\documentclass[11pt]{article}
\usepackage{microtype}
\usepackage{graphicx}
\usepackage[top=1in,bottom=1in,left=1in,right=1in]{geometry}
\usepackage{natbib} 
\usepackage{amsfonts, times, amsmath, amssymb, amsthm, constants, bbm}
\usepackage{MnSymbol}
\usepackage{mathtools,caption}
\usepackage[mathscr]{eucal}
\usepackage{graphicx,url,fullpage}
\usepackage{color,times,float}
\usepackage{setspace}

\usepackage{amsmath}		
\usepackage{amssymb}		
\usepackage{amsfonts}		
\usepackage{amsthm}		    

\usepackage{mathtools}		
\mathtoolsset{%
    above-intertext-sep = -1.5ex,
    below-intertext-sep = -2ex
}

\usepackage[utf8]{inputenc}		
\usepackage[T1]{fontenc}		

\usepackage{dsfont}		

\usepackage{inconsolata}

\usepackage[
cal=cm,
]
{mathalfa}

\usepackage{wrapfig}
\usepackage[labelfont={bf,small},labelsep=colon,font=small, singlelinecheck=false]{caption}	
\usepackage[dvipsnames,svgnames]{xcolor}		

\usepackage{placeins}

\usepackage{wasysym}		

\usepackage{fancybox}		
\usepackage{subcaption}		
\usepackage{tikz}		
\usetikzlibrary{calc,patterns}

\usepackage{array}		
\usepackage{tabularx,booktabs}		
\usepackage[inline,shortlabels]{enumitem}		
\usepackage{multirow}
\usepackage{threeparttable}   

\usepackage{microtype}		

\usepackage{acronym}		
\usepackage{latexsym}		
\usepackage{xfrac}		
\usepackage{xspace}		
\usepackage{xparse}     

\usepackage{titletoc}

\usepackage{colortbl}

\definecolor{Gray}{gray}{0.85}
\definecolor{LightCyan}{rgb}{0.88,1,1}

\newcolumntype{a}{>{\columncolor{Gray}}c}

\usepackage{hyperref}
\hypersetup{
colorlinks=true,
linktocpage=true,
pdfstartview=FitH,
breaklinks=true,
pdfpagemode=UseNone,
pageanchor=true,
pdfpagemode=UseOutlines,
plainpages=false,
bookmarksnumbered,
bookmarksopen=false,
bookmarksopenlevel=1,
hypertexnames=true,
pdfhighlight=/O,
urlcolor=Maroon,linkcolor=MediumBlue!90!black,citecolor=MediumBlue!90!black,	
pdftitle={},
pdfauthor={},
pdfsubject={},
pdfkeywords={},
pdfcreator={pdfLaTeX},
pdfproducer={LaTeX with hyperref}
}
\usepackage[sort&compress,capitalize,nameinlink]{cleveref}		
\crefname{figure}{Figure}{Figure}
\crefname{assumption}{Assumption}{Assumptions}
\crefname{assumptionloc}{Assumption}{Assumptions}

\usepackage{algorithm}		
\usepackage{algorithmic}

\usepackage{thmtools,thm-restate,thm-autoref}

\theoremstyle{plain}
\newtheorem{theorem}{Theorem}		
\newtheorem{lemma}{Lemma}		
\newtheorem{proposition}{Proposition}		

\newtheorem*{corollary*}{Corollary}		

\theoremstyle{definition}
\newtheorem{assumption}{Assumption}		

\newtheorem*{definition*}{Definition}		
\newtheorem*{assumption*}{Assumptions}		

\makeatletter		
\newcommand{\asmtag}[1]{
  \let\oldtheassumption\theassumption
  \renewcommand{\theassumption}{#1}
  \g@addto@macro\endassumption{
    \addtocounter{assumption}{-1}
    \global\let\theassumption\oldtheassumption}
  }
\makeatother

\makeatletter		
\newcommand{\algtag}[1]{
  \let\oldthealgorithm\thealgorithm
  \renewcommand{\thealgorithm}{#1}
  \g@addto@macro\endalgorithm{
    \addtocounter{algorithm}{0}
    \global\let\thealgorithm\oldthealgorithm}
  }
\makeatother

\theoremstyle{remark}
\newtheorem{remark}{Remark}		
\newtheorem*{remark*}{Remark}		

\usepackage{hhline}
\usepackage{makecell}

\usepackage{macros}

\makeatletter
\newcommand{\smallsym}[2]{#1{\mathpalette\make@small@sym{#2}}}
\newcommand{\make@small@sym}[2]{%
  \vcenter{\hbox{$\m@th\downgrade@style#1#2$}}%
}
\newcommand{\downgrade@style}[1]{%
  \ifx#1\displaystyle\scriptstyle\else
    \ifx#1\textstyle\scriptstyle\else
      \scriptscriptstyle
  \fi\fi
}
\makeatother

\newcommand{\smallin}{\mathrel{\scalebox{0.9}[.9]{$\in$}}}

\setenumerate{itemsep=4pt, parsep=0pt, leftmargin=*}

\title{Uplifting Bandits}

\author{%
    Yu-Guan Hsieh\thanks{Université Grenoble Alpes, \texttt{yu-guan.hsieh@univ-grenoble-alpes.fr}. Work done during internship at Amazon.}
   \and
    Shiva Prasad Kasiviswanathan\thanks{Amazon, \texttt{kasivisw@gmail.com}.}
    \and
    Branislav Kveton\thanks{Amazon, \texttt{bkveton@amazon.com}.}
}
\date{}

\begin{document}

\maketitle


\begin{abstract}
We introduce a multi-armed bandit model where the reward is a sum of multiple random variables, and each action only alters the distributions of some of them. After each action, the agent observes the realizations of all the variables. This model is motivated by marketing campaigns and recommender systems, where the variables represent outcomes on individual customers, such as clicks. We propose UCB-style algorithms that estimate the \emph{uplifts} of the actions over a baseline. We study multiple variants of the problem, including when the baseline and affected variables are unknown, and prove sublinear regret bounds for all of these. We also provide lower bounds that justify the necessity of our modeling assumptions. Experiments on synthetic and real-world datasets show the benefit of methods that estimate the uplifts over policies that do not use this structure.

\end{abstract}

\newpage
\begin{spacing}{0.85}
\tableofcontents
\end{spacing}
\newpage

\section{Introduction}
\label{sec:intro}

\Ac{MAB} is an important framework for sequential decision making under uncertainty~\citep{LR85,bubeck2012regret,LS20}. In this problem, a learner repeatedly takes action and receives their rewards, while the outcomes of the other actions are unobserved. The goal of the learner is to maximize their cumulative reward over time by balancing exploration (select actions with uncertain reward estimates) and exploitation (select actions with high reward estimates). \ac{MAB} has applications in online advertisement, recommender systems, portfolio management, and dynamic channel selection in wireless networks.

One prominent question in the \ac{MAB} literature is how the dependencies between the actions can be exploited to improve statistical efficiency. Popular examples working toward this goal include {\em linear bandits}~\citep{dani2008stochastic} and {\em combinatorial bandits}~\citep{cesa2012combinatorial}.
In this work, we study a different structured bandit problem with the following three features:
\begin{enumerate*}[\textup(\itshape i\textup)]
    \item the reward is the sum of the payoffs of a fixed set of variables; 
    \item these payoffs are observed; and
    \item each action only affects a small subset of the variables.
\end{enumerate*}
This structure arises in many applications, as discussed below.
%
%
\begin{enumerate}[label=\textbf{(\alph*)}, wide=0pt]

\item \textbf{Online Marketing.}
\label{ex:marketing}
An ecommerce platform can opt among several marketing strategies (actions) to encourage customers to make more purchases on their website.
As different customers can be sensitive to a different marketing strategy, regarding each of them as a variable, it is natural to expect that each action would likely affect only a subset of the variables. 
The payoff associated to a customer can be for example the revenue generated by that customer; then the reward is just the total revenue received by the platform.

\item\textbf{Product Discount.} 
\label{ex:product}
Consider a company that uses discount strategies to increase their sales. It is common to design discount strategies that only apply to a small subset of products. In this case, we can view the sales of each product as a variable and each discount strategy as an action, and assume that each action only has a significant impact on the sales of those products discounted by this action.

\item \textbf{Movie Recommendation.}
\label{ex:movie}
Consider a bandit model for movie recommendation where actions correspond to different recommendation algorithms, and the variables are all the movies in the catalog. For each user, define a set of binary payoffs that indicate whether the user watches a movie in the catalog, and the reward is the number of movies from the catalog that this user watches. Since a recommendation (action) for this user contains (promotes) only a subset of movies, it is reasonable to assume that only their associated payoffs are affected by the action.
\end{enumerate}

\paragraph{Our Contributions.} To begin with, we formalize an \emph{uplifting bandit} model that captures the aforementioned structure.
The term \emph{uplift} is borrowed from the field of {\em uplift modeling}~\citep{radcliffe2007using,gutierrez2017causal} and indicates that the actions' effects are incremental over a certain baseline.
We consider a stochastic model, that is, the payoffs of the variables follow an unknown distribution that depends on the chosen action,
and study this problem under various assumptions on the learners' prior knowledge about
    \begin{enumerate*}[\textup(\bf 1\textup)]
    \item \label{a} the baseline payoff of each variable and\item \label{b} the set of affected variables of each action. 
    \end{enumerate*}
Our first result (\cref{sec:UpUCB-BL}) shows that when both \ref{a} and \ref{b} are known, a simple modification of the \ac{UCB} algorithm~\citep{auer2002using} (Algorithm~\UpUCB) for estimating the uplifts has a much lower regret than its standard implementation. Roughly speaking, when $\nVariables$ is the number of variables and $\nAffected$ is the number of variables affected by each action, we get a $O(\nAffected^2)$ regret bound instead of $O(\nVariables^2)$. 
This results in major regret reduction for $\nAffected \ll \nVariables$, and is a distinguishing feature of all our results.
    
Going one step further, we design algorithms that have minimax optimal regret bounds without assuming that either~\ref{a} or~\ref{b} is known 
in \cref{sec:known-affected,sec:unknown-affected}.
When ~\ref{a} the baseline payoffs are unknown and \ref{b} the affected variables are known, we compute differences of \ac{UCB}s to estimate the uplifts (Algorithm~\UpUCBwb).
In contrast to standard \ac{UCB} methods, these differences are \emph{not} optimistic; this is because the feedback of the actions also reveals important information about the baseline.
When ~\ref{b} the affected variables are  
also unknown, we identify them on the fly to maintain suitable estimates of the uplifts. We study two approaches, which differ in what they know about the affected variables. Algorithm~\UpUCBLwb knows an upper bound on number of affected variables, whereas Algorithm~\UpUCBDeltawb knows a lower bound on individual uplift. Our regret bounds are summarized in \cref{tab:summary}.
These results are further completed with lower bounds that justify the need for our modeling assumptions (\cref{sec:lower}).

    


\begin{table*}[!t]
\renewcommand{\arraystretch}{1.3}
\setlength\tabcolsep{6pt}
\begin{center}
\begin{tabularx}{\textwidth}{c|a|cc|c|c}
\toprule
Algorithm & UCB & \UpUCB & \UpUCBwb & \UpUCBLwb &  \UpUCBDeltawb     \\ \hline \hline
Affected variables known & No & Yes & Yes & No & No \\ 
Baseline payoffs known & No & Yes & No &  No &  No \\[2pt] 
\hline
\rule{0ex}{4ex}
Regret Bound
& $\displaystyle \frac{\nArms\nVariables^2}{\gapbandit}$ 
&
\multicolumn{2}{c|}{ $\displaystyle \frac{\nArms\nAffected^2}{\gapbandit}$} & 
\multicolumn{1}{c|}{ $\displaystyle \frac{\nArms\nAffected^2}{\gapbandit}$} & 
\multicolumn{1}{c}{$\displaystyle \frac{\nArms\clip(\gapbandit/\gapUplift,\nAffected,\nVariables)^2}{\gapbandit}$} 
\\[0.4em] \bottomrule
\end{tabularx}
\caption{Summary of our regret bounds for uplifting bandits. Constant and logarithmic factors are ignored throughout.
For simplicity, we assume here all actions affect exactly $\nAffected$ of the $\nVariables$ variables and all the suboptimal actions have the same suboptimality gap $\gapbandit$. $K$ and $\gapUplift$ are respectively the number of actions and a lower bound on individual uplift ($\gapUplift$ is formally introduced in \cref{sec:part2}).
The operator $\clip$ restricts the value of its first variable to the range defined by its second and third variable, \ie $\clip(x,\alpha,\beta)=\max(\alpha,\min(\beta,x))$.}
\vspace{-1.4em}
\label{tab:summary}
\end{center}
\end{table*}

\paragraph{Organization.} We introduce our uplifting bandit model along with its various variations in~\cref{sec:model}. Over Sections~\ref{sec:UpUCB-BL},~\ref{sec:known-affected},~\ref{sec:unknown-affected}, we provide regret bounds for these variations. Lower bounds and numerical experiments are presented in~\cref{sec:lower} and \cref{sec:exp}.
To demonstrate the generality of our setup and how our algorithmic ideas extend beyond vanilla multi-armed bandits, we further discuss contextual extensions of our model in \cref{sec:context}.
Comparison to related work is deferred to \cref{sec:related}.

\section{Problem Description}
\label{sec:model}
We start by formally introducing our uplifting bandit model. We illustrate it in~\cref{fig:model} and summarize our notations in~\cref{tab:notation}. Contextual extensions of this model are discussed in \cref{sec:context}.

A $(\nArms,\nVariables)$-uplifting bandit is
a stochastic bandit with $\nArms$ actions and $\nVariables$ underlying variables.
Each action $\arm\in\arms\defeq\intinterval{1}{\nArms}$ is associated with a distribution $\va[\distribution]$ on $\R^{\nVariables}$. 
At each round $\run$, the learner chooses an action $\vt[\arm]\in\arms$
and receives reward $\vt[\reward]=\sum_{\indv\in\variables}\vtv$ where
$\variables\defeq\intinterval{1}{\nVariables}$ the set of all variables
and $\vt[\valuev]=(\vtv[\valuev])_{\indv\in\variables}\sim\va[\distribution][\vt[\arm]]$ 
is the payoff vector.\footnote{The terms \emph{reward} and \emph{payoff} distinguish $\vt[\reward]$ and $(\vtv[\valuev])_{\allvariables}$.}
Our model is distinguished by the two assumptions that we describe below. 
%
\begin{enumerate}[label=\textbf{(\Roman*)}, wide=0pt]
\item \textbf{Limited Number of Affected Variables.}
\label{asm:limited}
Let $\va[\variables]\subseteq\variables$ be the subset of variables affected by action $\arm$ and $\va[\distribution][0]$ be the baseline distribution that the variables' payoffs follow when no action is taken.
By definition $\va[\distribution]$ and $\va[\distribution][0]$ have the same marginal distribution on $\setcomplement{\va[\variables]}\defeq\setexclude{\variables}{\va[\variables]}$, the variables unaffected by action $\arm$.\footnote{%
If the actions in fact have small impact on the variables in $\setcomplement{\va[\variables]}$,
our model is misspecified and incurs additional linear regret whose size is proportional to the impact of the actions on $\setcomplement{\va[\variables]}$.
An interesting question is how we can design algorithms that self-adapt to the degree of misspecification.\label{footnote:small-impact}}
While the above condition is always satisfied with  $\va[\variables]=\variables$, we are interested in the case of $\va[\nAffected]\defeq\card(\va[\variables])\ll\nVariables$, meaning only a few variables are affected by $\arm$.
We define $\nAffected$ as a uniform bound on $\va[\nAffected]$, so that $\max_{\arm\in\arms}\va[\nAffected]\le\nAffected$.
For convenience of notation, we write $\arms_0=\arms\union\{0\}$ and $\oneto{\nRuns}:=\intinterval{1}{\nRuns}$.

\item\textbf{Observability of Individual Payoff.} 
\label{asm:observability}
In addition to the reward $\vt[\reward]$, we assume that the learner observes all the payoffs $(\vtv[\valuev])_{\indv\in\variables}$ after an action is taken in round $\run$.


%
\end{enumerate}
%

\paragraph{Uplift and Noise\afterhead}
Let $\va[\valuev]=(\vav[\valuev])_{\allvariables}$ be a random variable with distribution $\va[\distribution]$. 
We define $\vav[\meanReward]=\ex[\vav[\valuev]]$ and $\vav[\snoise]=\vav[\valuev]-\vav[\meanReward]$ respectively as the expected value of and the noise in $\vav[\valuev]$.
We use $\va[\meanReward]$ (resp.\ $\vav[\meanReward][0]$) to denote the vector of $(\vav[\meanReward])_{\allvariables}$ (resp.\ $(\vav[\meanReward][0])_{\allvariables}$), and refer to $\va[\meanReward][0]$ as the baseline payoffs.
The \emph{individual} uplift associated to a pair $(\arm,\indv)\in\arms\times\variables$
is defined as $\vav[\meanUplift] =\vav[\meanReward] - \vav[\meanReward][0]$. An individual uplift can be \emph{positive} or \emph{negative}.
We obtain the (total) uplift of an action by summing its individual uplifts over all the variables affected by that action
\begin{equation}
\label{eq:total-uplift}
\va[\uplift]
=\sum_{\indv\in\va[\variables]}\vav[\meanUplift]
=\sum_{\indv\in\va[\variables]}(\vav[\meanReward]-\vav[\meanReward][0]).
\end{equation}
%
Let $\va[\reward]=\sum_{\indv\in\variables}\vav[\meanReward]$ be the expected reward of an action or of pure observation. We also have $\va[\uplift]=\va[\reward]-\va[\reward][0]$
since $\vav[\meanReward]=\vav[\meanReward][0]$ as long as $\indv\notin\va[\variables]$.

A real-value random variable $\rv$ is said to be $\noisedev$-sub-Gaussian 
if for all $\scalar\in\R$, it holds $\ex[\exp(\scalar\rv)]\le\exp(\noisevar\scalar^2/2)$.
Throughout the paper, we assume that $\vav[\snoise]$ is $1$-sub-Gaussian for all $\arm\in\arms_0$ and $\indv\in\variables$.
Note that we do not assume that $(\vav[\valuev])_{\indv\in\variables}$ are independent, i.e., the elements in the noise vector $(\vav[\snoise])_{\allvariables}$ may be correlated,
for the following two reasons: 
\begin{itemize}[leftmargin=*]
    \item The independence assumption is not always realistic.
    In our first example, it excludes any potential correlation between two customers' purchases.
    \item While a learner can exploit their knowledge on the noise covariance matrix to reduce the regret, as for example shown in \citep{DP16}, incorporating such knowledge complicates the algorithm design and the accompanying analysis.
    We however believe this is an interesting future direction to pursue.
\end{itemize}

\begin{figure*}[t]
\centering
\begin{minipage}[b]{.31\textwidth}
  \centering
   \includegraphics[width=0.94\textwidth]{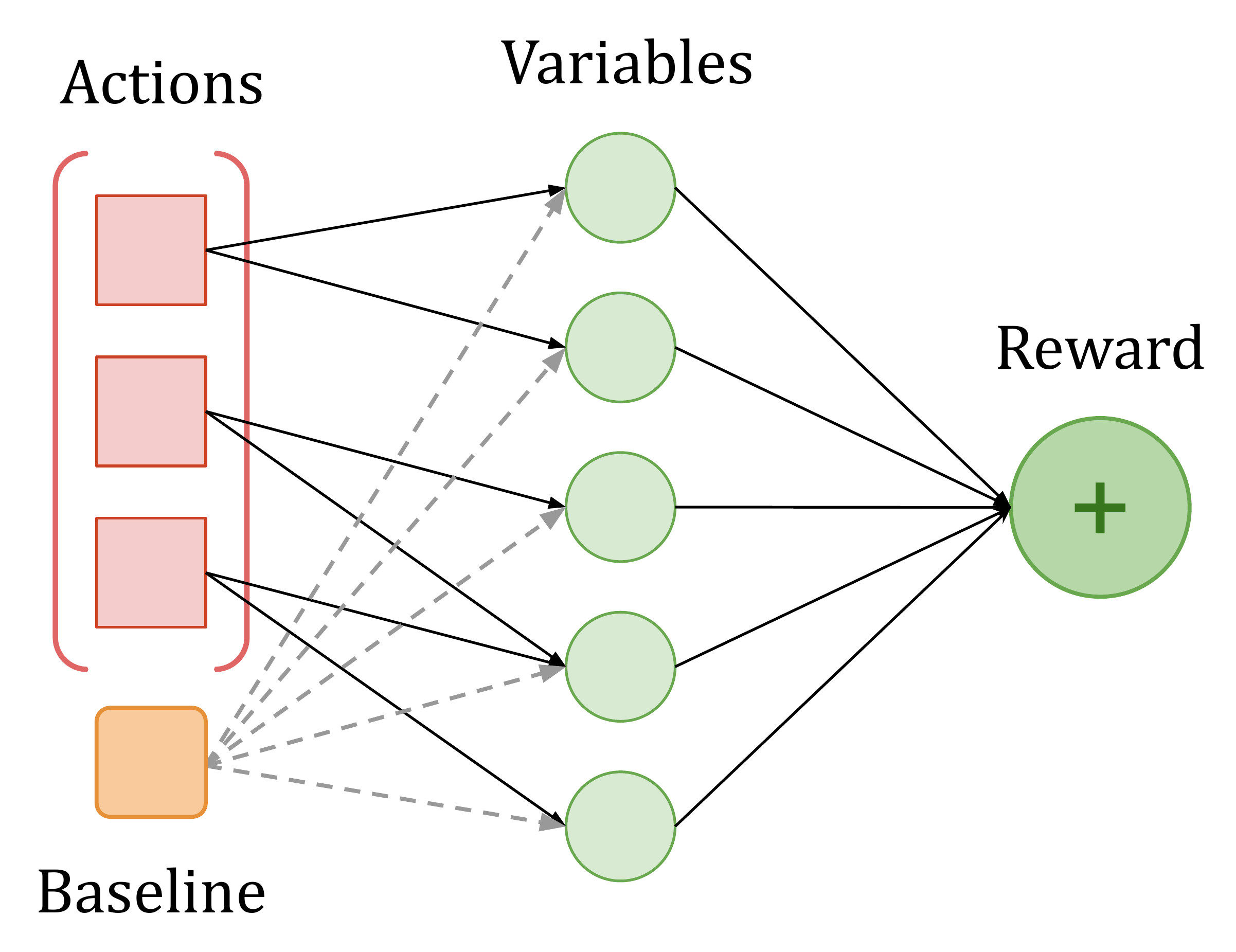}
  \captionof{figure}{Illustration of the uplifting bandit model. This example has $\nArms=3$ arms, $\nVariables=5$ variables and each action affects  $\va[\nAffected]=2$ variables.
  Dash lines indicate the variables' payoffs follow the baseline distribution $\va[\distribution][0]$ by default.} 
  \label{fig:model}
\end{minipage}
\hfill
\begin{minipage}[b]{.66\textwidth}
  \centering
  \begin{subfigure}{0.475\textwidth}
  \centering
  \includegraphics[width=0.86\textwidth]{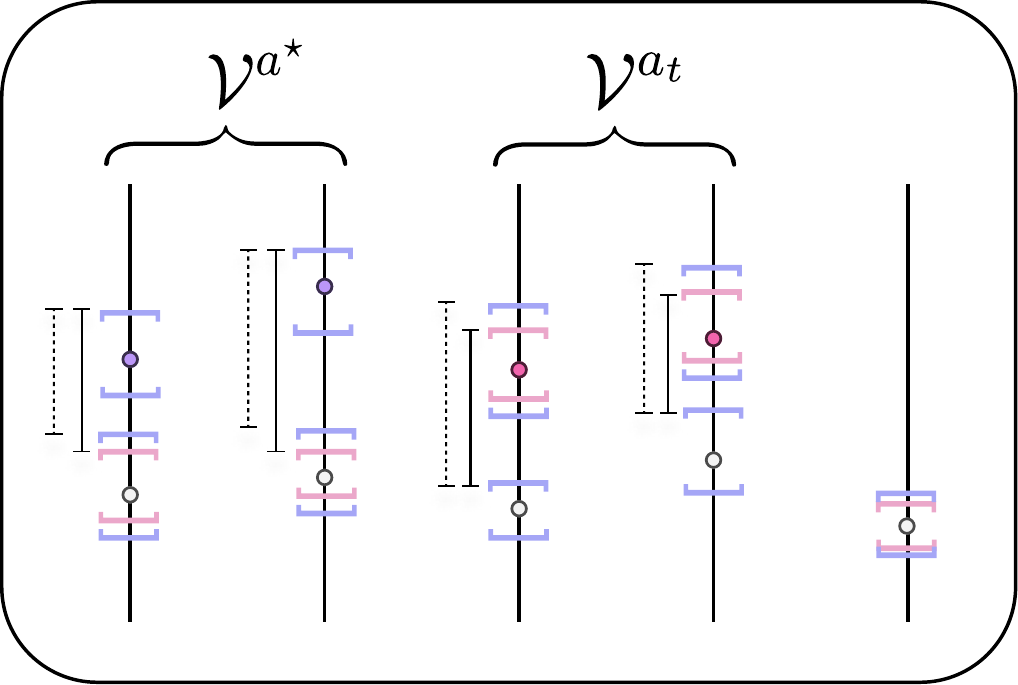}
  \vskip 5pt
  \caption{\UpUCBwb: After suboptimal action $\vt[\arm]$ is taken, the confidence intervals of $(\vav[\meanReward])_{\indv\in\va[\variables][\vt[\arm]]}$ and $(\vav[\meanReward][0])_{\indv\in\setexclude{\va[\variables][\sol[\arm]]}{\va[\variables][\vt[\arm]]}}$ shrink (from purple to pink). Hence the uplifting index of $\vt[\arm]$ decreases while that of $\sol[\arm]$ increases (from dash to solid).}
  \label{subfig:UpUCBwb}
  \end{subfigure}
  \hfill
  \begin{subfigure}{0.475\textwidth}
  \centering
  \includegraphics[width=0.86\textwidth]{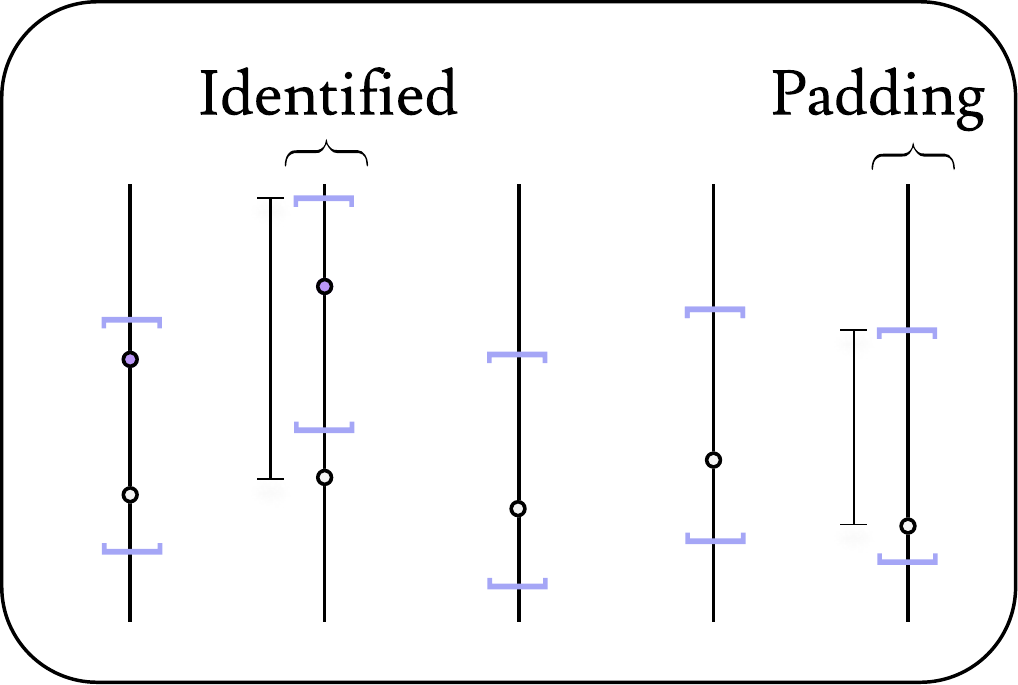}
  \vskip 5pt
  \caption{\UpUCBL: To compute the uplifting index of an action, we identify a set of variables whose associated confidence intervals do not contain the baseline payoff and pad it with the variables with the largest UCBs on uplifts.}
  \label{subfig:UpUCBL}
  \end{subfigure}
  \addtocounter{figure}{-1}
  \captionof{figure}{Explanation about \UpUCBwb and \UpUCBL using the model illustrated in \cref{fig:model}.}
  \label{fig:algorithms}
\end{minipage}
\vspace{-0.6em}
\end{figure*}

\paragraph{Regret.}
The learner's performance is characterized by their regret.
To define it, we denote by $\sol[\arm]\smallin\argmax_{\arm\in\arms}\va[\reward]$ an action with the largest expected reward and
$\sol[\reward]=\va[\reward][\sol[\arm]]=\max_{\arm\in\arms}\va[\reward]$
the largest expected reward.
The regret of the learner after $\nRuns$ rounds is then given by
\begin{equation}
    \label{eq:regret-decomp}
    \begin{aligned}[b]
    \reg_{\nRuns} &= \sol[\reward]\nRuns - \sum_{\run=1}^{\nRuns}\va[\reward][\vt[\arm]]
    = \sum_{\arm\in\arms}\sum_{\run=1}^{\nRuns}\one\{\vt[\arm]=\arm\}\va[\gapbandit],
    \end{aligned}
\end{equation}
where $\va[\gapbandit]=\sol[\reward]-\va[\reward]$ is the suboptimality gap of action $\arm$.
In words, we compare the learner's cumulative (expected) reward against the best we can achieved by taking  an optimal action at each round.
For posterity, we write $\gapbandit=\min_{\arm\in\arms,\va[\gapbandit]>0}\va[\gapbandit]$ for the minimum non-zero suboptimality gap and refer to it as the suboptimality gap of the problem.

\paragraph{UCB for Uplifting Bandits.}
The \ac{UCB} algorithm~\citep{auer2002using}, at each round, constructs an \acdef{UCB} on the expected reward 
of each action and chooses the action with the highest \ac{UCB}.
When applied to our model,
we get a regret of $\bigoh(\nArms\nVariables^2\log\nRuns/\gapbandit)$,
as the noise in the reward is $\nVariables$-sub-Gaussian (recall that we do not assume independence of the payoffs).
However, this approach completely ignores the structure of our problem.
As we show in \cref{sec:UpUCB-BL}, we can achieve much smaller regret by focusing on the uplifts.

%
%

\paragraph{Problem Variations.} In the following sections, we study several variants of the above basic problem that differ in the prior knowledge about $(\va[\distribution])_{\arm\in\arms_0}$ that the learner possesses.
\begin{enumerate}[label=\textbf{(\alph*)}, wide=0pt]
\item\textbf{Knowledge of Baseline Payoffs.} We consider two scenarios based on whether the learner knows the baseline payoffs $\va[\meanReward][0] \coloneqq (\vav[\meanReward][0])_{\allvariables}$ or not.
\item\textbf{Knowledge of Affected Variables.} We again consider two  scenarios based on whether the learner knows the affected variables associated with each action $(\va[\variables])_{\arm\in\arms}$ or not. 
\end{enumerate}

\vspace*{-1ex}
\section{Case of Known Baseline and Known Affected Variables}

\label{sec:UpUCB-BL}
We start with the simplest setting where both the affected variables and the baseline payoffs are known.
To address this problem, we make the crucial observation that
for any two actions, the difference in their rewards equals to that of their uplifts.
As an important consequence, uplift maximization has the same optimal action as total reward maximization, and we can replace rewards by uplifts in the definition of the regret \eqref{eq:regret-decomp}.
Formally, we write $\sol[\uplift]=\va[\uplift][\sol[\arm]]=\max_{\arm\in\arms}\va[\uplift]$ for the largest uplift; then 
\[\reg_{\nRuns} = \sol[\uplift]\nRuns - \sum_{\run=1}^{\nRuns}\va[\uplift][\vt[\arm]].\]
By making this transformation, we gain in statistical efficiency because
$\va[\uplift]=\ex[\sum_{\indv\in\va[\variables]}(\vav[\valuev]-\vav[\meanReward][0])]$ can now be estimated much more efficiently under our notion of sparsity.
Since both $\va[\meanReward][0]$ and $(\va[\variables])_{\arm\in\arms}$ are known, we can directly construct a \ac{UCB} on $\va[\uplift]$. For this, we define
for all rounds $\run\in\oneto{\nRuns}$, actions $\arm\in\arms$, and variables $\indv\in\variables$ the following quantities.
\begin{equation}
    \label{eq:mean-estimate-a}
    \begin{aligned}[b]
    \vta[\pullcount] = \sum_{\runalt=1}^{\run}\one\{\vt[\arm][\runalt]=\arm\},
    ~~~
    \vta[\radiusConf] = \sqrt{
    \frac{2\log(1/\alt{\smallproba})}{\vta[\pullcount]}},
    ~~~
    \vtav[\est{\meanReward}] =
    \sum_{\runalt=1}^{\run}
    \frac{\vtv[\valuev][\runalt]\one\{\vt[\arm][\runalt]=\arm\}}{\max(1, \vta[\pullcount])},
    \end{aligned}
\end{equation}
where $\alt{\smallproba}>0$ is a tunable parameter.
In words, $\vta[\pullcount]$, $\vtav[\est{\meanReward}]$, and $\vta[\radiusConf]$ represent respectively the number of times that action $\arm$ is taken, the empirical estimate of $\vav[\meanReward]$, and the associated radius of \acl{CI} calculated at the end of round $\run$.
The UCB for action $\arm\in\arms$ and variable $\indv\in\va[\variables]$ at round $\run$ is $\vtav[\UCBindex]=\vtav[\est{\meanReward}][\run-1]+\vta[\radiusConf][\run-1]$.
We further define $\vta[\UPindexSum]=\sum_{\indv\in\va[\variables]}(\vtav[\UCBindex]-\vav[\meanReward][0])$ as the \emph{uplifting index},
and refer to \UpUCB as the algorithm that takes an action with the largest uplifting index $\vta[\UPindexSum]$ at each round $\run$ (\cref{algo:UpUCB-BL}).
Here, the suffix (bl) indicates that the method operates with the knowledge of the baseline payoffs.

\begin{algorithm}[t!]
    \caption{\UpUCB} 
    \label{algo:UpUCB-BL}
\begin{algorithmic}[1]
    \STATE {\bfseries Input:} Error probability $\alt{\smallproba}$, Baseline payoffs $\va[\meanReward][0]$, Sets of affected variables $\setdef{\va[\variables]}{\arm\in\arms}$ 
    \STATE {\bfseries Initialization:} Take each action once
    \FOR{$\run = \nArms+1, \ldots, \nRuns$}
    \STATE Compute the statistics following \eqref{eq:mean-estimate-a}
    \STATE For $\arm\in\arms$, compute uplifting index $\vta[\UPindexSum]\subs\sum_{\indv\in\va[\variables]}(\vtav[\est{\meanReward}][\run-1]+\vta[\radiusConf][\run-1]-\vav[\meanReward][0])$
    \STATE Select action $\vt[\arm]\in\argmax_{\arm\in\arms}\vta[\UPindexSum]$
    \ENDFOR
\end{algorithmic}
\end{algorithm}

Since \UpUCB is nothing but a standard UCB with transformed rewards $\vt[\alt{\reward}]=\sum_{\indv\in\vt[\variables][\vt[\arm]]}(\vtv[\valuev]-\vav[\meanReward][0])$, and 
$\ex[\vt[\alt{\reward}]]=\va[\uplift][\vt[\arm]]$ defines exactly the same regret as the original reward, a standard analysis for \ac{UCB} yields the following result.

\begin{proposition}
\label{prop:UpUCB-BL-regret}
Let $\alt{\smallproba}=\smallproba/(2\nArms\nRuns)$. Then the regret of \UpUCB (\cref{algo:UpUCB-BL}), with probability at least $1-\smallproba$, satisfies:
\begin{equation}
     \label{eq:UpUCB-BL-regret}
    \vt[\reg][\nRuns] \le
    \sum_{\arm\in\arms:\va[\gapbandit]>0}
    \left(\frac{8(\va[\nAffected])^2\log(2\nArms\nRuns/\smallproba)}{\va[\gapbandit]}
    +\va[\gapbandit]\right).
\end{equation}
\end{proposition} 

%
As expected, the regret bound does not depend on $\nVariables$ and scales with $\nAffected^2$.
This is because the transformed reward of action $\arm$ is only $\va[\nAffected]$-sub-Gaussian.
The improvement is significant when $\nAffected\ll\nVariables$. 
However, this also comes at a price, as both the baseline payoffs and affected variables need to be known.
We address these shortcomings in \cref{sec:known-affected,sec:unknown-affected}.

\paragraph{Estimating Baseline Payoffs from Observational Data.}
In practice, the baseline payoffs $\va[\meanReward][0]$ can be estimated from observational data, which gives confidence intervals where $\vav[\meanReward][0]$ lie with high probability.
The uplifting indices can then be constructed by subtracting the lower confidence bounds of $\vav[\meanReward][0]$ from $\vtav[\UCBindex]$.
If the number of samples is $\nSamples$, the widths of the confidence intervals are in $\bigoh(1/\sqrt{\nSamples})$.
Identification of $a^*$ is possible only when the sum of these widths over $\nAffected$ variables is at most $\gapbandit$, which requires $\nSamples=\Omega(\nAffected^2/\gapbandit^2)$.
Otherwise, these errors persist at each iteration and $\nSamples$ must be in the order of $\nRuns$ to ensure $\bigoh(\sqrt{\nRuns})$ regret.

\paragraph{About Expected and Gap-Free Bounds.}
In \cref{prop:UpUCB-BL-regret}, we state a high-probability instance-dependent regret bound, and we will continue to do so for all the regret upper bounds that we present for the non-contextual variants.
This type of result can be directly transformed to bound on $\ex[\vt[\reg][\nRuns]]$ by taking $\smallproba=1/\nRuns$.
Following the routine of separating $\va[\gapbandit]$ into two groups depending on their scale, most of our proofs can also be easily modified to obtain a gap-free regret bound, which is typically in the order of $\bigoh(\nAffected\sqrt{\nArms\nRuns\log(\nRuns)})$.
We will not state these results to avoid unnecessary repetitions.



\section{Lower Bounds}
\label{sec:lower}

In this section, we discuss the necessity of our modeling assumptions for obtaining the improved regret bounds of \cref{prop:UpUCB-BL-regret}.
We prove these results in  \cref{apx:lower-bound}, where we also introduce a general information-theoretic lower bounds for bandit problems with similar design.

Intuitively, the regret can be improved both because the noise in the effect of an action is small, and because the observation of this effect does not heavily deteriorate with noise.
These two points correspond respectively to assumptions \ref{asm:limited} and \ref{asm:observability}.
Moreover, the knowledge on $(\va[\variables])_{\arm\in\arms}$ allows the learner to distinguish between problems with different structures.
Without such distinction, 
there is no chance that the learner can leverage the underlying structure.
Therefore, the aforementioned three points are crucial for obtaining \eqref{eq:UpUCB-BL-regret}.
Below, we establish this formally for algorithms that are {\em consistent} \citep{LR85} over the class of $1$-sub-Gaussian uplifting bandits,
which means the induced regret of the algorithm on any uplifting bandit with $1$-sub-Gaussian noise satisfies $\vt[\reg]=\smalloh(\run^p)$ for all $p>0$.

For conciseness, in the following we say that an uplifting bandit has parameters $(\nArms,\nVariables,(\va[\gapbandit],\va[\nAffected])_{\arm\in\arms})$ if the number of actions, the number of variables, the suboptimality gaps, and the number of affected variables of this instance are respectively $\nArms$, $\nVariables$, $(\va[\gapbandit])_{\arm\in\arms}$, and $(\va[\nAffected])_{\arm\in\arms}$.
We focus on instance-dependent lower-bounds and assume that the learner has full knowledge of the baseline distribution. (Of course, the problem only becomes more challenging if the learner does not know the baseline distribution.)

\begin{proposition}
\label{prop:lower-bound}
Let $\policy$ be a consistent algorithm over the class of $1$-sub-Gaussian uplifting bandits that at most uses knowledge of $\va[\distribution][0]$, $(\va[\variables])_{\arm\in\arms}$, and the fact that the noise is $1$-sub-Gaussian.
Let $\nArms,\nVariables>0$ and sequence $(\va[\nAffected],\va[\gapbandit])_{1\le\arm\le\nArms}\in(\oneto{\nVariables}\times\R_+)^{\nArms}$ satisfy $\va[\gapbandit][1]=0$.
Assume either of the following holds.
\begin{enumerate}[(a)]
    \item 
    \label{low-cond:affected}
    $\va[\nAffected]=\nVariables$ for all $\arm\in\arms$, so that in the bandits considered below all actions affect all variables.
    \item
    \label{low-cond:observed}
    Only the reward is observed.
    \item
    \label{low-cond:knowledge}
    The algorithm $\policy$ does not make use of any prior knowledge about the arms' expected payoffs $(\va[\meanReward])_{\arm\in\arms}$ (in particular, the knowledge of $(\va[\variables])_{\allvariables}$ is not used by $\policy$).
\end{enumerate}
Then, there exists a $1$-sub-Gaussian uplifting bandit 
with parameters $(\nArms,\nVariables,(\va[\gapbandit],\va[\nAffected])_{\arm\in\arms})$
such that the regret induced by $\policy$ on it satisfies: 
\[
\liminf_{\toinf[\nRuns]} \frac{\ex[\vt[\reg][\nRuns]]}{\log\nRuns} \ge \sum_{\arm\in\arms:\va[\gapbandit]>0}\frac{2\nVariables^2}{\va[\gapbandit]}.
\]
\end{proposition}

As discussed above, \cref{prop:lower-bound} shows that there would be a higher regret in absence of any of the following: \begin{enumerate*}[\textup(a\textup)]
\item Limited number of affected variables,
\item Side information about variables beyond reward, and
\item Some prior knowledge on how the variables are affected.
\end{enumerate*}
We discuss below these three points in detail.
\begin{enumerate}[(a), wide, topsep=4pt]
    \item This point justifies the necessity of assumption \ref{asm:limited}.
    It can be further refined to account for non-uniform numbers of affected variables.
\vspace*{-3pt}
\begin{proposition}
\label{prop:lower-bound-L}
Let $\policy$ be a consistent algorithm over the class of $1$-sub-Gaussian uplifting bandits that at most uses knowledge of $\va[\distribution][0]$, $(\va[\variables])_{\arm\in\arms}$, and the fact that the noise is $1$-sub-Gaussian.
Then, for any $\nArms,\nVariables>0$ and sequence $(\va[\nAffected],\va[\gapbandit])_{1\le\arm\le\nArms}\in(\oneto{\nVariables}\times\R_+)^{\nArms}$
with $\va[\gapbandit][1]=0$, there exists a $1$-sub-Gaussian uplifting bandit $(\nArms,\nVariables,(\va[\gapbandit],\va[\nAffected])_{\arm\in\arms})$
such that the regret induced by $\policy$ on it satisfies
\begin{equation}
    \label{eq:lower-bound-L-1}
    \liminf_{\toinf[\nRuns]} \frac{\ex[\vt[\reg][\nRuns]]}{\log\nRuns} \ge \sum_{\arm\in\arms:\va[\gapbandit]>0}\frac{2(\va[\nAffected])^2}{\va[\gapbandit]}.
\end{equation} 
\end{proposition}
%
\cref{prop:lower-bound-L} essentially posits that the regret must scale with the magnitudes of the noises.
The stated lower bound \eqref{eq:lower-bound-L-1} matches an expected version of \eqref{eq:UpUCB-BL-regret} up to a constant factor,
suggesting the minimax optimatlity of \cref{prop:UpUCB-L-BL-regret}.
One way to argue this lower bound holds is by reducing a carefully designed uplifting bandits with parameters $(\nArms,\nVariables,(\va[\gapbandit],\va[\nAffected])_{\arm\in\arms})$ to a $\nArms$-arm bandit with the corresponding noise scales.
We will take a different approach in \cref{apx:lower-bound} that allows us prove all the lower bounds following the same schema.

\item This point partially justifies assumption \ref{asm:observability}.
Note that it is however not always necessary to observe all the payoffs. For example, observing the aggregated payoff $\sum_{\indv\in\va[\variables][\vt[\arm]]}\vtv[\valuev]$ is sufficient for computing the uplifting indices of \UpUCB.
Having observation of the entire payoff vector $(\vtv[\valuev])_{\indv\in\variables}$ is particularly relevant for the case of unknown affected variables since it helps identify the affected variables relatively easily, as we will see in \cref{sec:unknown-affected}.

\item
This point justifies the learner using knowledge about the affected variables.
To get some more intuition on the meaning of this condition, let us consider two bandits with exactly the same baseline and noise distributions and very close expected payoffs of the arms.
Then, with finite observations, they provide very similar feedback which prevents the learner from distinguishing between the two problems.
However, the two bandits may have completely different structures, and in particular, one of them may have all variables affected by all actions since slight perturbation to the baseline would mean that a variable is affected.
Therefore, the lower bound from \ref{low-cond:affected} applies to this instance.
Now, provided that the learner is doomed to treat the two instances equally, the lower bound applies to the other instance as well.
Of course, one can argue that it is unnatural that the regrets of two similar problems differ significantly even when the learner is provided some prior knowledge.
We believe this is the drawback of the asymptotic analysis that we look at here, since these differences may indeed be small within a reasonable number of runs. This is exactly related to the problem of misspecified model that we mentioned in \cref{footnote:small-impact}.
\end{enumerate}

With \cref{prop:lower-bound}, what remains unclear at this point is whether similar improvement of the regret is still possible when the baseline is unknown or when the learner only has access to more restricted knowledge than $(\va[\variables])_{\arm\in\arms}$.
We give affirmative answers to these two questions in the next two sections.

\vspace{0.4em}
\begin{remark}
Our lower bounds are derived with respect to the worst correlation structure of the noise. It is also possible to obtain lower bounds that depend on the covariance of the noise. 
For example, if the variables $(\vav[\valuev])_{\allvariables}$ are independent, the lower bound in \eqref{eq:lower-bound-L-1} only scales linearly with $\va[\nAffected]$.
\end{remark}


%

\section{Case of Unknown Baseline}
\label{sec:known-affected}

In this section, we consider the situation where the sets of affected variables $(\va[\variables])_{\arm\in\arms}$ are known by the learner, but not the baseline payoffs. Missing proofs are in~\cref{apx:upper-bound}.

Since the actual uplift at each round is never observed, the uplifts of the actions can hardly be estimated directly without knowing the baseline payoffs.
Instead, we follow a two-model approach.
Leveraging the fact that $\va[\distribution]$ and $\va[\distribution][0]$ have the same marginal distribution on $\setcomplement{\va[\variables]}$, we can estimate the baseline payoffs by aggregating information from the feedback of different actions. This leads to
\begin{equation}
    \label{eq:mean-estimate-0}
    \vtav[\pullcount][\run][0] =
    \sum_{\runalt=1}^{\run}\one\{\indv\notin\va[\variables][\vt[\arm][\runalt]]\},
    ~~~
    \vtav[\radiusConf][\run][0] = 
    \sqrt{\frac{2\log(1/\alt{\smallproba})} {\vtav[\pullcount][\run][0]}},
    ~~~
    \vtav[\est{\meanReward}][\run][0] =
    \frac{\sum_{\runalt=1}^{\run}\vtv[\valuev][\runalt]\one\{\indv\notin\va[\variables][\vt[\arm][\runalt]]\}}
    {\max(1, \vtav[\pullcount][\run][0])}.
\end{equation} 
Compared to \eqref{eq:mean-estimate-a}, we notice that both $\vta[\pullcount][\run][0]$ and $\vta[\radiusConf][\run][0]$ are functions of $\indv$.
This is because for each taken action $\arm$, we only increase the counters $\vtav[\pullcount][\run][0]$ for those $\indv\notin\va[\variables]$, which causes a non-uniform increase of $(\vtav[\pullcount][\run][0])_{\allvariables}$.
Then, by looking at all the rounds that variable $\indv$ is not influenced by the chosen action, we manage to compute $\vtav[\est{\meanReward}][\run][0]$, an estimate of $\vav[\meanReward][0]$.
An alternative here is to consider a specific `action $0$' and estimate the baseline payoffs only using feedback from this action.
However, this necessarily requires action $0$ to be taken sufficiently often and would thus result in a very high regret.
As we will see, our algorithm does not make use of this specific action and achieves regret comparable to the one presented in \cref{prop:UpUCB-BL-regret}.
To proceed, we define the following UCB indices for all the pairs $(\arm,\indv)\in\arms_0\times\variables$
\begin{gather}
    \label{eq:UCBs}
    \vtav[\UCBindex] = \begin{cases}
    0
    & \text{if} ~ \arm=0 ~\text{and}~ \indv\in\bigcap_{\arm\in\arms}\va[\variables],
    \\
    \vtav[\est{\meanReward}][\run-1] + \vta[\radiusConf][\run-1] & \text{if} ~ \arm\in\arms ~\text{and}~ \indv\in\va[\variables], \\
    \vtav[\est{\meanReward}][\run-1][0] + \vtav[\radiusConf][\run-1][0]
    & \text{otherwise.}
    \end{cases}
    \raisetag{1em}
\end{gather} 
%
%
The second and the third lines of \eqref{eq:UCBs} contain the usual definition of UCBs using the empirical estimates and the radii of the confidence intervals defined in \eqref{eq:mean-estimate-a} and \eqref{eq:mean-estimate-0}.
In the special case that a variable is affected by all the actions
(recall that we do not use `action $0$'),
it is impossible to estimate $\vav[\UCBindex][0]$ but it is enough to compare $\vav[\UCBindex]$ directly against $\vav[\UCBindex][\armalt]$ for any two actions $\arm,\armalt\in\arms$, so we just set $\vtav[\UCBindex][\run][0]$ to $0$ in this case.


We outline the proposed method, \UpUCBwb, in~\cref{algo:UpUCB}. 
The uplifting indices are given by
$\vta[\UPindexSum]=\sum_{\indv\in\va[\variables]}(\vtav[\UCBindex]-\vtav[\UCBindex][\run][0])$.
It may be counter-intuitive to compare the differences between two \ac{UCB}s.
Indeed, $\vta[\UPindexSum]$ is no longer an optimistic estimate of the uplifting effect $\va[\uplift]$, but it captures the essential trade-off between learning action $\arm$'s payoffs and learning the baseline $\va[\meanReward][0]$. 
To provide some intuition, we give an informal justification of \UpUCBwb in \cref{subfig:UpUCBwb}:
If a suboptimal action $\arm$ is taken in round $\run$, the estimates of all $\vtav[\UCBindex]$ move closer to the actual mean from above. As a result, $\vta[\UPindexSum]$ decreases, since all $\vtav[\UCBindex]$ for $\indv\in\va[\variables]$ do. Thus action $\arm$ is less likely to be taken next. Moreover, $\vta[\UPindexSum][\run][\sol[\arm]]$ increases, since $\vtav[\UCBindex][\run][0]$ decrease for any $\indv$ affected by $\sol[\arm]$ but not $\arm$. Thus $\sol[\arm]$ is more likely to be taken next.
The effectiveness of \UpUCBwb is confirmed by the following theorem.

\begin{algorithm}[tb]
    \caption{\UpUCBwb} 
    \label{algo:UpUCB}
\begin{algorithmic}[1]
    \STATE {\bfseries Input:} Error probability $\alt{\smallproba}$, the sets of variables each action affects $\setdef{\va[\variables]}{\arm\in\arms}$ 
    \STATE {\bfseries Initialization:} Take each action once 
    \FOR{$\run = \nArms+1, \ldots, \nRuns$}
    \STATE Compute the UCB indices following \eqref{eq:UCBs}
    \STATE For $\arm\in\arms$, set $\vta[\UPindexSum]\subs\sum_{\indv\in\va[\variables]}(\vtav[\UCBindex]-\vtav[\UCBindex][\run][0])$
    \STATE Select action $\vt[\arm]\in\argmax_{\arm\in\arms}\vta[\UPindexSum]$
    \ENDFOR
\end{algorithmic}
\end{algorithm}



\vspace{0.2em}
\begin{restatable}{theorem}{UpUCBRegret}
\label{thm:UpUCB-regret}
Let $\alt{\smallproba}=\smallproba/(4\nArms\nAffected\nRuns)$. Then the regret of \UpUCBwb (\cref{algo:UpUCB}) with probability at least $1-\smallproba$, satisfies:
\vspace{-0.6em}
\begin{gather}
    \label{eq:UpUCB-regret}
    \vt[\reg][\nRuns] \le
    \sum_{\substack{\arm\in\arms:\va[\gapbandit]>0}}
    \left(\frac{8(\va[\nAffected]+\va[\nAffected][\sol[\arm]])^2\log(4\nArms\nAffected\nRuns/\smallproba)}{\va[\gapbandit]}
    +\va[\gapbandit]\right).
    \raisetag{0.6em}
\end{gather}
\end{restatable}

As in \cref{prop:UpUCB-BL-regret}, the regret is $\bigoh(\nArms\nAffected^2\log\nRuns/\gapbandit)$.
In fact, all decisions in \UpUCBwb are made on uncertain estimates of at most $\nAffected$ variables; thus the statistical efficiency scales with $\nAffected$ and not $\nVariables$.
Moreover, for any action $\arm$, the width of the confidence intervals for baseline estimates $\vtav[\radiusConf][\run][0]$
are lower than those of $\arm$ whenever $\indv\in\setcomplement{\va[\variables]}$. 
The factors $\vtav[\radiusConf][\run][0]$ thus play a secondary role in the analysis of the regret.
A detailed comparison with \cref{prop:UpUCB-BL-regret} reveals that the difference is in 
the order of $\nArms(\va[\nAffected][\sol[\arm]])^2\log\nRuns/\gapbandit$; this is only significant when 
the optimal actions affect many more variables then the suboptimal ones. 
Hence, the price of not knowing the baseline payoffs is generally quite small.

\section{Case of Unknown Affected Variables}
\label{sec:unknown-affected}
Now we study the more challenging setting where the affected variables $(\va[\variables])_{\arm\in\arms}$ are unknown to the learner.
\cref{prop:lower-bound} states that improvement is impossible if the learner does not possess any prior knowledge that allow them to exploit the structure.
To circumvent this negative result, we study two weak assumptions motivated by practice: the learner has either access to \begin{enumerate*}[\textup(\itshape i\textup)]
\item an upper bound on the number of affected variables or
\item a lower bound on individual uplift.
\end{enumerate*}



\subsection{Known Upper Bound on Number of Affected Variables}
\label{sec:part1}
Here we assume that we have a known upper bound on number of affected variables $\nAffected$ (\ie $\nAffected\ge\max_{\arm\in\arms}\va[\nAffected]$). We design algorithms with $\bigoh(\nArms\nAffected^2\log\nRuns/\gapbandit)$ regret bounds
that takes $\nAffected$ as input. We consider the cases of known and unknown baseline payoffs.

\subsubsection{Known Baseline Payoffs}
\label{subsubsec:UpUCB-L-BL}
To illustrate our ideas, we start by assuming that the baseline payoffs are known.
We propose an optimistic algorithm that maintains a \ac{UCB} on the total uplift with an overestimate in the order of $\nAffected$.
Let $\vta[\pullcount]$, $\vta[\radiusConf]$, and $\vtav[\est{\meanReward}]$ be defined as in \eqref{eq:mean-estimate-a}.
The UCB, uplifting indices, and the confidence intervals for each (action, variable) pair $(\arm,\indv)\in\arms\times\variables$ are 
\begin{equation}
    \label{eq:UCBs-arm}
    \begin{aligned}[b]
    \vtav[\UCBindex] &= \vtav[\est{\meanReward}][\run-1] + \vta[\radiusConf][\run-1],~~
    \vtav[\UPindex] = \vtav[\UCBindex]-\vav[\meanReward][0],\\
    \vtav[\intConf]&=[\vtav[\est{\meanReward}][\run-1]-\vta[\radiusConf][\run-1],\,\vtav[\est{\meanReward}][\run-1]+\vta[\radiusConf][\run-1]].
    \end{aligned}
\end{equation} 
%
In the sequel, we refer to $\vtav[\UPindex]$ as the \emph{individual uplifting index} of the pair $(\arm,\indv)$. It serves as an estimate of the individual uplift $\vav[\meanUplift]$.
Our algorithm, \UpUCBL, leverages two important procedures to compute an optimistic estimate of the uplift 
$\va[\up{\reward}]$: identification of affected variables, and padding with variables with the highest individual uplifting indices. 

To begin,  in line 6, \UpUCBL constructs the set of \emph{identified variables} 
%
\begin{equation}
    \notag
    \vta[\setId] = \setdef{\indv\in\variables}{\vav[\meanReward][0]\notin\vtav[\intConf]}.
\end{equation}
This set is contained in $\va[\variables]$ with high probability.
In fact, by concentration of measure, with high probability $\vav[\meanReward]\in\vtav[\intConf]$, in which case $\vav[\meanReward][0]\notin\vtav[\intConf]$ indicates $\indv\in\va[\variables]$.
However, $\vta[\setId]$ is not guaranteed to capture all the affected variables, so
we also need to provide an upper bound for $\sum_{\indv\in\setexclude{\va[\variables]}{\vta[\setId]}}\vav[\meanUplift]$, the uplift contributed by the unidentified affected variables.
Since the individual uplifting index $\vtav[\UPindex]$ is in fact a \ac{UCB} on the individual uplift $\vav[\meanUplift]$ here and $\card(\va[\variables])\le\nAffected$, we can simply choose the $\nAffected-\card(\vta[\setId])$ variables in $\setexclude{\variables}{\vta[\setId]}$ with the largest individual uplifting index $\vtav[\UPindex]$, as done in line 8 of \cref{algo:UpUCB-L-BL}.
We refer to this set as $\vta[\setLargeUp]$.
We then get a proper \ac{UCB} on the uplift of action $\arm$ by computing $\vta[\UPindexSum]=\sum_{\indv\in\vta[\setId]\union\vta[\setLargeUp]}\vtav[\UPindex]$ in line 9.
This process is summarized in \cref{algo:UpUCB-L-BL} and illustrated in \cref{subfig:UpUCBL}.

\subsubsection{Unknown Baseline Payoffs}
Now we focus on the most challenging setting, where also the baseline payoffs are unknown.
In this case, neither the sets of identified variables nor the uplifting indices of \UpUCBL can be defined.
We also cannot estimate the baseline payoffs using \eqref{eq:mean-estimate-0} since the sets of affected variables are unknown.
To overcome these challenges, we note that for any two actions $\arm,\armalt\in\arms$, $\va[\meanReward][\arm]$ and $\va[\meanReward][\armalt]$ only differ on $\va[\variables]\union\va[\variables][\armalt]$, and 
$\card(\va[\variables]\union\va[\variables][\armalt])\le2\nAffected$.
Therefore, $\va[\meanReward][\arm]$ and $\va[\meanReward][\armalt]$ differ in at most $2\nAffected$ variables, and we recover a similar problem structure by taking the payoffs of any action as the baseline.

Combining this idea with the elements that we have introduced previously,
we obtain \UpUCBLwb (\cref{algo:UpUCB-L}).
In each round, \UpUCBLwb starts by picking a most frequently taken action $\vt[\mostpullarm]$ (Line 4) whose payoffs are treated as the baseline during this round.
Then, in Line 6, \UpUCBLwb chooses variables that are guaranteed to be either in $\va[\variables]$ or $\va[\variables][\vt[\mostpullarm]]$. 
This generalizes the identification step of \UpUCBL.
The individual uplifting indices $\vtav[\UPindex]=\vtav[\UCBindex]-\vtav[\UCBindex][\run][\vt[\mostpullarm]]$ are computed in Line 7. The definition as differences of \ac{UCB}s is inspired by similar construction in \UpUCBwb.
Lines 8 and 9 constitute the padding step, during which variables with the highest uplifting indices are selected, and finally in Line 10 we combine the above to get the uplifting index of the action.
%
%
%

The similarity between \UpUCBLwb and \UpUCBL can be shown by supposing that one action has been taken frequently. Then the baseline payoffs are precisely estimated and do not change much between consecutive rounds.

\subsubsection{Regret Bounds}
Both \UpUCBL and \UpUCBLwb choose $\bigoh(\nAffected)$ variables for estimating the uplift of an action, and the decisions are made based on these estimates.
Therefore, the statistical efficiencies of these algorithms only scale with $\nAffected$ and not $\nVariables$.
This in turn translates into an improvement of the regret, as demonstrated by the theorem below.



\begin{algorithm}[tb]
    \caption{\UpUCBL} 
    \label{algo:UpUCB-L-BL}
\begin{algorithmic}[1]
    \STATE {\bfseries Input:} Error probability $\alt{\smallproba}$, Baseline payoffs $\va[\meanReward][0]$, Upper bound $\nAffected$ on the number of affected variables
    \STATE {\bfseries Initialization:} Take each action once
    \FOR{$\run = \nArms+1, \ldots, \nRuns$}
    \FOR{$\arm\in\arms$}
        \STATE Compute UCBs, uplifting indices,
        and confidence intervals 
        following \eqref{eq:UCBs-arm}
        \STATE Set $\vta[\setId] \subs \setdef{\indv\in\variables}{\vav[\meanReward][0]\notin\vtav[\intConf]}$
        \STATE Set $\vta[{\nAffected}] \subs \max(0, \nAffected-\card(\vta[\setId]))$
        \STATE Set $\displaystyle \vta[\setLargeUp] \subs
            \argmax_{\substack{\setLargeUp\subseteq\setexclude{\variables}{\vta[\setId]}\\
            \card(\setLargeUp)=\vta[{\nAffected}]}}
            \sum_{\indv\in\setLargeUp}\vtav[\UPindex]$
        \STATE Compute uplifting index $\vta[\UPindexSum]\subs\sum_{\indv\in\vta[\setId]\union\vta[\setLargeUp]}\vtav[\UPindex]$ 
    \ENDFOR
    \STATE Select action $\vt[\arm]\in\argmax_{\arm\in\arms}\vta[\UPindexSum]$
    \ENDFOR
\end{algorithmic}
\end{algorithm}
\begin{algorithm}[tb]
    \caption{\UpUCBLwb} 
    \label{algo:UpUCB-L}
\begin{algorithmic}[1]
    \STATE {\bfseries Input:} Error probability $\alt{\smallproba}$, Upper bound $\nAffected$ on the number of variables that each action affects 
    \STATE {\bfseries Initialization:} Take each action once
    \FOR{$\run = \nArms+1, \ldots, \nRuns$}
    \STATE Choose $\vt[\mostpullarm]\in\argmax_{\arm\in\arms}\vta[\pullcount][\run-1]$
    and compute UCBs and confidence intervals using \eqref{eq:UCBs-arm}
    \FOR{$\arm\in\arms$}
        \STATE Set $\vta[\setId]\subs\setdef{\indv\in\variables}
        {\vtav[\intConf]\intersect\vtav[\intConf][\run][\vt[\mostpullarm]]=\varnothing}$
        \STATE For $\indv\in\variables$, compute $\vtav[\UPindex]\subs\vtav[\UCBindex]-\vtav[\UCBindex][\run][\vt[\mostpullarm]]$
        \STATE Set $\vta[{\nAffected}] \subs \max(0, 2\nAffected-\card(\vta[\setId]))$
        \STATE Set
        $\displaystyle \vta[\setLargeUp] \subs
        \argmax_{\substack{\setLargeUp\subseteq\setexclude{\variables}{\vta[\setId]},\\
        \card(\setLargeUp)\le\vta[{\nAffected}]}}
        \sum_{\indv\in\setLargeUp}\vtav[\UPindex]$
         \STATE Compute uplifting index $\vta[\UPindexSum]\subs\sum_{\indv\in\vta[\setId]\union\vta[\setLargeUp]}\vtav[\UPindex]$
    \ENDFOR
    \STATE Select action $\vt[\arm]\in\argmax_{\arm\in\arms}\vta[\UPindexSum]$
    \ENDFOR
\end{algorithmic}
\end{algorithm}


\vspace{0.4em}
\begin{restatable}{theorem}{UpUCBLWBRegret}
\label{thm:UpUCB-L-regret}
Let $\alt{\smallproba}=\smallproba/(2\nArms\nVariables\nRuns)$. Then the regret of \UpUCBL (\cref{algo:UpUCB-L-BL}) (resp.\ \UpUCBLwb, \cref{algo:UpUCB-L}), with probability at least $1-\smallproba$, satisfies:
%
\begin{equation}
    \label{eq:UpUCB-L-regret-main}
    \vt[\reg][\nRuns] \le
    \sum_{\arm\in\arms:\va[\gapbandit]>0}
    \left(\frac{\alpha\nAffected^2\log(2\nArms\nVariables\nRuns/\smallproba)}{\va[\gapbandit]}+\va[\gapbandit]\right),
\end{equation}
where $\alpha=32$ (resp.\ $512$) in the above inequality.
\end{restatable}
\vspace{-0.3em}

The proof of \cref{thm:UpUCB-L-regret} (\cref{apx:upper-bound-unknown-affected}) is notable for two reasons.
First, tracking of the identified variables guarantees that the uplifting index $\vta[\UPindexSum]$ does not overestimate the uplift $\va[\uplift]$ much.
Take \UpUCBL as an example.
An alternative to constructing a \ac{UCB} on $\va[\uplift]$ is to choose the $\nAffected$ variables with the highest individual uplifting indices $\vtav[\UPindex]$.
However, this would result in a severe overestimate when a negative individual uplift is present.
Second, to prove \eqref{eq:UpUCB-L-regret-main} for \UpUCBLwb, we use that the widths of confidence intervals of the chosen $\vt[\mostpullarm]$ are always smaller than those of the taken action. This is ensured by taking $\vt[\mostpullarm]$ as the most frequent action (Line 4 in \cref{algo:UpUCB-L}). 

\subsection{Known Lower Bound on Individual Uplift}
\label{sec:part2}
In this second part of this section, we assume a lower bound on individual uplift is known.
This means the learner has access to $\gapUplift>0$ satisfying that for all $\arm\in\arms$ and $\indv\in\va[\variables]$, $\abs{\vav[\meanReward]-\vav[\meanReward][0]}\ge\gapUplift$.
This gives us an indicator on how many times we need to take each action in order to identify all the affected variables. Similar assumption was made by \citet{LMT21} to identify which node in the causal graph affects the reward variable.

Again in this scenario, we derive regret bounds for both when the baseline payoffs are known or not to the learner. 
Our algorithms combine UCB, successive elimination \citep{even2006action}, and the idea that the affected variables can be identified after an action is taken sufficiently many times.
Our regret bounds feature the ratio $\va[\gapbandit]/\gapUplift$, suggesting that this knowledge is helpful when $\gapUplift$ is relatively large. 

Hereinafter, for any $x,\alpha,\beta\in\R$ with $\alpha\le\beta$, we use $\clip(x,\alpha,\beta)$ to denote the clipping function that restricts $x$ to the interval $[\alpha,\beta]$, \ie $\clip(x,\alpha,\beta)=\max(\alpha,\min(\beta,x))$. Missing proofs are collected in~\cref{apx:lower-bound-unknown-affected}.

\subsubsection{Known Baseline Payoffs}

As we have seen in \cref{subsubsec:UpUCB-L-BL},
if the baseline payoffs $\va[\meanReward][0]$ are known, we can directly check whether $\vav[\meanReward][0]$ is in the confidence interval.
Moreover, with the knowledge $\abs{\vav[\meanReward]-\vav[\meanReward][0]}\ge\gapUplift$ for all $\indv\in\va[\variables]$, we deduce that $\va[\variables]$ can be identified correctly with probability $1-2\nVariables\alt{\smallproba}$ after action $\arm$ is taken $\pullcount_0=\ceil{(8/\gapUplift^2)\log(1/\alt{\smallproba})}$ times.
In fact, let $\vta[\valuev][1],\ldots,\vta[\valuev][\pullcount_0]$ be $\pullcount_0$ i.i.d. realizations of $\va[\valuev]\sim\va[\distribution]$ and $\vtav[\tilde{\meanReward}][\pullcount_0]=(\sum_{\run=1}^{\pullcount_0}\vtav[\valuev])/\pullcount_0$ be the empirical mean of variable $\indv$.
Applying the sub-Gaussian concentration inequality (\ref{eq:sub-Gaussian-concentration}, \cref{apx:concentration}) gives
\begin{equation}
    \prob\left(\abs{\vtav[\tilde{\meanReward}][\pullcount_0]-\vav[\meanReward]}\ge\frac{\gapUplift}{2}\right)
    \le2
    \exp\left(-\frac{\pullcount_0\gapUplift^2}{8}\right)
    \le 2\alt{\smallproba}.
\end{equation}
With a union bound we conclude that with probability at least $1-2\nVariables\alt{\smallproba}$ it holds for all $\indv$ that $\abs{\vtav[\tilde{\meanReward}][\pullcount_0]-\vav[\meanReward]}<\gapUplift/2$.
We distinguish between the following two situations
\begin{enumerate}
    \item $\indv\in\va[\variables]$, then $\displaystyle \abs{\vtav[\tilde{\meanReward}][\pullcount_0]-\vav[\meanReward][0]}\ge\abs{\vav[\meanReward]-\vav[\meanReward][0]}-\abs{\vtav[\tilde{\meanReward}][\pullcount_0]-\vav[\meanReward]}>\gapUplift-\frac{\gapUplift}{2}=\frac{\gapUplift}{2}$.
    \item $\indv\notin\va[\variables]$, then $\displaystyle \abs{\vtav[\tilde{\meanReward}][\pullcount_0]-\vav[\meanReward][0]}=\abs{\vtav[\tilde{\meanReward}][\pullcount_0]-\vav[\meanReward]}<\frac{\gapUplift}{2}$.
\end{enumerate}
This implies that with the choice
$\va[\setId] = \setdef{\indv\in\variables}{\abs{\vtav[\tilde{\meanReward}][\pullcount_0]-\vav[\meanReward][0]}>\gapUplift/2}$, we have effectively $\prob(\va[\setId]=\va[\variables])\ge1-2\nVariables\alt{\smallproba}$.
One natural algorithm is to thus first pull each arm $\pullcount_0$ times and run \UpUCB (\cref{algo:UpUCB-BL}) 
in the remaining rounds by regarding $\va[\setId]$ as the variables that action $\arm$ affects.
A direct computation shows that the regret of this algorithm is in
$\bigoh(\sum_{\arm\in\arms:\va[\gapbandit]>0}(\va[\gapbandit]/\gapUplift+\va[\nAffected])^2\log\nRuns/\va[\gapbandit])$ with high probability.

However, $\pullcount_0$ can be arbitrarily large when $\gapUplift$ gets closer to $0$, and taking an action much fewer times is probably sufficient to deduce that action is not optimal (with high probability).
We thus propose to implement the above idea in a more flexible manner:
We run \UpUCB with the uplifting indices $\vta[\UPindexSum]=\sum_{\indv\in\vta[\setId]}(\vtav[\UCBindex]-\vav[\meanReward][0])$, where
\begin{equation}
    \notag
    \vta[\setId] =
    \begin{cases}
    \variables & \text{if} ~ \vta[\pullcount]<\pullcount_0,\\
    \setdef{\indv\in\variables}{\abs{\vtav[\est{\meanReward}][\run][\vt[\arm]]-\vav[\meanReward][0]}>\gapUplift/2} & \text{otherwise}.
    \end{cases}
\end{equation}
%
We refer to this algorithm as \UpUCBDelta, with iLift standing for individual uplift, and summarize it in \cref{algo:UpUCB-GapUp-BL}. \UpUCBDelta has the following regret guarantee.

\begin{algorithm}[tb]
    \caption{\UpUCBDelta} 
    \label{algo:UpUCB-GapUp-BL}
\begin{algorithmic}[1]
    \STATE {\bfseries Input:} Error probability $\alt{\smallproba}$, Baseline payoffs $\va[\meanReward][0]$, Lower bound on individual uplift $\gapUplift$
    \STATE {\bfseries Initialization:} Take each action once;
    Set $\pullcount_0\subs\ceil{(8/\gapUplift^2)\log(1/\alt{\smallproba})}$ and $\va[\setId]\subs\variables$
    \FOR{$\run = \nArms+1, \ldots, \nRuns$}
    \STATE Compute the UCB indices following \eqref{eq:UCBs-arm}
    \STATE For $\arm\in\arms$, compute uplifting index $\vta[\UPindexSum]\subs\sum_{\indv\in\va[\setId]}(\vtav[\UCBindex]-\vav[\meanReward][0])$
    \STATE Select action $\vt[\arm]\in\argmax_{\arm\in\arms}\vta[\UPindexSum]$
    \IF{$\vta[\pullcount][\run][\vt[\arm]]\ge\pullcount_0$}
        \STATE Set $\va[\setId][\vt[\arm]]\subs\setdef{\indv\in\variables}{\abs{\vtav[\est{\meanReward}][\run][\vt[\arm]]-\vav[\meanReward][0]}>\gapUplift/2}$
    \ENDIF
    \ENDFOR
\end{algorithmic}
\end{algorithm}

\begin{restatable}{theorem}{UpUCBGapUpBL}
\label{thm:UpUCB-GapUp-BL-regret}
Let $\alt{\smallproba}=\smallproba/(2\nArms\nVariables\nRuns)$. Then the regret of \UpUCBDelta (\cref{algo:UpUCB-GapUp-BL}), 
with probability at least $1-\smallproba$, satisfies:
\begin{equation}
    \label{eq:UpUCB-GapUp-BL-regret}
    \vt[\reg][\nRuns] \le
    \sum_{\arm\in\arms:\va[\gapbandit]>0}
    \left(\frac{8\thinspace\clip(\va[\gapbandit]/\gapUplift,\va[\nAffected],\nVariables)^2\log(2\nArms\nVariables\nRuns/\smallproba)}{\va[\gapbandit]}+\va[\gapbandit]\right).
\end{equation}
\end{restatable}

\cref{thm:UpUCB-GapUp-BL-regret} shows that \cref{algo:UpUCB-GapUp-BL} provides a smooth transition between standard UCB and the naive strategy that tries to identify the affected variables of all the actions.
When $\va[\gapbandit]/\gapUplift\ge\nVariables$, the loss of taking action $\arm$ is so large that UCB prevents it from being taken further even if $\vta[\pullcount]\le\pullcount_0$.
Otherwise, the affected variables get identified after an action is taken sufficient number of times and the algorithm benefits from this knowledge to improve the estimate of the uplift.
It is also quite straightforward to combine \cref{algo:UpUCB-L-BL} with \cref{algo:UpUCB-GapUp-BL} when both $\nAffected$ and $\gapUplift$ are known; this results in a regret bound that replaces $\clip(\va[\gapbandit]/\gapUplift,\va[\nAffected],\nVariables)$ by $\clip(\va[\gapbandit]/\gapUplift,\va[\nAffected],\nAffected)$ in \eqref{eq:UpUCB-GapUp-BL-regret}.

\subsubsection{Unknown Baseline Payoffs}
Several additional challenges emerge when we want to adapt the aforementioned strategies to the case where the baseline payoffs are unknown.
First, without additional assumption, it is impossible to tell whether two arbitrarily close estimates indicate the same effect of two actions on a variable.
Then, without a proper baseline to compare with, we can never exclude the possibility that all the variables are affected by all the actions.
To circumvent this issue, we make the following more stringent assumption.

\begin{assumption}
\label{asm:gapUplift-ubl}
The effect of any two actions on a variable differ by at least $\gapUplift$ as long as one of them affects this variable, \ie for all $\arm\in\arms$, $\indv\in\va[\variables]$, and $\alt{\arm}\in\setexclude{\arms}{\{\arm\}}$, it holds $\abs{\vav[\meanReward]-\vav[\meanReward][\alt{\arm}]}\ge\gapUplift$.
\end{assumption}

\begin{algorithm}[tb]
    \caption{\UpUCBDeltawb} 
    \label{algo:UpUCB-GapUp}
\begin{algorithmic}[1]
    \STATE {\bfseries Input:} Error probability $\alt{\smallproba}$, Lower bound on effect gap $\gapUplift$
    \vskip 0.1em
    \STATE {\bfseries Initialization:}
    Set $\run\subs0$,  $\vt[\arms][1]\subs\arms$, and $\pullcount_0\subs\ceil{(32/\gapUplift^2)\log(1/\alt{\smallproba})}$
    \STATE \texttt{\underline{Phase \rom{1}: Successive elimination}}
    \vskip 0.3em
    \FOR{$\round = 1, \ldots, \pullcount_0$}
    \IF{$\nRuns-\run\ge\card(\vt[\arms][\round])$}
    \STATE Take each action $\arm\in\vt[\arms][\round]$ once
    \STATE Set $\run\subs\run+\card(\vt[\arms][\round])$
    
    \STATE Compute the empirical estimates $\vta[\estalt{\reward}][\round]$ for $\arm\in\vt[\arms][\round]$ and the radius of confidence interval $\vt[\radiusConf][\round]$
    \STATE Update active set
    $\vt[\arms][\round+1]\subs\setdef{\arm\in\vt[\arms][\round]}{
    \vta[\estalt{\reward}][\round]+2\vt[\radiusConf][\round]\ge
    \max_{\armalt}\vta[\estalt{\reward}][\round][\armalt] }$
    \ELSE
    \STATE Take $\nRuns-\run$ actions from $\vt[\arms][\round]$ and terminate the algorithm
    \ENDIF
    \ENDFOR
    \vskip 0.1em
    \STATE \texttt{\underline{Phase \rom{2}: UpUCB}}
    \vskip 0.3em
    \STATE Construct the sets $\vv[\setBaseEst]$ as in \eqref{eq:setBaseEst} and $\va[\setId]\subs\setdef{\indv\in\variables}{\arm\notin\vv[\setBaseEst]}$
    \FOR{$\run = \run+1, \ldots, \nRuns$}
    \STATE Compute the UCB indices following \eqref{eq:UCBs-arm},\eqref{eq:UCBs-0-UpUCB-GapUp}
    \STATE For $\arm\in\vt[\arms][\pullcount_0+1]$, compute uplifting index $\vta[\UPindexSum]\subs\sum_{\indv\in\va[\setId]}(\vtav[\UCBindex]-\vtav[\UCBindex][\run][0])$
    \STATE Select action $\vt[\arm]\in\argmax_{\arm\in\vt[\arms][\round]}\vta[\UPindexSum]$
    \ENDFOR
\end{algorithmic}
\end{algorithm}

In other words, $\gapUplift$ is not only a lower bound on individual uplift, but also a lower bound on (individual) effect gap.
Second, with a UCB-style algorithm, the optimal actions may not be taken sufficiently many times, in which case we can only conservatively assume that they affect all the variables.
This would incur additional regret in the analysis when $\va[\meanReward][0]$ is unknown. 
Therefore, we would like to ensure that all the (potentially optimal) actions are taken the same number of times at the point that the affected variables get identified.
This leads to a two-stage method as described in \cref{algo:UpUCB-GapUp}.

In the first stage, we perform successive elimination until the uneliminated actions are taken sufficiently many times.
Here $\vta[\estalt{\reward}][\round]$ is the empirical estimate of the reward associated to action $\arm$ and $\vt[\radiusConf][\round]=\nVariables\sqrt{2\log(1/\alt{\smallproba})/\round}$ is the associated radius of confidence interval.
Subsequently, for each variable $\indv\in\variables$ we construct the set
\begin{equation}
    \label{eq:setBaseEst}
    \vv[\setBaseEst]
    = \setdef{\arm\in\arms}{\exists\thinspace \armalt\in\setexclude{\arms}{\{\arm\}}, \vtav[\intConf][\run_0+1]\intersect\vtav[\intConf][\run_0+1][\armalt]\neq\varnothing},
\end{equation}
where $\run_0$ is the number of iterations in the elimination phase and the confidence intervals are defined as in \eqref{eq:UCBs-arm}.
If \cref{asm:gapUplift-ubl} is verified, then with high probability $\arm\in\vv[\setBaseEst]$ only if $\indv\notin\va[\variables]$.
Therefore, the observed results from these arms can be used to estimate the baseline payoff of $\indv$.
On the other hand, the identified affected variables for action $\arm\in\arms$ are then $\va[\setId]=\setdef{\indv\in\variables}{\arm\notin\vv[\setBaseEst]}$.

The second stage of the algorithm reuses the idea from \UpUCBwb.
We define the UCB indices following \eqref{eq:UCBs-arm} for $\arm\in\arms$ and for the baseline estimate we choose among $\vv[\setBaseEst]$ the action that is taken the most frequently whenever this set is non-empty. Otherwise we arbitrarily set it to $0$. That is,
\begin{equation}
    \label{eq:UCBs-0-UpUCB-GapUp}
    \vtav[\UCBindex][\run][0] = \begin{cases}
    0
    & \text{if} ~ \vv[\setBaseEst]=\varnothing,
    \\
    \vtav[\UCBindex][\run][\vtav[\arm][\run][0]] \text{ where $\vtav[\arm][\run][0]=\argmax_{\arm\in\vv[\setBaseEst]}\vta[\pullcount][\run-1]$} & \text {otherwise}.
    \end{cases}
\end{equation}
The uplifting indices that guide the decision of the algorithm are defined by $\vta[\UPindexSum]=\sum_{\indv\in\vta[\setId]}(\vtav[\UCBindex]-\vtav[\UCBindex][\run][0])$.

As shown below, the regret of \cref{algo:UpUCB-GapUp} takes roughly the same form as the one in \cref{thm:UpUCB-GapUp-BL-regret}, but as in \cref{thm:UpUCB-regret}, the number of variables that an optimal action affects also plays a crucial role in the regret bound.



\begin{restatable}{theorem}{UpUCBGapUpregret}
\label{thm:UpUCB-GapUp-regret}
Let \cref{asm:gapUplift-ubl} hold and $\alt{\smallproba}=\smallproba/(2\nArms\nVariables\nRuns)$.  Then the regret of \UpUCBDeltawb (\cref{algo:UpUCB-GapUp}), with probability at least $1-\smallproba$, satisfies:
\begin{equation}
    \label{eq:UpUCB-GapUp-regret}
    \vt[\reg][\nRuns] \le
    \sum_{\arm\in\arms:\va[\gapbandit]>0}
    \left(\frac{8\thinspace\clip(2\va[\gapbandit]/\gapUplift,\va[\nAffected]+\va[\nAffected][\sol[\arm]],2\nVariables)^2\log(2\nArms\nVariables\nRuns/\smallproba)}{\va[\gapbandit]}+\va[\gapbandit]\right).
\end{equation}
\end{restatable}

\vspace{0.4em}
\begin{remark}
A natural candidate for $\vtav[\UCBindex][\run][0]$ is the one that uses aggregated information as in \eqref{eq:mean-estimate-0}.
The problem of this choice is that how we aggregate the observations collected in the first phase are only determined at the end of the the phase, while this choice itself depends on these observations.
Mathematically, this prevents us from defining a suitable martingale difference sequence for which we establish concentration bound.
An easy fix is to only aggregate observations from the second phase, while the use of the observations from the first phase should be limited to those coming from just one arm.
\end{remark}

\begin{figure*}[t!]
    \centering
    \vspace{-2em}
    \begin{subfigure}{0.45\linewidth}
    \includegraphics[width=\linewidth]{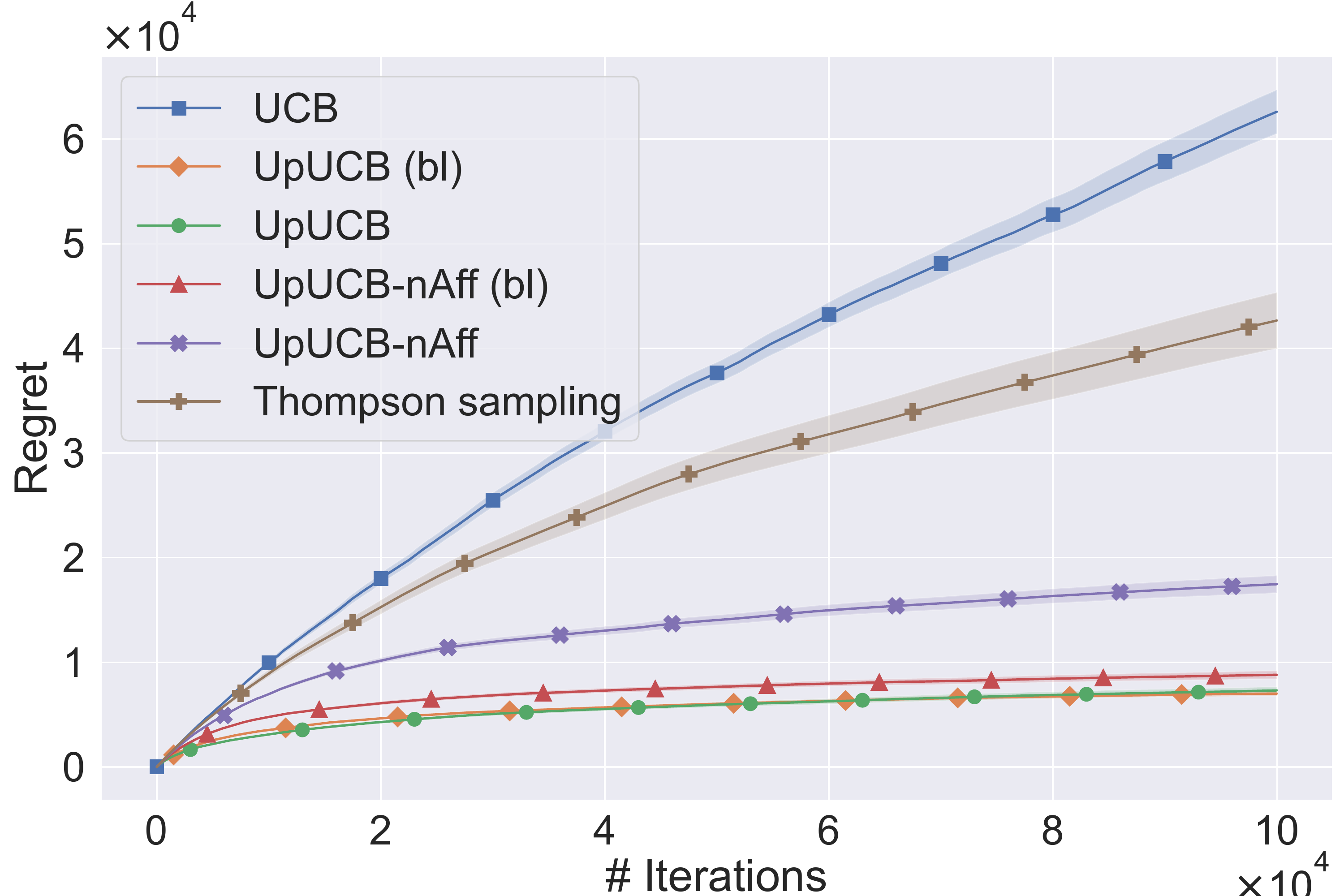}
    \caption{Gaussian uplifting bandit}
    \label{subfig:exp-gau}
    \end{subfigure}
    \hspace{1em}
    \begin{subfigure}{0.45\linewidth}
    \includegraphics[width=\linewidth]{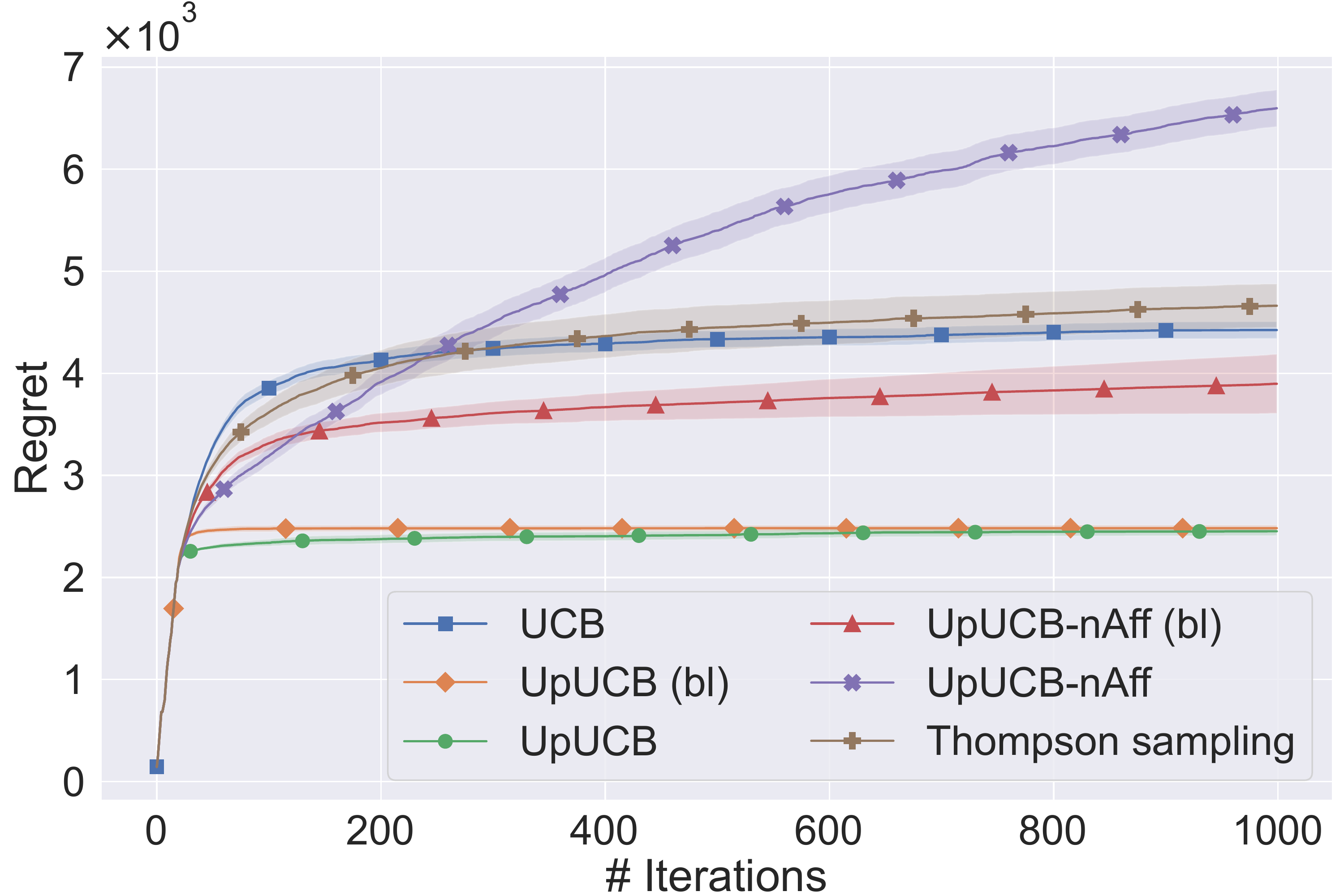}
    \caption{Bernoulli uplifting bandit}
    \label{subfig:exp-ber}
    \end{subfigure}
    \captionsetup{labelsep=space}
    \caption{Experimental results on a synthetic and real-world dataset.
    All the curves are averaged over $100$ runs and the shaded areas represent the standard errors.}
    \label{fig:exp-main}
\end{figure*}
\section{Numerical Experiments}
\label{sec:exp}




In this section, we present numerical experiments to demonstrate the benefit of estimating uplifts in our model.
We compare
\UpUCB, \UpUCBwb, \UpUCBL, and \UpUCBLwb
(\cref{algo:UpUCB-BL,algo:UpUCB,algo:UpUCB-L-BL,algo:UpUCB-L})
against UCB and Thompson sampling with Gaussian prior and Gaussian noise that only use the observed rewards $(\vt[\reward])_{\run\in\oneto{\nRuns}}$.
To ensure a fair comparison, we tune all the considered algorithms and report the results for the parameters that yield the best averaged performance.
The experimental setup details are provided in \cref{apx:exp-detail}. The experiments for the contextual combinatorial uplifting model introduced in~\cref{subsec:C2UpBandit} are presented in \cref{apx:exp-c2}.


\paragraph{Gaussian Uplifting Bandit\afterhead}
\label{subsec:exp-gaussian}
We first study our algorithms 
in a Gaussian uplifting bandit,
\ie the noise vectors $(\va[\snoise])_{\arm\in\arms}$ are Gaussian,
with $\nArms = 10$ actions, $\nVariables = 100$ variables, and  $\va[\nAffected] \equiv 10$ meaning that each action affects $10$ variables.
The expected payoffs are contained in $[0,1]$,
and the covariance matrix of the noise is taken the same for all the actions.
The suboptimality gap of the created problem is around $0.2$, and the variance of the total noise $\sum_{\allvariables}\vav[\snoise]$ is around $80$.

\paragraph{Bernoulli Uplifting Bandit with Criteo Uplift\afterhead}
\label{subsec:exp-ber}
We use the Criteo Uplift Prediction Dataset~\citep{Diemert2018} 
with `visit' as the outcome variable to build a Bernoulli uplifting bandit, where the payoff of each variable has a Bernoulli distribution.
This dataset is designed for uplift modeling, and has outcomes for both treated and untreated individuals.
Thus it is suitable for our simulations.
To build the model, we sample $10^5$ examples from the dataset, and use K-means to partition these samples into $20$ clusters of various sizes.
The $10^5$ examples are taken as our variables.
We consider $20$ actions that correspond to treating individuals of each cluster.
Assuming that each cluster contains a single type of users that react in the same pattern,
we sample the payoffs $(\vtv[\valuev])_{\allvariables}$ from independent Bernoulli distributions with means that equal the average visit rates of the treated or untreated individuals of the clusters, depending on which action is chosen.
Here, $\nAffected=12654$ and $\gapbandit$ is around~$30$.

\paragraph{Results\afterhead}
\cref{fig:exp-main} confirms that we can effectively achieve much smaller regret by restricting our attention to the uplift.
Moreover, when the sets of affected are known, the loss of not knowing the baseline payoffs seems to be minimal.
Not knowing the affected variables has a more severe effect, and in particular causes much larger regret in the second experiment.
In fact, the design of \UpUCBLwb and \UpUCBL heavily rely on the additive structure of the uplifting index, and can thus hardly benefit from the payoff independence which allows the other four algorithms to achieve smaller regret in this case.

\begin{figure*}[t]
    \centering
    \begin{subfigure}{0.49\linewidth}
    \includegraphics[width=0.49\linewidth]{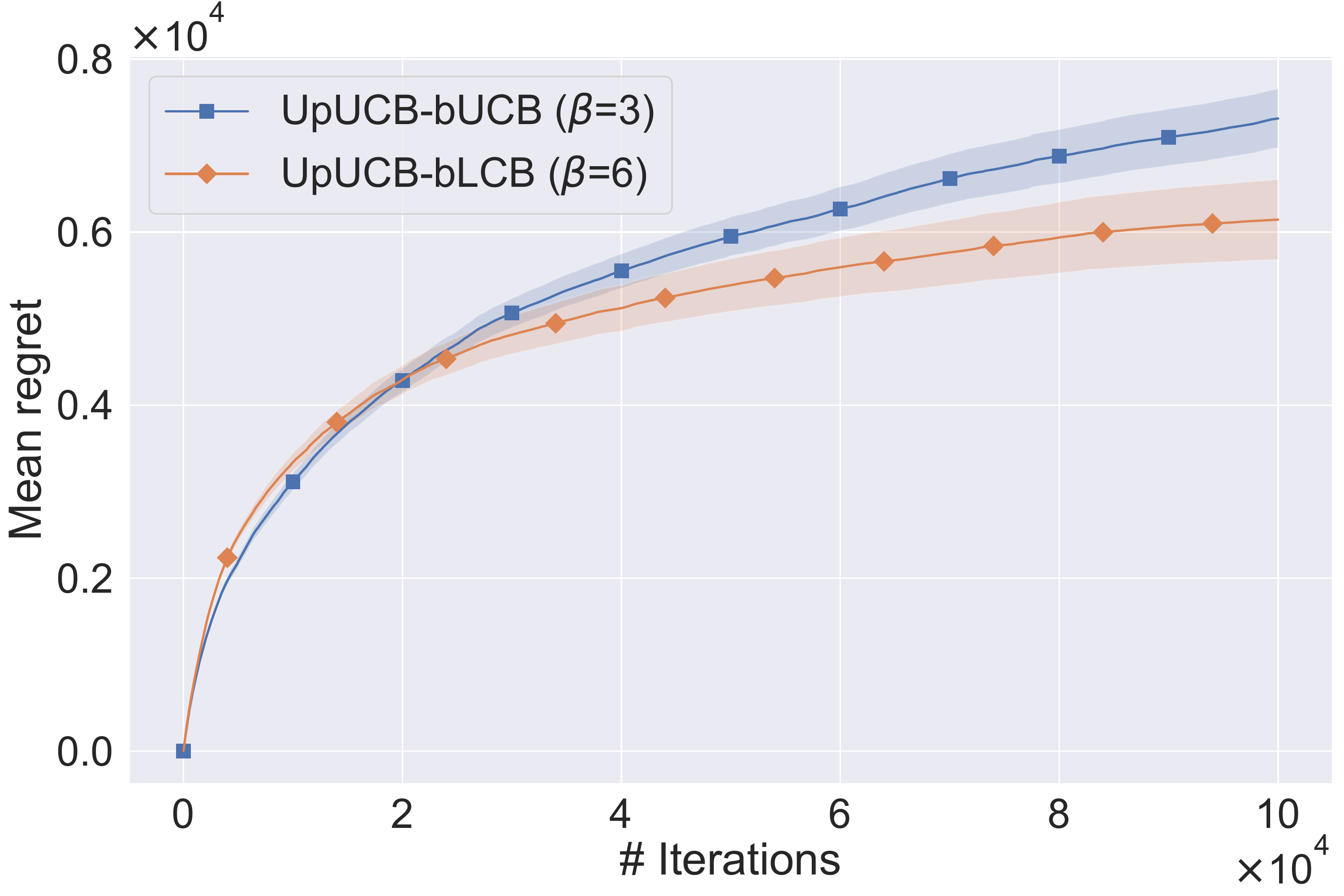}
    \includegraphics[width=0.49\linewidth]{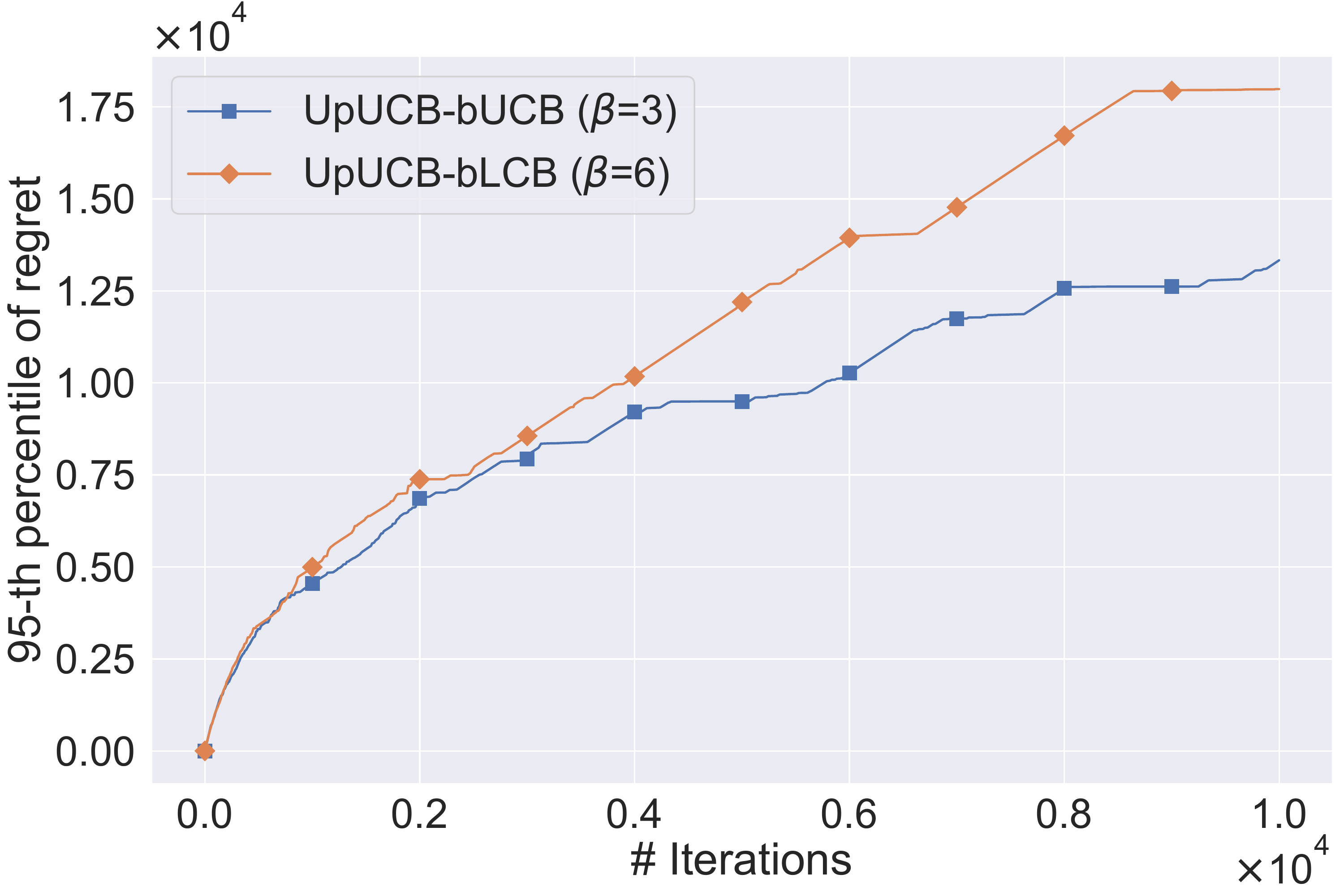}
    \caption{Baseline UCB versus Baseline LCB}
    \label{subfig:UCB-LCB}
    \end{subfigure}
    \begin{subfigure}{0.49\linewidth}
    \includegraphics[width=0.49\linewidth]{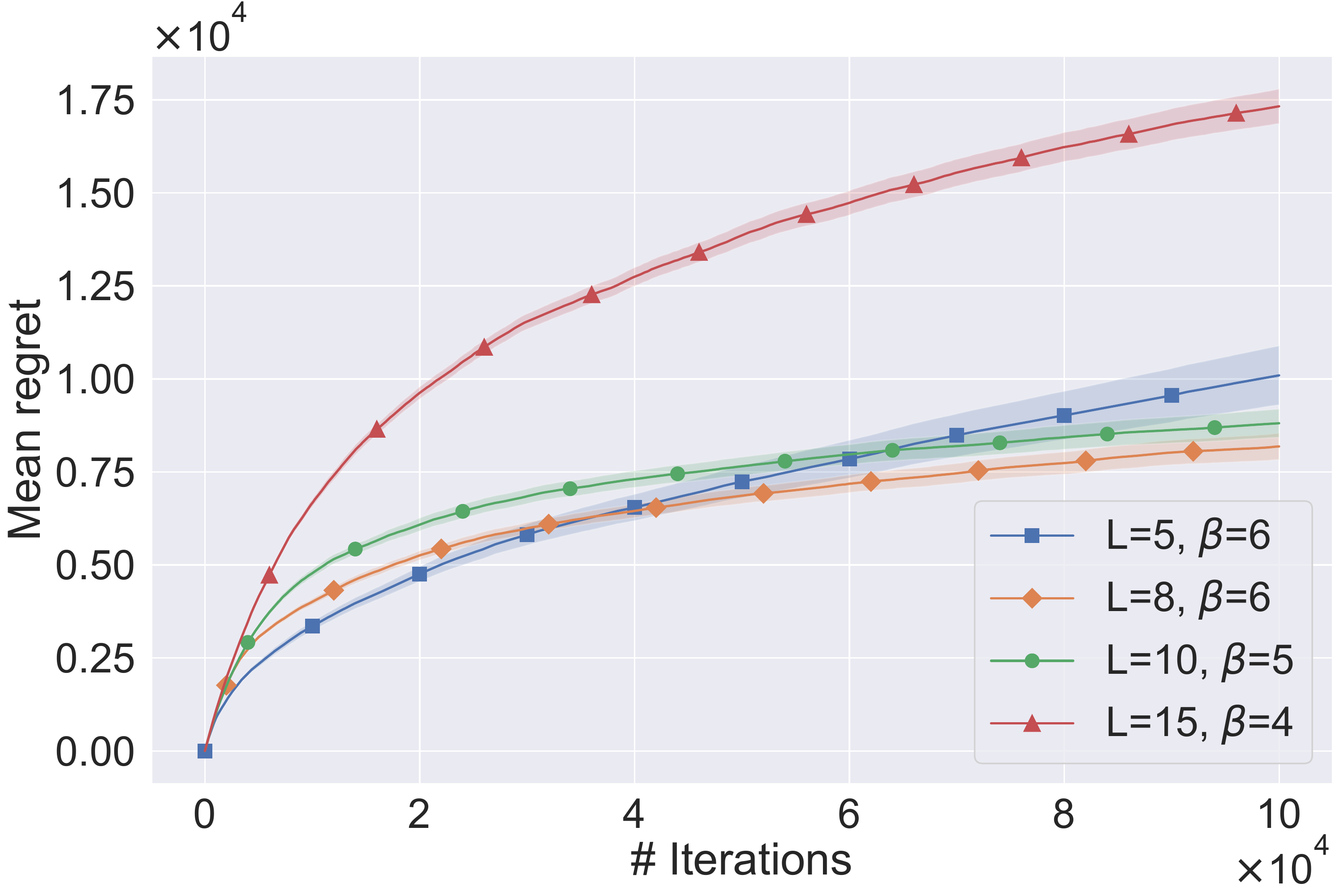}
    \includegraphics[width=0.49\linewidth]{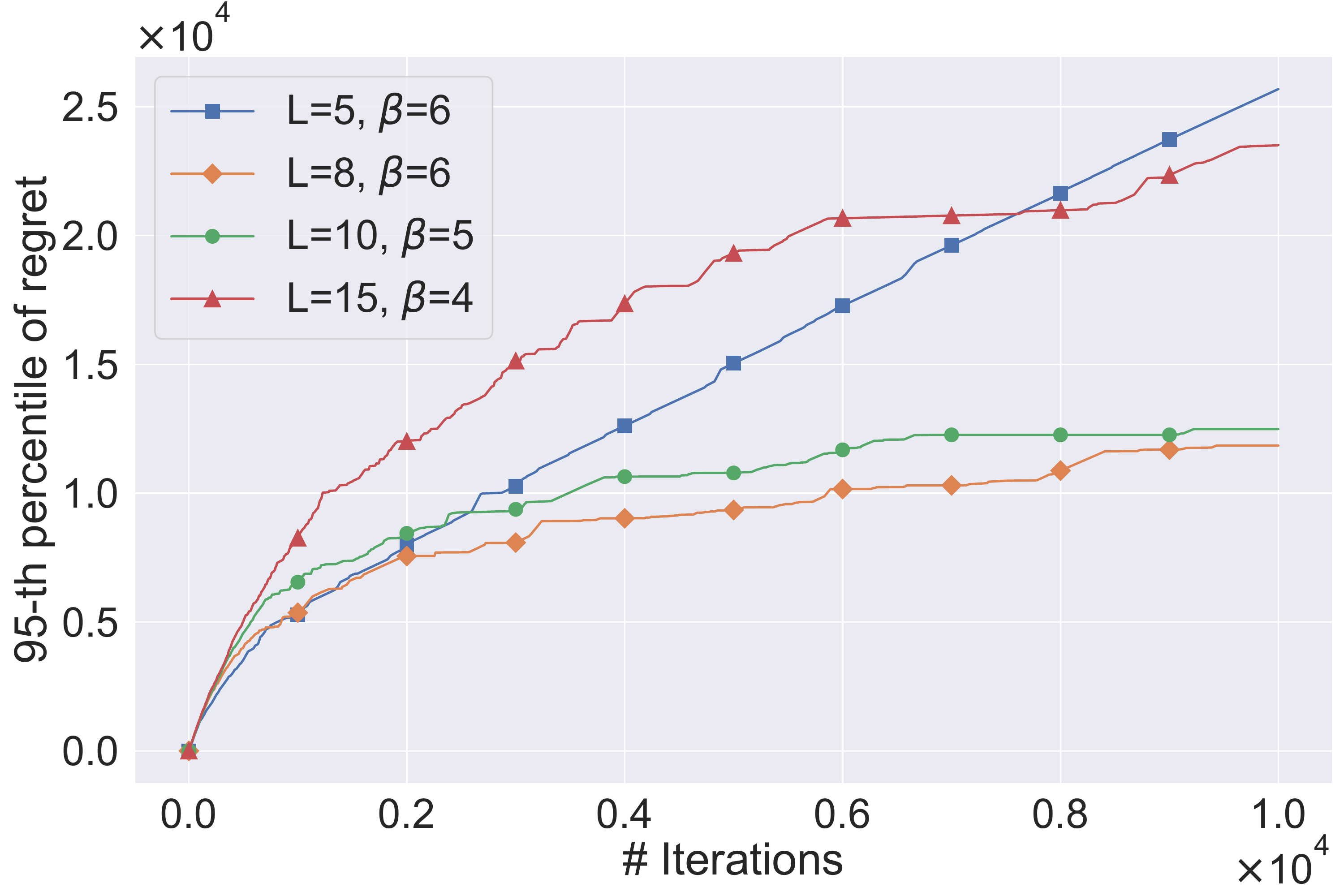}
    \caption{Consequence of misspecification of $\nAffected$}
    \label{subfig:misL}
    \end{subfigure}
    \caption{Ablation study on Gaussian uplifting bandit to investigate two aspects of our algorithms. Besides mean regret with standard error, we also plot the $95$-th percentile of the regrets as an indicator for the robustness of a method.}
    \vspace{-1em}
\end{figure*}
\subsection{Ablation Study on Gaussian Uplifting Bandit}
 \label{sec:abalation}

Below we further present ablation studies on  the Gaussian uplifting bandit model.

\paragraph{Baseline UCB versus Baseline LCB\afterhead}
In \cref{algo:UpUCB}, we suggest subtracting the UCBs of the baseline payoffs from the UCBs of the actions.
This could be counter-intuitive because to get an optimistic estimate of the uplift, we should instead subtract the LCBs of the baseline payoffs.
However, the latter strategy prevents the learner from learning more about the baselines that are badly estimated, and can thus lead to failure more frequently.
In the following, we refer to the two strategies respectively as \UpUCB-bUCB and \UpUCB-bLCB.

In \cref{subfig:UCB-LCB} (Left), we illustrate the mean regret of the two designs when their parameters are optimally tuned.
It turns out there the two strategies perform similarly here.
We would just like to highlight that to achieve the optimal performance, \UpUCB-bLCB needs to use a larger exploration parameter.
Therefore, in terms of average performance, it seems that the effect of using LCBs of the baseline is close to 
a shrinkage of the exploration parameter.

Next, we further inspect the robustness of the algorithms by plotting the $95$th percentile of the regrets (which roughly corresponds to the $5$th largest regret among those recorded in the $100$ runs).
The same parameters give the optimal values for this metric.
However, as shown in \cref{subfig:UCB-LCB} (Right), it turns out that \UpUCB-bLCB achieves a much higher $95$th percentile of regret, which indicates that it indeed fails more frequently.

\paragraph{Misspecification of $\nAffected$.}
We also investigate the effect of misspecification of $\nAffected$ on \cref{algo:UpUCB-L-BL}.
For the bandit instance that we employ here, $\max_{\arm\in\arms}\va[\nAffected]=10$ and thus according to \cref{prop:UpUCB-L-BL-regret}, \cref{algo:UpUCB-L-BL} guarantees sublinear regret as long as it is provided with $\nAffected\ge10$. On the flip side, the algorithm may incur linear regret if $\nAffected$ is underestimated.

To verify this, we run the algorithm with $\nAffected\in\{5,8,10,15\}$ and plot the averaged regret and the $95$-th percentile of regret for optimally tuned exploration parameters in \cref{subfig:misL}.
We effectively observe sublinear regret when $\nAffected$ is overspecified, but the regret becomes much larger because the algorithm is more conservative and explores more.
On the other hand, under severe underspecification of $\nAffected$, the algorithm is doomed to fail whatever the exploration parameter used.
Finally, it seems that slight underspecification of $\nAffected$ can be compensated by choosing a larger exploration parameter. In fact, in the figure we see that for $\nAffected=8$ choosing $\expparam=6$ gives the smallest regret among the presented curves.

\section{Contextual Extensions}
\label{sec:context}

In this section, we briefly discuss potential contextual extensions of our model.
As in contextual bandits, 
context is a side information that helps the learner to make a more informed decision, which results in a higher reward. To incorporate context, one possibility is to associate each variable with a feature vector $\vtv[\feature]\in\R^\vdim$.
The subscript $\run$ indicates that the context can change from one round to another.
We also associate each action with a function $\va[\obj]$ so that the expected payoff of action $\arm$ acting on a variable with feature $\vtv[\feature]$ is $\va[\obj](\vtv[\feature])$.
The expected reward of choosing $\arm$ at round $\run$ is then $\va[\reward](\vt[\feature])=\sum_{\allvariables}\va[\obj](\vtv[\feature])$.
The optimal action of round $\run$ is $\vt[\sol[\arm]]=\argmax_{\arm\in\arms}\va[\reward](\vt[\feature])$ and the regret of a learner that takes the actions $(\vt[\arm])_{\run\in\oneto{\nRuns}}$ is given by
\[
\vt[\reg][\nRuns]
    = \sum_{\run=1}^{\nRuns}\sum_{\indv\in\variables}
    (\va[\obj][\vt[\sol[\arm]]](\vtv[\feature]) - \va[\obj][\vt[\arm]](\vtv[\feature])).
\]
As in \cref{sec:model}, upon taking an action, the learner observes the noisy payoffs of all the variables.
The key structure in our model is there exists a baseline payoff vector $\va[\meanReward][0]$ such that for any given action $\arm$, the equality $\vav[\meanReward]=\vav[\meanReward][0]$ holds for most $\indv\in\variables$.
Given context, this can be translated into the existence of a baseline function $\va[\obj][0]$ such that for any given $\arm$ and $\run$, the equality $\va[\obj](\vtv[\feature])=\va[\obj][0](\vtv[\feature])$ holds for most $\indv\in\variables$.
The uplift of action $\arm$ is defined as \[\va[\up{\reward}](\vt[\feature])=\sum_{\allvariables}\va[\obj](\vtv[\feature])-\va[\obj][0](\vtv[\feature]).\]
We provide two concrete examples of the contextual extension below.

\medskip
\paragraph{Contextual Uplifting Bandit with Linear Payoffs\afterhead}
\label{subsec:contextual-linear}
In this model, each action is associated with an unknown parameter $\va[\param]$ and the expected payoff is the scalar product of $\va[\param]$ and the feature of the variable, \ie $\va[\obj](\vtv[\feature])=\product{\va[\param]}{\vtv[\feature]}$.
We also assume the aforementioned equality to hold for the baseline function $\va[\obj][0]$
and we use 
\[\vta[\variables]=\setdef{\indv\in\variables}{\product{\va[\param]}{\vtv[\feature]}\neq\product{\va[\param][0]}{\vtv[\feature]}}\]
for the variables affected by action $\arm$ at round $\run$.
One sufficient condition for $\vta[\variables]$ to be small is sparsity in both the parameter difference $\va[\up{\param}]=\va[\param]-\va[\param][0]$ and the context vector $\vtv[\feature]$.
In fact, $\product{\va[\param]}{\vtv[\feature]}=\product{\va[\param][0]}{\vtv[\feature]}$ as long as the supports of $\va[\up{\param}]$ and $\vtv[\feature]$ are disjoint.
In the sequel, we assume $\card(\vta[\variables])$ is uniformly bounded by $\nAffected$.

Clearly, our algorithms can be directly applied as long as we can construct a \ac{UCB} on $\product{\va[\param]}{\vtv[\feature]}$.
This can for example be done using the construction of Linear UCB~\citep{APS11}.
In this way, the decision of the learner is again based on the uncertain estimates of at most $\bigoh(\nAffected)$ variables, and we thus expect the improvements that we have shown in our theorems would persist.
As an example, when both $\va[\param][0]$ and $\vta[\variables]$ are known, \UpUCB adapted to this situation constructs \ac{UCB} for $\sum_{\indv\in\vta[\variables]}\product{\va[\up{\param}]}{\vtv[\feature]}$, and it is straightforward to show that the regret of such algorithm can be as small as $\bigoh(\nAffected\vdim\sqrt{\nArms\nRuns})$.\footnote{As in the previous sections, this bounds applies to the case where the noises are not necessarily independent.}
In contrast, if the learner works directly with the total reward, the regret is in $\bigoh(\nVariables\vdim\sqrt{\nArms\nRuns})$.

\paragraph{Contextual Combinatorial Uplifting Bandit\afterhead}
\label{subsec:C2UpBandit}

\begin{algorithm}[t]
    \caption{\CUpUCB} 
    \label{algo:C2UpUCB}
\begin{algorithmic}[1]
    \STATE {\bfseries Input:} Budget constraint $\nAffected$, Parameters for computing the UCBs indices
    \FOR{$\run = 1, \ldots, \nRuns$}
    \FOR{$\indv\in\variables$}
    \STATE Compute UCB indices $(\vtav[\UCBindex][\run][\indtreat])_{\indtreat\in\{0, 1\}}$
    \STATE Compute uplifting index $\vtv[\UPindex][\run]$
    \hfill
    \COMMENT{$\vtv[\UPindex][\run]=\vtav[\UCBindex][\run][1]-\vtav[\UCBindex][\run][0]$}~~
    \ENDFOR
    \STATE Set $\vt[\variables]\subs\argmax_{\set\subset\variables,\card(\set)\le\nAffected}\sum_{\indv\in\set}\vtv[\UPindex]$
    \STATE Select action $\vt[\arm]=(\one\{\indv\in\vt[\variables]\})_{\allvariables}$
    \hfill
    \COMMENT{this defines the set of treated customers}~~
    \ENDFOR
\end{algorithmic}
\end{algorithm}

To demonstrate the flexibility of our framework, we also consider a different model that tackles the problem of targeted campaign. 
Here, we refer to the variables as individuals (customers) that we may want to treat (serve) with some treatment (campaign). 
To improve the overall return on investment, it is often beneficial to treat only a subset of the individuals.
The bandit problem thus consists of finding the right set of individuals to treat at each round.
Concretely, if the learner selects at most $\nAffected$ individuals to treat in a round, 
the action set can be written as $\arms=\setdef{\arm\in\{0,1\}^{\nVariables}}{\norm{\arm}_0\le\nAffected}$, where $\norm{\arm}_0$ is the number of non-zero coordinates of $\arm$.
In this form, the $\indv$-th coordinate of $\arm$ is $1$ if and only if individual $\indv$ receives the treatment.
Similar to before, we observe the payoffs of all the individuals after treatments are applied.


Here, we make a practical {\em no interference} assumption, that the payoff of each individual only depends on whether they are treated or not (i.e., independent of others). 
We can thus define functions $\va[\objalt][\indtreat]$ for $\indtreat\in\{0,1\}$ so that $\va[\obj](\vtv[\feature])=\va[\objalt][\vv[\arm]](\vtv[\feature])$.\footnote{%
Strictly speaking, $\va[\obj]$ is a function of $(\vtv[\feature], \indv)$ here.}
Considering the action of not treating any individual as the baseline,
the uplift of action $\arm$ in round $\run$ is given by \[\va[\up{\reward}](\vt[\feature])=\sum_{\indv\in\va[\variables]}\va[\objalt][\vv[\arm]](\vtv[\feature])-\va[\objalt][0](\vtv[\feature]),\]
where $\va[\variables]$ is the set of individuals treated by action $\arm$.
For UCB-type methods, we should thus provide upper confidence bounds $\vtav[\UCBindex][\run][\indtreat]$ for $\va[\objalt][\indtreat](\vtv[\feature])$.
Depending on whether the baseline function is known or not, we may subtract $\va[\objalt][0](\vtv[\feature])$ or $\vtav[\UCBindex][\run][0]$ from $\vtav[\UCBindex][\run][1]$ to compute the uplifting indices for each individual.
The algorithm then chooses the $\nAffected$ individuals with the highest uplifting indices.
We refer to the resulting algorithm as \CUpUCB, where C2 is the abbreviation of ``contextual combinatorial''.
We provide its pseudo code in \cref{algo:C2UpUCB} (where we suppose the baseline is unknown).

In \cref{apx:C2UpUCB}, we further present a generalization of the algorithm to the case of multiple treatments, a theoretical analysis thereof for linear payoffs, and accompanying numerical experiments.

\section{Related Work}
\label{sec:related}
In this section, we link our work to other existing bandit models.
To begin, as suggested by the name, the concept of \emph{uplift} plays a crucial role in our approach.
From a high-level perspective, the goals of both uplifting modeling and \ac{MAB} is to help selecting the optimal action.
The former achieves it by modelling the incremental effect of an action on an individual's behavior.
Despite this apparent connection between the two concepts, few papers explicitly link them together.
We believe that this is because the resulting problem can often be solved by classic bandit algorithms with a redefined reward. This approach was taken in \citep{sawant2018contextual,berrevoets2019optimising}.

Instead, our paper focuses on bandit problems in which estimating the uplifts improves the statistical efficiency of the algorithms, and this is possible thanks to the `sparsity' of the actions' effects.
Prior to our work, sparsity assumptions in bandits primarily concerned the sparsity of the parameter vectors in linear bandits, 
which could be of great interest in high-dimensional settings \citep{abbasi2012online,bastani2020online,hao2021information}.
A notable exception is \citet{kwon2017sparse} who studied a variant of the \ac{MAB} where the sparsity is reflected by the fact that the number of arms with positive reward is small.
Our work is orthogonal to all of these in that we look at a different form of sparsity. As we saw in \cref{sec:context}, while sparsity in parameter can be a cause of sparsity in action effect, the improvement of regret is established with a different mechanism.
In the upcoming paragraphs, we further exhibit how our model is related to causal and combinatorial bandits.

\paragraph{Causal Bandits\afterhead}
The term causal bandits captures the general idea that in a bandit model the generation of reward may be governed by some causal mechanism while the actions correspond to interventions on the underlying causal graph.
This idea was first formalized in \citep{lattimore2016causal} and subsequently explored in a series of work \citep{lee2018structural,lu2020regret,nair2021budgeted} that cover multiples variants of the problem, including both the minimization of simple and cumulative regrets.
Our problem fits into the causal bandit framework by regarding \cref{fig:model} as a simple causal graph where the variables consist of all the directed causes of the reward and the actions correspond to atomic interventions on the action nodes.
The observability of these direct causes are also key assumptions in many of the aforementioned works.
Moreover, in the unknown affected case our causal graph is in fact partially unknown.
This is an intriguing problem in causal bandits as argued by \citep{LMT21}.
While we have shown improvement of the algorithms for this specific model, a interesting question is whether similar improvement can still be shown in more general causal bandit models under the assumption that each actions only affects a limited number of variables of the causal graph.

\paragraph{Combinatorial Bandits\afterhead}
In combinatorial bandits~\citep{cesa2012combinatorial}, the learner selects at each round a subset of ground items depending on which the reward is computed.
In its simplest form, the reward is just the sum of the payoffs of the selected items.
Of particular relevance to us is the semi-bandit setting~\citep{CWY13,KWAS15}, which assumes that the payoffs of the selected items are also observed.
This structure of the reward and observability of individual payoffs both suggest a strong similarity between our model and semi-bandits.
However, this comes with an important conceptual difference: the set of items, or variables, for which we observe the outcomes are selected in semi-bandits, while for us they are fixed and inherent to the reward generation mechanism.
As an example, in example \ref{ex:movie}, combinatorial bandit models optimize the number of movies that are watched among the recommended ones, while we also consider non-recommended movies.
In fact, as argued in \citep{WLCB20}, a recommendation is only effective if the user actually watches the movie and would not watch it without the recommendation

On the other hand, mathematically, it is possible to fit our model into the stochastic combinatorial semi-bandit framework when the sets of affected variables are known.
For this, consider a semi-bandit problem with $\nVariables+\sum_{\arm\in\arms}\va[\nAffected]$ items that correspond to the random variables $\setdef{\vav[\valuev][0]}{\indv\in\variables}\union\bigcup_{\arm\in\arms}\setdef{\vav[\valuev]}{\indv\in\va[\variables]}$.
At each round, the learner selects $\nVariables$ items that depends on the taken action and observes the payoffs of the selected items.
More precisely, when action $\arm$ is taken, the selected items are $\setdef{\vav[\valuev][0]}{\indv\in\setcomplement{\va[\variables]}}\union\setdef{\vav[\valuev]}{\indv\in\va[\variables]}$.

In terms of algorithm, let us consider the \UpUCBwb algorithm described in \cref{algo:UpUCB}, which is based on selecting the action with the largest uplift index $\vta[\UPindexSum]=\sum_{\indv\in\va[\variables]}(\vtav[\UCBindex]-\vtav[\UCBindex][\run][0])$.
We notice that the quantity $\vta[\UCBindex]=\sum_{\indv\in\variables}\vtav[\UCBindex]=\vta[\UPindexSum]+\sum_{\indv\in\variables}\vtav[\UCBindex][\run][0]$ is indeed an upper confidence bound on the reward of action $\arm$ and \UpUCBwb is equivalent to choosing the action with the largest $\vta[\UCBindex]$ in each round.
This observation indeed reduces \UpUCBwb to CombUCB~\citep{CWY13,KWAS15} on the transformed model.
Nonetheless, the \UpUCBwb formulation has several advantages:
\begin{enumerate}[topsep=2pt, itemsep=1.5pt, parsep=2pt, leftmargin=*]
    \item It captures the fact that we really care about is the uplifting effect and the variables affected by each action.
    \item The uplifting interpretation facilitates algorithm design in more involved situations as we have shown in \cref{sec:unknown-affected}.
    \item While the general analysis of CombUCB yields a regret in 
    $\bigoh((\nVariables^2+\nVariables\nArms\nAffected)\log\nRuns/\gapbandit)$, we have shown in \cref{thm:UpUCB-regret} that the regret of \UpUCBwb is in $\bigoh(\nAffected^2\log\nRuns/\gapbandit)$.
    Our analysis presented in \cref{apx:proof-UpUCBwb} also greatly benefits from the \UpUCBwb formulation of the algorihhm.
\end{enumerate}

\section{Concluding Remarks}
\label{sec:conc}

This paper addresses \acl{MAB} problems whose rewards can be written as a sum of some observable variables.
We showed that when each of the action only affects a limited number of these variables, much smaller regret can be achieved, and we developed algorithms with such guarantee under different variations of knowledge the learner possesses. While the focus of this paper is on \ac{UCB}s for stochastic bandits, we believe understanding how similar problem structure can be exploited for other algorithms
such as Thompson sampling, information directed sampling, or even methods for adversarial bandits is also important.


\newcommand\Tstrut{\rule{0pt}{2.8ex}}         
\newcommand\Bstrut{\rule[-1.4ex]{0pt}{0pt}} 
\begin{table}[!h]
    \caption{Main notations used in the paper.}
    \label{tab:notation}
    \renewcommand{\arraystretch}{0.7}
    \label{tab:notation_full}
    \begin{tabularx}{\textwidth}{@{\hspace{1em}}l@{\hspace{1.6em}}l@{\hspace{1.4em}}X@{\hspace{1em}}}
    \toprule
         & Notation & Explanation \\
         \midrule
         \multirow{12}{*}{Bandits} & $\nArms$ & Number of actions\\
         & $\arm$ & An action\\
         & $\sol[\arm]$ & An optimal action, $\sol[\arm]\in\argmax_{\arm\in\arms}\va[\reward]$\\
         & $\arms$ & The set of actions, $\arms=\intinterval{1}{\nArms}$\\
         & $\run$ & Round index \\
         & $\nRuns$ & Number of rounds / Time horizon \\
         & $\oneto{\nRuns}$ & The set of all the rounds, $\oneto{\nRuns}=\intinterval{1}{\nRuns}$\\
         & $\vt[\reward]$ & Reward received at round $\run$\\
         & $\va[\reward]$ & Expected reward of action $\arm$\\
         & $\sol[\reward]$ & Largest expected reward, $\sol[\reward]=\va[\reward][\sol[\arm]]=\max_{\arm\in\arms}\va[\reward]$ \\
         & $\va[\gapbandit]$ & Suboptimality gap of action $\arm$, $\va[\gapbandit]=\sol[\reward]-\va[\reward]$ \\
         & $\gapbandit$ & 
         Minimum non-zero suboptimality gap, \ie $\gapbandit=\min_{\arm\in\arms,\va[\gapbandit]>0}\va[\gapbandit]$ \Bstrut\\
         \hline
         \multirow{19}{*}{
         Uplifting Bandits} & $\nVariables$ & Number of underlying variables \Tstrut\\
         & $\indv$ & A variable \\
         & $\variables$ & The set of variables, $\variables=\intinterval{1}{\nVariables}$\\
         & $\va[\nAffected]$ & Number of variables affected by action $\arm$ \\
         & $\nAffected$ & 
         Upper bound on number of affected variables, \ie $\nAffected\ge\max_{\arm\in\arms}\va[\nAffected]$\\
         & $\va[\variables]$ & The set of variables affected by action $\arm$\\
         & 0 & A convenient notation to represent all quantities related to the baseline \\
         & $\arms_0$ & A notation to include both the actions and the baseline \\
         & $\va[\reward][0]$ & Baseline reward \\
         & $\va[\uplift]$ & Expected uplift of action $\arm$, $\va[\uplift]=\va[\reward]-\va[\reward][0]$ \\
         & $\sol[\uplift]$ & Largest expected uplift, $\sol[\uplift]=\va[\uplift][\sol[\arm]]=\max_{\arm\in\arms}\va[\uplift]$ \\
         & $\vt[\valuev]$ & Payoffs observed at round $\run$,  $\vt[\valuev]=(\vtv[\valuev])_{\allvariables}$ \\
         & $\va[\meanReward]$ & Expected payoffs of action $\arm$, $\va[\meanReward]=(\vav[\meanReward])_{\allvariables}$ \\
         & $\va[\meanReward][0]$ & Baseline payoffs, $\va[\meanReward][0]=(\vav[\meanReward][0])_{\allvariables}$ \\
         & $\va[\meanUplift]$ & Expected individual uplifts of action $\arm$, $\va[\meanUplift]=(\vav[\meanUplift])_{\allvariables}$\\
         & $\va[\distribution]$ & The distribution of the payoffs associated to action $\arm$ \\
         & $\va[\distribution][0]$ & Baseline distribution of the payoffs \\
         & $\gapUplift$ & 
         \makecell[l]{Lower bound on individual uplift,\\ \ie $\forall\arm\in\arms,~\indv\in\va[\variables]$, $\abs{\vav[\meanReward]-\vav[\meanReward][0]}\ge\gapUplift$}
         \Bstrut\\[8pt]
         \hline
         \multirow{8}{*}{UCB} & $\vta[\pullcount]$ & Number of times action $\arm$ is taken in the first $\run$ rounds \Tstrut\\
         & $\vta[\est{\meanReward}]$ & Empirical estimates of expected payoffs \\
         & $\vta[\radiusConf]$ & Widths of confidence intervals \\
         & $\vta[\intConf]$ & Confidence intervals \\
         & $\vta[\UCBindex]$ & Upper confidence bounds \\
         & $\vta[\UPindexSum]$ & Uplifting index of action $\arm$ \\
         & $\vta[\UPindex]$ & Individual uplifting indices \\
         & $\alt{\smallproba}$ & Error probability provided as input for \ac{UCB}-type algorithms\\
    \bottomrule
    \end{tabularx}
\end{table}

\bibliography{references}
\bibliographystyle{plainnat}


\newpage
\appendix






\section{Proofs for Upper Bound Results}
\label{apx:upper-bound}
In this appendix, we provide proofs of the regret bounds from \cref{sec:known-affected,sec:unknown-affected}.
Our proofs have the following structure.

\begin{enumerate}
    \item Leveraging \cref{lem:concentration-mean-a,lem:concentration-mean-0} presented in \cref{apx:concentration}, we define a high-probability event $\event$ on which all the expected payoffs belong to their confidence intervals.
    \item 
    For UCB-like methods, we use the fact that the uplifting index of the taken action is larger than that of any optimal action by design to show that on event $\event$ a suboptimal action can only be taken when its associated width of confidence interval is sufficiently large.
    \item The previous point allows us to bound the number of times that a suboptimal action is taken on $\event$.
    \item We conclude with the following regret decomposition (which was already stated in \eqref{eq:regret-decomp}).
    \begin{equation}
    \label{eq:regret-decomp-apx}
    \reg_{\nRuns} =
    \sum_{\arm\in\arms}\vta[\pullcount]\va[\gapbandit].
\end{equation}
\end{enumerate}

\subsection{Concentration Bounds}
\label{apx:concentration}
To prove the regret upper bounds, we will make use of concentration bounds that apply to quantities defined in \eqref{eq:mean-estimate-a} and \eqref{eq:mean-estimate-0}.
Underlying all these results is the following fundamental concentration inequality for $\noisedev$-sub-Gaussian random variables~\citep{vershynin2018high}
\begin{equation}
    \label{eq:sub-Gaussian-concentration}
    \prob(\abs{\rv}\ge\scalar)\le2\exp(-\scalar^2/(2\noisevar)) \text{ for all } \scalar>0.
\end{equation}

However, the estimates in \eqref{eq:mean-estimate-a} and \eqref{eq:mean-estimate-0} aggregate observed values in a way that depend on the decisions that have been made.
Therefore, concentration bounds for the sum of independent variables, which can be directly derived from \eqref{eq:sub-Gaussian-concentration}, are not sufficient for our purposes.
The standard trick to circumvent this issue in bandit literature is to combine Doob's optional skipping (\citep[Chap.\ 3]{Doob53}, \citep[Chap.\ 9]{Chung01}) with concentration inequality for martingale difference sequence.
The following lemma adapted from \citep[Lem.\ A.1]{SS11} is proved exactly using this technique.

\begin{lemma}
\label{lem:concentration-MDS}
Let $(\vt[\filter])_{\run\in\N}$ be a filtration and let $(\vt[\indevent],\vt[\rv])_{\run\in\N}$ be a sequence of $\{0,1\}\times\R$- valued random variables such that $\vt[\indevent]$ is $\vt[\filter][\run-1]$-measurable, and $\vt[\rv]$ is $\vt[\filter]$-measurable and conditionally $\noisedev$-sub-Gaussian, \ie $\exof{\exp(\scalar\vt[\rv])\given{\vt[\filter][\run-1]}}\le\exp(\scalar^2\noisedev^2/2)$ for all $\scalar\in\R$.
Let $\pullcount>0$ and let $\stoptime=\min\setdef{\run\ge1}{\sum_{\runalt=1}^{\run}\vt[\indevent][\runalt]=\pullcount}$, where we take $\stoptime=+\infty$ when $\sum_{\runalt=1}^{+\infty}\vt[\indevent][\runalt]<\pullcount$.
Then, for any $\smallproba\in(0,1)$,
\begin{equation}
    \prob\left(\left|\sum_{\run=1}^{\stoptime}\vt[\indevent]\vt[\rv]\right|
    \ge\noisedev\sqrt{2\pullcount\log(2/\smallproba)}
    \right)\le\smallproba.
\end{equation}
\end{lemma}
\begin{proof}
The proof of \citep[Lem. A.1]{SS11} essentially applies.
We just need to replace the use of Hoeffding-Azuma inequality by the use of concentration inequality for conditionally sub-Gaussian martingale difference sequences, which is itself derived from \eqref{eq:sub-Gaussian-concentration} and the fact that the sum of conditionally sub-Gaussian martingale differences is sub-Gaussian.
\end{proof}

In the following, $(\vt[\filter])_{\run}$ will represent the natural filtration associated to $(\vt[\valuev], \vt[\arm][\run+1])_{\run}$ such that $\vt[\arm]$ is $\vt[\filter][\run-1]$-measurable, and $\vt[\valuev]$ is $\vt[\filter]$-measurable.
Applying \cref{lem:concentration-MDS} to properly defined $(\vt[\indevent],\vt[\rv])_{\run}$ along with the use of a union bound gives the following concentration results that are key to our analyses.

\begin{lemma}
\label{lem:concentration-mean-a}
Let $\arm\in\arms$, $\indv\in\variables$, and let $\vtav[\meanReward]$ be defined as in \eqref{eq:mean-estimate-a}. Then for any $\smallproba\in(0,1)$, it holds
\begin{equation}
    \notag
    \prob
    \left(
    \exists \thinspace \run\in\nRuns, \thinspace
    \abs{\vtav[\est{\meanReward}] - \vav[\meanReward]}
    \ge
    \sqrt{\frac{2\log(2/\smallproba)}{\vta[\pullcount]}}
    \right)
    \le \nRuns\smallproba.
\end{equation}
\end{lemma}
\begin{proof}
For $\pullcount\in\oneto{\nRuns}$, we define the random variables
\begin{equation}
    \notag
    \vt[\stoptime][\pullcount]=\min
    \left\{\run\ge1 : \sum_{\runalt=1}^{\run}\one\{\vt[\arm][\runalt]=\arm\}=\pullcount\right\}
    \text{~~~and~~~}
    \vtav[\srvsum][\pullcount]=
    \sum_{\runalt=1}^{\vt[\stoptime][\pullcount]}\vtv[\valuev][\runalt]\one\{\vt[\arm][\runalt]=\arm\}.
\end{equation}
We then define the failure events
\begin{gather*}
    \vv[\event] =
    \left\{
    \exists \thinspace \run\in\oneto{\nRuns}, \thinspace
    \abs{\vtav[\est{\meanReward}] - \vav[\meanReward]}
    \ge
    \sqrt{\frac{2\log(2/\smallproba)}{\vta[\pullcount]}}
    \right\},\\
    \vv[\widetilde{\event}] =
    \left\{
    \exists \thinspace \pullcount\in\oneto{\nRuns}, \thinspace
    \left|
    \vtav[\srvsum][\pullcount]
    - \vav[\meanReward]
    \left(\sum_{\runalt=1}^{\vt[\stoptime][\pullcount]}\one\{\vt[\arm][\runalt]=\arm\}\right)
    \right|
    \ge
    \sqrt{2\pullcount\log(2/\smallproba)}.
    \right\}
\end{gather*}
It holds $\vv[\event]\subseteq\vv[\widetilde{\event}]$.
In fact, if $\vv[\event]$ happens then $\abs{\vtav[\est{\meanReward}] - \vav[\meanReward]} \ge \sqrt{2\log(2/\smallproba)/\vtav[\pullcount]}$ for some $\run$ and this is only possible when $\vtav[\pullcount]\ge 1$.
With the definition $\vtav[\pullcount]=\sum_{\runalt=1}^{\run}\one\{\vt[\arm][\runalt]=\arm\}$, we further see that $\vtav[\pullcount]\le\run\le\nRuns$ and $\vt[\stoptime][\vta[\pullcount]]\le\run<+\infty$.
The latter means that we indeed have $\sum_{\runalt=1}^{\vt[\stoptime][\vta[\pullcount]]}\one\{\vt[\arm][\runalt]=\arm\}=\vta[\pullcount]$. Thus multiplying $\abs{\vtav[\est{\meanReward}] - \vav[\meanReward]} \ge \sqrt{2\log(2/\smallproba)/\vtav[\pullcount]}$ by $\pullcount=\vta[\pullcount]\in\oneto{\nRuns}$ gives exactly
\begin{equation}
    \notag
    \left|
    \vtav[\srvsum][\pullcount]
    - \vav[\meanReward]
    \left(\sum_{\runalt=1}^{\vt[\stoptime][\pullcount]}\one\{\vt[\arm][\runalt]=\arm\}\right)
    \right|
    \ge
    \sqrt{2\pullcount\log(2/\smallproba)},
\end{equation}
which shows that $\vv[\widetilde{\event}]$ happens as well.
It is thus sufficient to prove that $\prob(\vv[\widetilde{\event}])\le\nRuns\smallproba$.

For $\pullcount\in\oneto{\nRuns}$, we apply \cref{lem:concentration-MDS} to $\vt[\indevent]\subs\one\{\vt[\arm]=\arm\}$ and $\vt[\rv]\subs\vtav[\snoise]=\vtav[\valuev]-\vav[\meanReward][\vt[\arm]]$. Since $\vav[\meanReward][\vt[\arm]]\one\{\vt[\arm]=\arm\}=\vav[\meanReward]\one\{\vt[\arm]=\arm\}$, we recover the exact inequality that appears in the definition of $\vv[\widetilde{\event}]$. We conclude with a union bound taking $\pullcount$ ranging from $1$ to $\nRuns$.
\end{proof}

\begin{lemma}
\label{lem:concentration-mean-0}
Let $\indv\in\variables$, and let $\vtav[\meanReward][\run][0]$ be defined as in \eqref{eq:mean-estimate-0}. Then for any $\smallproba\in(0,1)$, it holds
\begin{equation}
    \notag
    \prob
    \left(
    \exists \thinspace \run\in\nRuns, \thinspace
    \abs{\vtav[\est{\meanReward}][\run][0] - \vav[\meanReward][0]}
    \ge
    \sqrt{\frac{2\log(2/\smallproba)}{\vtav[\pullcount][\run][0]}}
    \right)
    \le \nRuns\smallproba.
\end{equation}
\end{lemma}
\begin{proof}
\cref{lem:concentration-mean-0} is proved in the same way as \cref{lem:concentration-mean-a}.
In particular, we apply \cref{lem:concentration-MDS} to $\vt[\indevent]\subs\one\{\indv\notin\va[\variables][\vt[\arm]]\}$ and $\vt[\rv]\subs\vtav[\snoise][\run][0]$.
\end{proof}

\subsection{Unknown Baseline: Proof of \cref{thm:UpUCB-regret}}
\label{apx:proof-UpUCBwb}

\UpUCBRegret*

\begin{proof}
We first notice that while there are $\nVariables$ variables, not all of them are involved in estimating the uplifts.
Therefore, we only need to focus on the set of relevant variables $\alt{\variables}=\bigcup_{\arm\in\arms}\va[\variables]$.
Clearly, $\card(\alt{\variables})\le\nArms\nAffected$.
With this in mind, we define the events
\begin{align*}
    \event_1 &=
    \{\forall\run\in\oneto{\nRuns},
    ~\forall\arm\in\arms,
    ~\forall\indv\in\va[\variables],
    ~ \abs{\vtav[\est{\meanReward}]-\vav[\meanReward]}\le\vta[\radiusConf]\},\\
    \event_2 &=
    \{\forall\run\in\oneto{\nRuns}, 
    ~\forall\indv\in\alt{\variables},
    ~ \abs{\vtav[\est{\meanReward}][\run][0]-\vav[\meanReward][0]}\le\vtav[\radiusConf][\run][0]\}.
\end{align*}
\cref{lem:concentration-mean-a} and \cref{lem:concentration-mean-0} in \cref{apx:concentration} along with union bounds over $\arm$ and $\indv$ guarantee
\begin{equation}
    \notag
    \prob(\event_1\union\event_2)
    \ge1-\left(\sum_{\arm\in\arms}\va[\nAffected]+\card(\alt{\variables})\right)\nRuns(2\alt{\smallproba})
    \ge1-4\nArms\nAffected\nRuns\alt{\smallproba}
    = 1 - \smallproba.
\end{equation}
It is thus sufficient to show that \eqref{eq:UpUCB-regret} holds on $\event_1\union\event_2$.

In the remainder of the proof, we consider an arbitrary realization of $\event_1\union\event_2$ and prove that \eqref{eq:UpUCB-regret} indeed holds for this realization.
From \eqref{eq:regret-decomp-apx} it is clear that we just need to bound $\vta[\pullcount][\nRuns]=\sum_{\run=1}^{\nRuns}\one\{\vt[\arm]=\arm\}$ from above for all suboptimal action $\arm\in\arms$ with $\va[\gapbandit]>0$.
This is done by providing a lower bound on $\vta[\radiusConf]$ for any $\run$ such that $\vt[\arm]=\arm$.
To proceed, suppose that a suboptimal action $\arm$ is taken for the last time at round $\run\in\intinterval{\nArms+1}{\nRuns}$.\footnote{%
If the action is not taken anymore after the first $\nRuns$ runs. We have $\vta[\pullcount][\nRuns]=1$ so the upper bound that we show later trivially holds.
}
This implies $\vta[\UPindexSum][\run]\ge\vta[\UPindexSum][\run][\sol[\arm]]$ and $\vta[\pullcount][\nRuns]=\vta[\pullcount][\run-1]+1$. By the definition of the uplifting indices, we get
\begin{equation}
    \notag
    \sum_{\indv\in\va[\variables]}(\vtav[\UCBindex][\run]-\vtav[\UCBindex][\run][0])
    \ge
    \sum_{\indv\in\va[\variables][\sol[\arm]]}
    (\vtav[\UCBindex][\run][\sol[\arm]]-\vtav[\UCBindex][\run][0]).
\end{equation}
Rearranging and dropping $\sum_{\indv\in\va[\variables]\intersect\va[\variables][\sol[\arm]]}\vtav[\UCBindex][\run][0]$ from both sides, we get
\begin{equation}
    \label{eq:UCB-larger-than-UCB}
    \sum_{\indv\in\va[\variables]}\vtav[\UCBindex][\run]
    +\sum_{\indv\in\setexclude{\va[\variables][\sol[\arm]]}{\va[\variables]}}\vtav[\UCBindex][\run][0]
    \ge
    \sum_{\indv\in\va[\variables][\sol[\arm]]}\vtav[\UCBindex][\run][\sol[\arm]]
    +\sum_{\indv\in\setexclude{\va[\variables]}{\va[\variables][\sol[\arm]]}}\vtav[\UCBindex][\run][0].
\end{equation} 
This is similar to the type of inequality that defines standard UCB.
In fact, the left and the right hand sides of \eqref{eq:UCB-larger-than-UCB} are respectively upper confidence bounds on $\sum_{\indv\in\va[\variables]\union\va[\variables][\sol[\arm]]}\vav[\meanReward]$
and upper confidence bounds on $\sum_{\indv\in\va[\variables]\union\va[\variables][\sol[\arm]]}\vav[\meanReward][\sol[\arm]]$.
We next bound these two quantities respectively from above and from below.
Since both $\event_1$ and $\event_2$ hold, we have the following inequalities for the UCB indices
%
\begin{gather*}
    \forall\runalt\in\oneto{\nRuns},
    ~\forall\arm\in\arms, 
    ~\forall\indv\in\va[\variables],
    ~\vav[\meanReward] + 2\vta[\radiusConf][\runalt-1]
    \ge \vtav[\UCBindex][\runalt]
    \ge \vav[\meanReward],\\
    \forall\runalt\in\oneto{\nRuns}, 
    ~\forall\indv\in\setexclude{\alt{\variables}}{\bigcap_{\arm\in\arms}\va[\variables]},
    ~\vav[\meanReward][0] + 2\vtav[\radiusConf][\runalt-1][0]
    \ge \vtav[\UCBindex][\runalt][0]
    \ge \vav[\meanReward][0].
\end{gather*}
Combining the above, we obtain
\begin{equation}
    \notag
    \sum_{\indv\in\va[\variables]}
    (\vav[\meanReward]+2\vta[\radiusConf][\run-1])
    +\sum_{\indv\in\setexclude{\va[\variables][\sol[\arm]]}{\va[\variables]}}
    (\vav[\meanReward][0]+2\vtav[\radiusConf][\run-1][0])
    \ge
    \sum_{\indv\in\va[\variables][\sol[\arm]]}\vav[\meanReward][\sol[\arm]]
    +\sum_{\indv\in\setexclude{\va[\variables]}{\va[\variables][\sol[\arm]]}}\vav[\meanReward][0].
\end{equation}
Adding $\sum_{\indv\in\va[\variables]\intersect\va[\variables][\sol[\arm]]}\vav[\meanReward][0]$ to both sides and rearranging leads to
\begin{equation}
    \notag
    \sum_{\indv\in\va[\variables][\sol[\arm]]}
    \vav[\meanReward][\sol[\arm]]
    - \sum_{\indv\in\va[\variables]}\vav[\meanReward]
    + \sum_{\indv\in\va[\variables]}\vav[\meanReward][0]
    - \sum_{\indv\in\va[\variables][\sol[\arm]]}\vav[\meanReward][0]
    \le \sum_{\indv\in\va[\variables]}2\vta[\radiusConf][\run-1]
    + \sum_{\indv\in\setexclude{\va[\variables][\sol[\arm]]}{\va[\variables]}}2\vtav[\radiusConf][\run-1][0].
\end{equation}
From this we can derive an inequality between the suboptimal gap $\va[\gapbandit]$ and the the widths of the confidence intervals as follows
\begin{equation}
    \label{eq:UpUCB-regret-bound-gap}
    \begin{aligned}[b]
    \va[\gapbandit]
    &= \va[\uplift][\sol[\arm]] - \va[\uplift]\\
    &= \sum_{\indv\in\va[\variables][\sol[\arm]]}
    (\vav[\meanReward][\sol[\arm]]-\vav[\meanReward][0])
    -\sum_{\indv\in\va[\variables]}(\vav[\meanReward]-\vav[\meanReward][0])\\
    &= \sum_{\indv\in\va[\variables][\sol[\arm]]}
    \vav[\meanReward][\sol[\arm]]
    - \sum_{\indv\in\va[\variables]}\vav[\meanReward]
    + \sum_{\indv\in\va[\variables]}\vav[\meanReward][0]
    - \sum_{\indv\in\va[\variables][\sol[\arm]]}\vav[\meanReward][0]\\
    &\le 2\va[\nAffected]\vta[\radiusConf][\run-1]
    + \sum_{\indv\in\setexclude{\va[\variables][\sol[\arm]]}{\va[\variables]}}2\vtav[\radiusConf][\run-1][0].
    \end{aligned}
\end{equation}
We next argue that for all $\indv\in\setexclude{\va[\variables][\sol[\arm]]}{\va[\variables]}$ we have $\vtav[\pullcount][\run-1][0]\ge\vta[\pullcount][\run-1]$ and hence $\vtav[\radiusConf][\run-1][0]\le\vta[\radiusConf][\run-1]$.
In fact, if $\indv\notin\va[\variables]$, then whenever action $\arm$ is taken, the count of $\vtav[\pullcount][\runalt][0]$ also increases by $1$. Formally,
\begin{equation}
    \notag
    \vtav[\pullcount][\run-1][0]
    = \sum_{\runalt=1}^{\run-1}\one\{\indv\notin\va[\variables][\vt[\arm][\runalt]]\}
    \ge \sum_{\runalt=1}^{\run-1}\one\{\vt[\arm][\runalt]=\arm\}\one\{\indv\notin\va[\variables][\vt[\arm][\runalt]]\}
    = \sum_{\runalt=1}^{\run-1}\one\{\vt[\arm][\runalt]=\arm\}
    = \vta[\pullcount][\run-1].
\end{equation}
%
Therefore, we further deduce $\va[\gapbandit]\le2(\va[\nAffected]+\va[\nAffected][\sol[\arm]])\vta[\radiusConf][\run-1]$. Equivalently,

\begin{equation}
    \notag
    \vta[\pullcount][\run-1]
    \le \frac{8(\va[\nAffected]+\va[\nAffected][\sol[\arm]])^2\log(1/\alt{\smallproba})}{(\va[\gapbandit])^2}.
\end{equation}
%
%
We conclude that action $\arm$ is taken at most
$\floor{8(\va[\nAffected]+\va[\nAffected][\sol[\arm]])^2\log(1/\alt{\smallproba})/(\va[\gapbandit])^2}+1$
times during the execution of the algorithm. Plugging this into \eqref{eq:regret-decomp-apx} gives exactly \eqref{eq:UpUCB-regret}.
\end{proof}


\subsection{Unknown Affected Variables: Proof of~\cref{thm:UpUCB-L-regret}}
\label{apx:upper-bound-unknown-affected}

Below we prove \cref{thm:UpUCB-L-regret} separately for \UpUCBL and \UpUCBLwb.

\begin{proposition} [Regret of \UpUCBL]
\label{prop:UpUCB-L-BL-regret}
Let $\alt{\smallproba}=\smallproba/(2\nArms\nVariables\nRuns)$. If the learner knows  and an upper bound $\nAffected$ on maximum number of variables an action affects and the baseline payoffs, then \UpUCBL (\cref{algo:UpUCB-L-BL}), with probability at least $1-\smallproba$, has regret bounded by:
\begin{gather}
    \label{eq:UpUCB-L-BL-regret}
    \vt[\reg][\nRuns] \le
    \sum_{\arm\in\arms:\va[\gapbandit]>0}
    \left(\frac{32\nAffected^2\log(2\nArms\nVariables\nRuns/ \smallproba)}{\va[\gapbandit]}+\va[\gapbandit]\right).
    \raisetag{0.75em}
\end{gather}
\end{proposition}
\begin{proof}
Let us consider the event
\begin{equation}
    \label{eq:UCB-L-event}
    \event =
    \{\forall\run\in\oneto{\nRuns},
    ~\forall\arm\in\arms,
    ~\forall\indv\in\variables,
    ~ \abs{\vtav[\est{\meanReward}]-\vav[\meanReward]}<\vta[\radiusConf]\},
\end{equation}
%
By \cref{lem:concentration-mean-0} and a union bound we get immediately $\prob(\event)\ge1-\nArms\nVariables\nRuns(2\alt{\smallproba})=1-\smallproba$.
In the following, we assume that $\event$ occurs and prove \eqref{eq:UpUCB-L-BL-regret}.

First, by the definition of $\vta[\setId]$ we have clearly $\vta[\setId]\subseteq\va[\variables]$.
In fact, if $\indv\notin\va[\variables]$ then $\vav[\meanReward][0]=\vav[\meanReward]\in\vtav[\intConf]$.
The inclusion $\vta[\setId]\subseteq\va[\variables]$ along with the condition $\card(\va[\variables])\le\nAffected$ imply  $\card(\setexclude{\va[\variables]}{\vta[\setId]})\le\nAffected-\card(\vta[\setId])$;
in particular, $\nAffected-\card(\vta[\setId])\ge0$ and thus $\vta[\nAffected]=\nAffected - \card(\vta[\setId])$.

Next, for any $\indv\in\setexclude{\variables}{\vta[\setId]}$, it holds $\vav[\meanReward][0]\in\vtav[\intConf]$. This implies $\vav[\meanReward][0]\le\vtav[\UCBindex]$, and accordingly $\vtav[\UPindex]\ge0$.
We can then write
\begin{equation}
    \notag
    \sum_{\indv\in\setexclude{\va[\variables]}{\vta[\setId]}}\vtav[\UPindex]
    \le
    \max_{\substack{\setLargeUp\subseteq\setexclude{\variables}{\vta[\setId]}\\
    \card(\setLargeUp)=\card(\setexclude{\variables}{\vta[\setId]})}}
    \sum_{\indv\in\setLargeUp}\vtav[\UPindex]
    \le
    \max_{\substack{\setLargeUp\subseteq\setexclude{\variables}{\vta[\setId]}\\
    \card(\setLargeUp)=\nAffected-\card(\vta[\setId])}}
    \sum_{\indv\in\setLargeUp}\vtav[\UPindex]
    =
    \sum_{\indv\in\vta[\setLargeUp]}\vtav[\UPindex].
\end{equation}
Since $\vtav[\UCBindex]\ge\vav[\meanReward]$ on $\event$, we have $\vtav[\UPindex]\ge\vav[\meanUplift]$; subsequently
\begin{equation}
    \notag
    \va[\uplift]
    = \sum_{\indv\in\va[\variables]}\vav[\meanUplift]
    =
    \sum_{\indv\in\vta[\setId]}\vav[\meanUplift]
    + \sum_{\indv\in\setexclude{\va[\variables]}{\vta[\setId]}}\vav[\meanUplift]
    \le
    \sum_{\indv\in\vta[\setId]}\vtav[\UPindex]
    + \sum_{\indv\in\vta[\setLargeUp]}\vtav[\UPindex]
    =\vta[\UPindexSum].
\end{equation} 
This shows that $\vta[\UPindexSum]$ is effectively an upper bound of $\va[\uplift]$.
With this in mind, we are ready to bound the number of times that a suboptimal action is taken.

Let $\arm\in\arms$ with $\va[\gapbandit]>0$ and assume that it is taken at round $\run\in\intinterval{\nArms+1}{\nRuns}$, which means that $\vta[\UPindexSum][\run]\ge\vta[\UPindexSum][\run][\sol[\arm]]$.
We have shown that $\vta[\UPindexSum][\run][\sol[\arm]]\ge\va[\uplift][\sol[\arm]]$.
It remains to provide an upper bound for $\vta[\UPindexSum][\run]$ that involves $\va[\uplift]$.
%
%
For this, we again use $\vta[\setId]\subseteq\va[\variables]$ and the fact that $\vtav[\UPindex]\ge0$ for all $\indv\in\setexclude{\variables}{\vta[\setId][\run]}$ to decompose
\begin{equation}
    \label{eq:UpUCB-L-BL-upindex-decomposition}
    \begin{aligned}
    \vta[\UPindexSum][\run]
    = \sum_{\indv\in\vta[\setId][\run]\union\vta[\setLargeUp][\run]}\vtav[\UPindex][\run]
    \le \sum_{\indv\in\va[\variables]\union\vta[\setLargeUp][\run]}\vtav[\UPindex][\run]
    = \sum_{\indv\in\va[\variables]}\vtav[\UPindex][\run]
    + \sum_{\indv\in\setexclude{\vta[\setLargeUp]}{\va[\variables]}}\vtav[\UPindex][\run].
    \end{aligned}
\end{equation}
The summation of the individual uplifting indices over $\indv\in\va[\variables]$ is naturally related to $\va[\uplift]$.
Since $\vav[\meanReward]\ge\vtav[\est{\meanReward}][\run-1]-\vta[\radiusConf][\run-1]$, for all $\indv\in\variables$ it holds
%
\begin{equation}
    \notag
    \vtav[\UPindex][\run]
    =
    \vtav[\UCBindex] - \vav[\meanReward]
    \le \vav[\meanReward] - \vav[\meanReward][0] +2\vta[\radiusConf][\run-1]
    =\vav[\meanUplift]+2\vta[\radiusConf][\run-1].
\end{equation}
In consequence,
\begin{equation}
    \label{eq:UpUCB-L-BL-decomp-upper1}
    \sum_{\indv\in\va[\variables]}\vtav[\UPindex][\run]
    \le\sum_{\indv\in\va[\variables]}(\vav[\meanUplift]+2\vta[\radiusConf][\run-1])
    \le\va[\uplift]+2\nAffected\vta[\radiusConf][\run-1].
\end{equation}
What we need to show next is that the definition of $\vta[\setId]$ guarantees $\vtav[\UPindex]$ to be small whenever $\indv\notin\vta[\setId]$.
In fact, if $\indv\in\setexclude{\variables}{\vta[\setId][\run]}$, we have $\vav[\meanReward][0]\in\vtav[\intConf][\run]$, and in particular
$\vav[\meanReward][0]\ge\vtav[\est{\meanReward}][\run-1]-\vta[\radiusConf][\run-1]$, implying that
\begin{equation}
    \notag
    \vtav[\UPindex][\run]
    =\vtav[\UCBindex][\run]-\vav[\meanReward][0]
    \le (\vtav[\est{\meanReward}][\run-1]+\vta[\radiusConf][\run-1])
    - (\vtav[\est{\meanReward}][\run-1]-\vta[\radiusConf][\run-1])
    = 2\vta[\radiusConf][\run-1].
\end{equation}
By definition, $\vta[\setLargeUp]\subseteq\setexclude{\variables}{\vta[\setId][\run]}$, and hence
\begin{equation}
    \label{eq:UpUCB-L-BL-decomp-upper2}
    \sum_{\indv\in\setexclude{\vta[\setLargeUp][\run]}{\va[\variables]}}\vtav[\UPindex][\run]
    \le\sum_{\indv\in\setexclude{\vta[\setLargeUp][\run]}{\va[\variables]}}2\vta[\radiusConf][\run-1]
    \le2\nAffected\vta[\radiusConf][\run-1].
\end{equation}
Putting \eqref{eq:UpUCB-L-BL-upindex-decomposition}, \eqref{eq:UpUCB-L-BL-decomp-upper1}, and \eqref{eq:UpUCB-L-BL-decomp-upper2} together, we get
\begin{equation}
    \notag
    \begin{aligned}
    \vta[\UPindexSum][\run]
    \le \sum_{\indv\in\va[\variables]}\vtav[\UPindex][\run]
    + \sum_{\indv\in\setexclude{\vta[\setLargeUp]}{\va[\variables]}}\vtav[\UPindex][\run]
    \le \va[\uplift] +4\nAffected\vta[\radiusConf][\run-1].
    \end{aligned}
\end{equation}
With $\vta[\UPindexSum][\run]\ge\vta[\UPindexSum][\run][\sol[\arm]]\ge\va[\uplift][\sol[\arm]]$ we deduce
\begin{equation}
    \notag
    \va[\uplift][\sol[\arm]] \le  \va[\uplift] +4\nAffected\vta[\radiusConf][\run-1].
\end{equation}
Therefore, 
$\va[\gapbandit] = \va[\uplift][\sol[\arm]]-\va[\uplift] \le 4\nAffected\vta[\radiusConf][\run-1]$, and equivalently
\begin{equation}
    \notag
    \vta[\pullcount][\run-1]
    \le \frac{32\nAffected^2\log(1/\alt{\smallproba})}{(\va[\gapbandit])^2}.
\end{equation} 
%
Following the argument in the proof of \cref{thm:UpUCB-regret}, this translates into an upper bound on $\vta[\pullcount][\nRuns]$ and
we conclude by 
invoking the decomposition of \eqref{eq:regret-decomp-apx}.
\end{proof}

We notice that it is quite straightforward to adapt \UpUCBL and its analysis to the setting where the learner knows an action-dependent upper bound $\overline{\va[\nAffected]}\ge\va[\nAffected]$. The only change in \eqref{eq:UpUCB-L-BL-regret} is that $\nAffected$ is replaced by $\overline{\va[\nAffected]}$. Thus we recover \eqref{eq:UpUCB-BL-regret} up to a multiplicative constant when $\overline{\va[\nAffected]}=\Theta(\va[\nAffected])$.
To derive the regret bound for \UpUCBLwb, we combine idea from the proofs of \cref{thm:UpUCB-regret,prop:UpUCB-L-BL-regret}.



\begin{proposition}[Regret of \UpUCBLwb] \label{prop:2}
Let $\alt{\smallproba}=\smallproba/(2\nArms\nVariables\nRuns)$. Then the regret of \UpUCBLwb (\cref{algo:UpUCB-L}), with probability at least $1-\smallproba$, satisfies:
\begin{equation}
    \label{eq:UpUCB-L-regret}
    \vt[\reg][\nRuns] \le
    \sum_{\arm\in\arms:\va[\gapbandit]>0}
    \left(\frac{512\nAffected^2\log(2\nArms\nVariables\nRuns/\smallproba)}{\va[\gapbandit]}+\va[\gapbandit]\right),
\end{equation}
\end{proposition}

\begin{proof}

We again place ourselves in the event $\event$ defined in \eqref{eq:UCB-L-event}, and prove \eqref{eq:UpUCB-L-regret} under this condition.
Recall that $\vt[\mostpullarm]\in\argmax_{\arm\in\arms}\vta[\pullcount][\run-1]$ is an action that is taken most frequently during the first $\run-1$ rounds and the associated estimates serve as baselines at round $\run$. We define $\vta[\mostpullvariables]=\va[\variables]\union\va[\variables][\vt[\mostpullarm]]$ as the set of variables that are affected by either action $\arm$ or $\vt[\mostpullarm]$.
Clearly, $\card(\vta[\mostpullvariables])\le2\nAffected$.

We first show that $\vta[\setId]\subseteq\vta[\mostpullvariables]$ whenever $\event$ occurs.
Let $\indv\in\setexclude{\variables}{\vta[\mostpullvariables]}$.
Under $\event$ we have both $\vav[\meanReward][0]=\vav[\meanReward]\in\vtav[\intConf]$ and $\vav[\meanReward][0]=\vav[\meanReward][\vt[\mostpullarm]]\in\vtav[\intConf][\run][\mostpullarm]$.
Therefore, $\vtav[\intConf]\intersect\vtav[\intConf][\run][\vt[\mostpullarm]]\neq\varnothing$; subsequently $\indv\notin\vta[\setId]$. The contrapositive of the above gives exactly $\vta[\setId]\subseteq\vta[\mostpullvariables]$.

We shall now bound $\vta[\pullcount][\nRuns]$ for any suboptimal action $\arm$, from which the proof concludes by applying \eqref{eq:regret-decomp-apx}.
Let $\arm\in\arms$ satisfying $\va[\gapbandit]>0$. If it is taken at round $\run\in\intinterval{\nArms+1}{\nRuns}$, we have $\vta[\UPindexSum]\ge\vta[\UPindexSum][\run][\sol[\arm]]$.
This can be written as
%
\begin{equation}
    \label{eq:UpUCB-L-pull-round-larger}
    \sum_{\indv\in\vta[\setId]}\vtav[\UPindex]
    +\sum_{\indv\in\vta[\setLargeUp]}\vtav[\UPindex]
    \ge
    \sum_{\indv\in\vta[\setId][\run][\sol[\arm]]}\vtav[\UPindex][\run][\sol[\arm]]
    +\sum_{\indv\in\vta[\setLargeUp][\run][\sol[\arm]]}\vtav[\UPindex][\run][\sol[\arm]].
\end{equation}

\paragraph{Decomposition of Suboptimality Gap using the New Baseline.}
Similar to before, we will use \eqref{eq:UpUCB-L-pull-round-larger} to relate the widths of the confidence intervals to the suboptimality gap $\va[\gapbandit]$.
For this, we notice that the suboptimality gap of $\arm$ can be rewritten as
\begin{equation}
    \label{eq:proof-UpUCBLWB-gap-decompose}
    \begin{aligned}[b]
    \va[\gapbandit]
    &=\sum_{\indv\in\variables}\vav[\meanReward][\sol[\arm]]
    -\sum_{\indv\in\variables}\vav[\meanReward]\\
    &=\sum_{\indv\in\variables}(\vav[\meanReward][\sol[\arm]]-\vav[\meanReward][\vt[\mostpullarm]])
    -\sum_{\indv\in\variables}(\vav[\meanReward]-\vav[\meanReward][\vt[\mostpullarm]])\\
    &=\sum_{\indv\in\vta[\mostpullvariables][\run][\sol[\arm]]}(\vav[\meanReward][\sol[\arm]]-\vav[\meanReward][\vt[\mostpullarm]])
    -\sum_{\indv\in\vta[\mostpullvariables]}(\vav[\meanReward]-\vav[\meanReward][\vt[\mostpullarm]])
    \end{aligned}
\end{equation} 
The last equality is true because $\vav[\meanReward][\sol[\arm]]=\vav[\meanReward][0]$ outside $\vta[\mostpullvariables][\run][\sol[\arm]]$, and the same holds when we replace $\sol[\arm]$ by $\arm$.
Therefore, our goal is to lower bound the \acl{RHS} of \eqref{eq:UpUCB-L-pull-round-larger} by an expression that contains $\sum_{\indv\in\vta[\mostpullvariables][\run][\sol[\arm]]}(\vav[\meanReward][\sol[\arm]]-\vav[\meanReward][\vt[\mostpullarm]])$
and upper bound the \acl{LHS} of \eqref{eq:UpUCB-L-pull-round-larger} by an expression that contains $\sum_{\indv\in\vta[\mostpullvariables]}(\vav[\meanReward]-\vav[\meanReward][\vt[\mostpullarm]])$.

\paragraph{Lower Bound on RHS of \eqref{eq:UpUCB-L-pull-round-larger}\afterhead}
Since $\sum_{\indv\in\vta[\setLargeUp][\run][\sol[\arm]]}\vtav[\UPindex][\run][\sol[\arm]]$ is the maximum padding, we can lower bound the \acl{RHS} of the inequality by $\sum_{\indv\in\vta[\mostpullvariables][\run][\sol[\arm]]}\vtav[\UPindex][\run][\sol[\arm]]$.
Formally, with $\vta[\setId][\run][\sol[\arm]]\subseteq\vta[\mostpullvariables][\run][\sol[\arm]]$ we have $\card(\setexclude{\vta[\mostpullvariables][\run][\sol[\arm]]}{\vta[\setId][\run][\sol[\arm]]})\le2\nAffected-\card(\vta[\setId][\run][\sol[\arm]])$ and
$\vta[{\nAffected}][\run][\sol[\arm]]=2\nAffected-\card(\vta[\setId][\run][\sol[\arm]])$.
Thus, by definition of $\vta[\setLargeUp][\run][\sol[\arm]]$,
\begin{equation}
    \notag
    \sum_{\indv\in\vta[\setLargeUp][\run][\sol[\arm]]}\vtav[\UPindex][\run][\sol[\arm]]
    =
    \max_{\substack{\setLargeUp\subseteq\setexclude{\va[\variables]}{\vta[\setId]}\\
    \card(\setLargeUp)\le2\nAffected-\card(\vta[\setId][\run][\sol[\arm]])}}
    \sum_{\indv\in\setLargeUp}\vtav[\UPindex][\run][\sol[\arm]]
    \ge
    \sum_{\indv\in\setexclude{\vta[\mostpullvariables][\run][\sol[\arm]]}{\vta[\setId][\run][\sol[\arm]]}}\vtav[\UPindex][\run][\sol[\arm]].
\end{equation}
To further provide a lower bound on $\sum_{\indv\in\vta[\mostpullvariables][\run][\sol[\arm]]}\vtav[\UPindex][\run][\sol[\arm]]$, we recall that $\vtav[\UPindex][\run][\sol[\arm]]=\vtav[\UCBindex][\run][\sol[\arm]]-\vtav[\UCBindex][\run][\vt[\mostpullarm]]$, so with $\vtav[\UCBindex][\run][\sol[\arm]]\ge\vav[\meanReward][\sol[\arm]]$, $\vtav[\UCBindex][\run][\vt[\mostpullarm]]\le\vav[\meanReward][\vt[\mostpullarm]]+2\vta[\radiusConf][\run-1][\vt[\mostpullarm]]$, and $\card(\vta[\mostpullvariables][\run][\sol[\arm]])\le2\nAffected$, we get
\begin{equation}
    \label{eq:UpUCB-L-pull-lower}
    \begin{aligned}[b]
    \sum_{\indv\in\vta[\setId][\run][\sol[\arm]]}\vtav[\UPindex][\run][\sol[\arm]]
    +\sum_{\indv\in\vta[\setLargeUp][\run][\sol[\arm]]}\vtav[\UPindex][\run][\sol[\arm]]
    &\ge
    \sum_{\indv\in\vta[\mostpullvariables][\run][\sol[\arm]]}\vtav[\UPindex][\run][\sol[\arm]]\\
    &\ge \sum_{\indv\in\vta[\mostpullvariables][\run][\sol[\arm]]}
    (\vav[\meanReward][\sol[\arm]] - \vav[\meanReward][\vt[\mostpullarm]] - 2\vta[\radiusConf][\run-1][\vt[\mostpullarm]])\\
    &\ge
    \sum_{\indv\in\vta[\mostpullvariables][\run][\sol[\arm]]}
    (\vav[\meanReward][\sol[\arm]] - \vav[\meanReward][\vt[\mostpullarm]]) - 4\nAffected\vta[\radiusConf][\run-1][\vt[\mostpullarm]].
    \end{aligned}
\end{equation}

\paragraph{Upper Bound on LHS of \eqref{eq:UpUCB-L-pull-round-larger}\afterhead}
To make the term $\sum_{\indv\in\vta[\mostpullvariables]}(\vav[\meanReward]-\vav[\meanReward][\vt[\mostpullarm]])$ appear,
we first relate it to the sum of the uplifting indices using $\vtav[\UCBindex]\le\vav[\meanReward]+2\vta[\radiusConf][\run-1]$.
\begin{equation}
    \label{eq:proof-UpUCBLWB-LHS-upper1}
    \sum_{\indv\in\vta[\mostpullvariables]}\vtav[\UPindex]
    \le
    \sum_{\indv\in\vta[\mostpullvariables]}(\vav[\meanReward]+2\vta[\radiusConf][\run-1]-\vav[\meanReward][\vt[\mostpullarm]])
    \le \sum_{\indv\in\vta[\mostpullvariables]}(\vav[\meanReward]-\vav[\meanReward][\vt[\mostpullarm]])
    +4\nAffected\vta[\radiusConf][\run-1].
\end{equation}
The last inequality uses $\card(\vta[\mostpullvariables])\le2\nAffected$.
Next, to relate $\sum_{\indv\in\vta[\mostpullvariables]}\vtav[\UPindex]$ to the \acl{LHS} of \eqref{eq:UpUCB-L-pull-round-larger}, we want to show
\begin{enumerate*}[\textup(\itshape i\textup)]
    \item $\sum_{\indv\in\vta[\setLargeUp]}\vtav[\UPindex]$ is small, and
    \item We can add additional terms from $\setexclude{\vta[\mostpullvariables]}{\vta[\setId]}$ to the \acl{LHS} of \eqref{eq:UpUCB-L-pull-round-larger}.
\end{enumerate*}
These two points correspond respectively to inequalities \eqref{eq:UpUCB-L-BL-decomp-upper2} and \eqref{eq:UpUCB-L-BL-upindex-decomposition} in the proof of \cref{prop:UpUCB-L-BL-regret}.
Again, this can be done by looking at the definition of $\vta[\setId]$.
In fact, if $\indv\in\setexclude{\variables}{\vta[\setId]}$, then by definition $\vtav[\intConf]\intersect\vtav[\intConf][\run][\vt[\mostpullarm]]\neq\varnothing$, implying that
\begin{equation}
    \notag
    \begin{aligned}
    \vtav[\est{\meanReward}][\run-1][\vt[\mostpullarm]]+\vta[\radiusConf][\run-1][\vt[\mostpullarm]]
    &\ge\vtav[\est{\meanReward}][\run-1]-\vta[\radiusConf][\run-1],\\
    \vtav[\est{\meanReward}][\run-1]+\vta[\radiusConf][\run-1]
    &\ge\vtav[\est{\meanReward}][\run-1][\vt[\mostpullarm]]-\vta[\radiusConf][\run-1][\vt[\mostpullarm]].
    \end{aligned}
\end{equation}
%
The above two inequalities respectively translate into
\begin{equation}
    \notag
    \begin{aligned}
    \vtav[\UCBindex]-\vtav[\UCBindex][\run][\vt[\mostpullarm]]
    &\le 2\vta[\radiusConf][\run-1],\\
    \vtav[\UCBindex]-\vtav[\UCBindex][\run][\vt[\mostpullarm]]
    &\ge -2\vta[\radiusConf][\run-1][\vt[\mostpullarm]],
    \end{aligned}
\end{equation}
%
In other words, for any $\indv\in\setexclude{\variables}{\vta[\setId]}$ we have $-2\vta[\radiusConf][\run-1][\vt[\mostpullarm]]\le\vtav[\UPindex]\le2\vta[\radiusConf][\run-1]$.
Subsequently,
\begin{align}
    \label{eq:UpUCB-L-ineq-unidentified1}
    \sum_{\indv\in\vta[\setLargeUp]}\vtav[\UPindex]
    &\le \sum_{\indv\in\vta[\setLargeUp]}2\vta[\radiusConf][\run-1],\\
    \label{eq:UpUCB-L-ineq-unidentified2}
    \sum_{\indv\in\setexclude{\vta[\mostpullvariables]}{\vta[\setId]}}
    -2\vta[\radiusConf][\run-1][\vt[\mostpullarm]]
    &\le
    \sum_{\indv\in\setexclude{\vta[\mostpullvariables]}{\vta[\setId]}}\vtav[\UPindex].
\end{align} 
Therefore
\begin{equation}
    \label{eq:proof-UpUCBLWB-LHS-upper2}
    \begin{aligned}[b]
    \sum_{\indv\in\vta[\setId]}\vtav[\UPindex]
    +\sum_{\indv\in\vta[\setLargeUp]}\vtav[\UPindex]
    &\le
    \sum_{\indv\in\vta[\setId]}\vtav[\UPindex]
    + \sum_{\indv\in\vta[\setLargeUp]}2\vta[\radiusConf][\run-1]
    + \sum_{\indv\in\setexclude{\vta[\mostpullvariables]}{\vta[\setId]}}
    (\vtav[\UPindex]+2\vta[\radiusConf][\run-1][\vt[\mostpullarm]])\\
    &=\sum_{\indv\in\vta[\mostpullvariables]}\vtav[\UPindex]
    + \sum_{\indv\in\vta[\setLargeUp]}2\vta[\radiusConf][\run-1]
    + \sum_{\indv\in\setexclude{\vta[\mostpullvariables]}{\vta[\setId]}}
    2\vta[\radiusConf][\run-1][\vt[\mostpullarm]]\\
    &\le\sum_{\indv\in\vta[\mostpullvariables]}\vtav[\UPindex]
    +4\nAffected\vta[\radiusConf][\run-1]
    +4\nAffected\vta[\radiusConf][\run-1][\vt[\mostpullarm]].
    \end{aligned}
\end{equation}
The first inequality makes use of \eqref{eq:UpUCB-L-ineq-unidentified1} and \eqref{eq:UpUCB-L-ineq-unidentified2}, and in particular the last term on the \acl{RHS} is non-negative thanks to \eqref{eq:UpUCB-L-ineq-unidentified2}.
The equality in the second line is true because $\vta[\setId]\subseteq\vta[\mostpullvariables]$.
Finally, for the last inequality we use the fact that $\card(\vta[\setLargeUp])\le2\nAffected$ and $\card(\vta[\mostpullvariables])\le2\nAffected$.
Along with \eqref{eq:proof-UpUCBLWB-LHS-upper2} we get an upper bound on $\sum_{\indv\in\vta[\setId]}\vtav[\UPindex]
    +\sum_{\indv\in\vta[\setLargeUp]}\vtav[\UPindex]$ that involves $\sum_{\indv\in\vta[\mostpullvariables]}(\vav[\meanReward]-\vav[\meanReward][\vt[\mostpullarm]])$.

\paragraph{Conclusion\afterhead}
Combining \eqref{eq:UpUCB-L-pull-round-larger}, \eqref{eq:UpUCB-L-pull-lower}, \eqref{eq:proof-UpUCBLWB-LHS-upper1}, and \eqref{eq:proof-UpUCBLWB-LHS-upper2}, we obtain
\begin{equation}
    \notag
    \sum_{\indv\in\vta[\mostpullvariables]}(\vav[\meanReward]-\vav[\meanReward][\vt[\mostpullarm]])
    + 8\nAffected\vta[\radiusConf][\run-1]
    + 4\nAffected\vta[\radiusConf][\run-1][\vt[\mostpullarm]]
    \ge \sum_{\indv\in\vta[\mostpullvariables][\run][\sol[\arm]]}
    (\vav[\meanReward][\sol[\arm]] - \vav[\meanReward][\vt[\mostpullarm]]) - 4\nAffected\vta[\radiusConf][\run-1][\vt[\mostpullarm]].
\end{equation}
By the choice of $\vt[\mostpullarm]$ we have $\vta[\radiusConf][\run-1]\ge\vta[\radiusConf][\run-1][\vt[\mostpullarm]]$.
With \eqref{eq:proof-UpUCBLWB-gap-decompose}, it follows that $16\nAffected\vta[\radiusConf][\run-1]\ge\va[\gapbandit]$ and thus
\begin{equation}
    \notag
    \vta[\pullcount][\run-1]
    \le \frac{512\nAffected^2\log(1/\alt{\smallproba})}{(\va[\gapbandit])^2}.
\end{equation}
Repeating the argument in the proof of \cref{thm:UpUCB-regret} (\ie choose $\run$ to be the last time that action $\arm$ is taken if it is taken more than once and apply decomposition \eqref{eq:regret-decomp-apx}) gives the desired result.
\end{proof}


\UpUCBLWBRegret*
\begin{proof}
The proof follows by combining~\cref{prop:UpUCB-L-BL-regret,prop:2}.
\end{proof}

\subsection{Unknown Affected Variables: Proofs of \cref{thm:UpUCB-GapUp-BL-regret,thm:UpUCB-GapUp-regret}}
\label{apx:lower-bound-unknown-affected}
\UpUCBGapUpBL*
\begin{proof}
As before, we prove \eqref{eq:UpUCB-GapUp-BL-regret} under the condition that $\event$ defined in \eqref{eq:UCB-L-event} occurs. Recall that we have shown in the proof of \cref{prop:UpUCB-L-BL-regret} that $\prob(\event)\ge1-\smallproba$ with our choice of $\alt{\smallproba}$, so this effectively leads to a high-probability regret bound.

In the following we assume that $\event$ occurs and show that for any suboptimal action $\arm\in\arms$, it holds
\begin{equation}
    \label{eq:UpUCB-GapUp-pullcount-bound}
    \vta[\pullcount][\nRuns]
    \le
    \max\left(
    \min\left(
    \frac{8\log(1/\alt{\smallproba})}{\gapUplift^2},
    \frac{8\nVariables^2\log(1/\alt{\smallproba})}{(\va[\gapbandit])^2}\right), \frac{8(\va[\nAffected])^2\log(1/\alt{\smallproba})}{(\va[\gapbandit])^2}\right)
    +1.
\end{equation}
The inequality \eqref{eq:UpUCB-GapUp-BL-regret} then follows directly from \eqref{eq:regret-decomp-apx} and a rearrangement of the terms.

We first claim that for any $\arm\in\arms$ if $\vta[\pullcount]\ge\pullcount_0$ then $\vta[\setId]=\va[\variables]$.
In fact, by the definition of $\pullcount_0$, $\vta[\pullcount]\ge\pullcount_0$ implies that $\vta[\radiusConf]<\gapUplift/2$.
Therefore, as argued in the text, $\indv\in\vta[\setId]$ if and only if $\indv\in\va[\variables]$.
Since $\vtav[\UPindex]\ge\vav[\meanUplift]$ for any $\arm$ and $\indv$, this also proves that $\vta[\UPindex]\ge\va[\uplift]$ always holds (we have either $\vta[\setId]=\variables$ or $\vta[\setId]=\va[\variables]$).

Let $\run\in\intinterval{\nArms+1}{\nRuns}$ be the last time that a suboptimal action $\arm$ is taken so that $\vta[\pullcount][\nRuns]=\vta[\pullcount][\run-1]+1$.
This indicates $\vta[\UPindexSum]\ge\vta[\UPindexSum][\run][\sol[\arm]]$, and hence $\vta[\UPindexSum]\ge\va[\uplift][\sol[\arm]]$.
We distinguish between the following two cases:
\begin{enumerate}[itemsep=2.5pt,topsep=2.5pt,parsep=2.5pt]
    \item $\vta[\pullcount][\run-1]<\pullcount_0$:
    Then $\vta[\setId]=\variables$ so $\vta[\UPindexSum]=\sum_{\indv\in\variables}(\vtav[\UCBindex]-\va[\meanReward][0])
    \le\va[\uplift]+2\nVariables\vta[\radiusConf][\run-1]$, which in turn leads to $\vta[\radiusConf][\run-1]\ge\va[\gapbandit]/(2\nVariables)$, or equivalently $\vta[\pullcount][\run-1]\le8\nVariables^2\log(1/\alt{\smallproba})/(\va[\gapbandit])^2$.
    \item $\vta[\pullcount][\run-1]\ge\pullcount_0$:
    Then $\vta[\setId]=\va[\variables]$ so $\vta[\UPindexSum]=\sum_{\indv\in\va[\variables]}(\vtav[\UCBindex]-\va[\meanReward][0])
    \le\va[\uplift]+2\va[\nAffected]\vta[\radiusConf][\run-1]$, which in turn leads to $\vta[\radiusConf][\run-1]\ge\va[\gapbandit]/(2\va[\nAffected])$, or equivalently $\vta[\pullcount][\run-1]\le8(\va[\nAffected])^2\log(1/\alt{\smallproba})/(\va[\gapbandit])^2$.
\end{enumerate}
Since we also have $\vta[\pullcount][\run-1]<\pullcount_0$ in the first case, the above gives
\begin{equation}
    \notag
    \vta[\pullcount][\run-1]
    \le
    \max\left(
    \min\left(
    \pullcount_0-1,\frac{8\nVariables^2\log(1/\alt{\smallproba})}{(\va[\gapbandit])^2}\right), \frac{8(\va[\nAffected])^2\log(1/\alt{\smallproba})}{(\va[\gapbandit])^2}\right).
\end{equation}
Plugging in the definition of $\pullcount_0$ we get immediately \eqref{eq:UpUCB-GapUp-pullcount-bound}.
\end{proof}

\UpUCBGapUpregret*
\begin{proof}
We shall again consider the event $\event$ defined in \eqref{eq:UCB-L-event}, which holds with probability $1-\smallproba$ as argued in the proof of \cref{prop:UpUCB-L-BL-regret}. Leveraging the decomposition in \eqref{eq:regret-decomp-apx} it is sufficient to prove that on $\event$ we have
\begin{equation}
    \label{eq:UpUCB-GapUp-noBL-pullcount-bound}
    \vta[\pullcount][\nRuns]
    \le
    \max\left(
    \min\left(
    \frac{32\log(1/\alt{\smallproba})}{\gapUplift^2},
    \frac{32\nVariables^2\log(1/\alt{\smallproba})}{(\va[\gapbandit])^2}\right), \frac{8(\va[\nAffected]+\va[\nAffected][\sol[\arm]])^2\log(1/\alt{\smallproba})}{(\va[\gapbandit])^2}\right)
    +1.
\end{equation}
We assume $\event$ happens in the rest of the proof.

Let us first focus on the elimination phase, notice that event $\event$ implies $\abs{\vta[\estalt{\reward}][\round]-\va[\reward]}<\vt[\radiusConf][\round]$ for every $\arm\in\vt[\arms][\round]$.
In this case, the optimal actions never get eliminated.
In fact, if $\sol[\arm]\in\vt[\arms][\round]$, then for all $\arm\in\vt[\arms][\round]$ it holds that
\begin{equation}
    \notag
    \vta[\estalt{\reward}][\round][\sol[\arm]]
    +2\vt[\radiusConf][\round]
    \ge
    \va[\reward][\sol[\arm]]+\vt[\radiusConf][\round]
    \ge
    \va[\reward][\arm]+\vt[\radiusConf][\round]
    \ge \vta[\estalt{\reward}][\round][\arm],
\end{equation}
As a consequence, we also have $\sol[\arm]\in\vt[\arms][\round+1]$ and we conclude by the principle of induction.
With this we further deduce that all action
$\arm$ with $\va[\gapbandit]\ge4\vt[\radiusConf][\round]$ gets eliminated
after it is taken at most $\round$ times.
To see this, notice that for such action we have
\begin{equation}
    \notag
    \vta[\estalt{\reward}][\round]+2\vt[\radiusConf][\round]
    < \va[\reward]+3\vt[\radiusConf][\round]
    = \va[\reward][\sol[\arm]]-\va[\gapbandit]+3\vt[\radiusConf][\round]
    < \vta[\estalt{\reward}][\round][\sol[\arm]]
    -\va[\gapbandit]+4\vt[\radiusConf][\round]
    \le \vta[\estalt{\reward}][\round][\sol[\arm]].
\end{equation}
Therefore, for all $\run$ in phase \rom{1} of the algorithm, it is true that
\begin{equation}
    \label{eq:UpUCB-GapUp-noBL-pullcount-bound-I}
    \vta[\pullcount]\le
    \min\left(
    \left\lceil\frac{32\log(1/\alt{\smallproba})}{\gapUplift^2}\right\rceil,
    \left\lceil\frac{32\nVariables^2\log(1/\alt{\smallproba})}{(\va[\gapbandit])^2}\right\rceil
    \right).
\end{equation}
This suggests that inequality \eqref{eq:UpUCB-GapUp-noBL-pullcount-bound} holds if the algorithm terminates in the elimination phase.

Otherwise, the algorithm constructs $\vv[\setBase]$'s, $\va[\setId]$'s, and switches to the \textsc{UpUCB} phase.
Let us denote by $\vv[\setBase]=\setdef{\arm\in\arms}{\indv\notin\va[\variables]}$ the set of actions that do not affect variable $\indv$
and let $\arms_{\text{res}}=\vt[\arms][\pullcount_0+1]$ be the actions that remain after phase \rom{1}.
We argue that the sets $\vv[\setBaseEst]$ are constructed such that when $\event$ occurs,
\begin{enumerate}[\textup(\itshape i\textup)]
    \item $\vv[\setBaseEst]\subseteq\vv[\setBase]$.
    \item  If there exist $\arm,\armalt\in\arms_{\text{res}}\intersect\vv[\setBase]$ such that $\arm\neq\armalt$ then we also have $\arm,\armalt\in\vv[\setBaseEst]$.
\end{enumerate}
The second point is proved by observing that under $\event$ we must have $\vav[\meanReward][0]\in\vtav[\intConf][\run_0+1]\intersect\vtav[\intConf][\run_0+1][\armalt]$ if $\arm,\armalt\in\arms_{\text{res}}\intersect\vv[\setBase]$.
To show the first point, notice that after the elimination phase, each of the action in $\arms_{\text{res}}$ is taken exactly $\ceil{(32/\gapUplift^2)\log(1/\alt{\smallproba})}$ times and thus $\vta[\radiusConf][\run_0]\le\gapUplift/4$.
By \cref{asm:gapUplift-ubl}, if $\arm\notin\vv[\setBase]$, then for all $\armalt\in\setexclude{\arms_{\text{res}}}{\{\arm\}}$ we have $\abs{\vav[\meanReward]-\vav[\meanReward][\alt{\arm}]}\ge\gapUplift$, which along with the definition of $\event$ implies $\vtav[\intConf][\run_0+1]\intersect\vtav[\intConf][\run_0+1][\armalt]=\varnothing$, and therefore $\arm\notin\vv[\setBaseEst]$.

The rest of the proof follows closely that of \cref{thm:UpUCB-regret}.
Suppose that a suboptimal action $\arm$ is taken at round $\run$ during phase \rom{2} of the algorithm.
This means $\vta[\UPindexSum]\ge\vta[\UPindexSum][\run][\sol[\arm]]$ (recall that $\sol[\arm]\in\arms_{\text{res}}$ since $\event$ happens), \ie
\begin{equation}
    \notag
    \sum_{\indv\in\va[\setId]}(\vtav[\UCBindex][\run]-\vtav[\UCBindex][\run][0])
    \ge
    \sum_{\indv\in\va[\setId][\sol[\arm]]}
    (\vtav[\UCBindex][\run][\sol[\arm]]-\vtav[\UCBindex][\run][0]).
\end{equation}
Rearranging the terms we get 
\begin{equation}
    \notag
    \sum_{\indv\in\va[\setId]}\vtav[\UCBindex][\run]
    +\sum_{\indv\in\setexclude{\va[\setId][\sol[\arm]]}{\va[\setId]}}
    \vtav[\UCBindex][\run][0]
    \ge
    \sum_{\indv\in\va[\setId][\sol[\arm]]}
    \vtav[\UCBindex][\run][\sol[\arm]]
    +\sum_{\indv\in\setexclude{\va[\setId]}{\va[\setId][\sol[\arm]]}}
    \vtav[\UCBindex][\run][0].
\end{equation}
If $\indv\in\setexclude{\va[\setId][\sol[\arm]]}{\va[\setId]}$ then $\arm\in\vv[\setBaseEst]$.
In particular, $\vv[\setBaseEst]\neq\varnothing$ so $\armalt=\vtav[\arm][\run][0]=\argmax_{\tilde{\arm}\in\vv[\setBaseEst]}\vta[\pullcount][\run-1][\tilde{\arm}]$ is well-defined and $\vtav[\UCBindex][\run][0]=\vtav[\UCBindex][\run][\armalt]$.
With $\vv[\setBaseEst]\subseteq\vv[\setBase]$ we know that $\vav[\meanReward][\armalt]=\vav[\meanReward][0]$ and accordingly $\abs{\vtav[\meanReward][\run-1][\armalt]-\vav[\meanReward][0]}\le\vta[\radiusConf][\run-1][\armalt]$ by the fact that $\event$ happens.
This shows $\vtav[\UCBindex][\run][0]\ge\vav[\meanReward][0]$.
Moreover, $\vta[\pullcount][\run][\armalt]\ge\vta[\pullcount]$ by the choice of $\armalt$; hence, $\vtav[\UCBindex][\run][0]\le\vav[\meanReward][0]+2\vta[\radiusConf][\run-1][\armalt]\le\vav[\meanReward][0]+2\vta[\radiusConf][\run-1]$.
The same argument applies to $\indv\in\setexclude{\va[\setId]}{\va[\setId][\sol[\arm]]}$, and the UCB indices of $\arm$ and $\sol[\arm]$ can be bounded from above and below as in previous proofs.
In summary, we obtain
\begin{equation}
    \notag
    \sum_{\indv\in\va[\setId]}(\vav[\meanReward]+2\vta[\radiusConf][\run-1])
    +\sum_{\indv\in\setexclude{\va[\setId][\sol[\arm]]}{\va[\setId]}}
    (\vav[\meanReward][0]+2\vta[\radiusConf][\run-1])
    \ge
    \sum_{\indv\in\va[\setId][\sol[\arm]]}
    \vav[\meanReward][\sol[\arm]]
    +\sum_{\indv\in\setexclude{\va[\setId]}{\va[\setId][\sol[\arm]]}}
    \vav[\meanReward][0].
\end{equation}
Equivalently,
\begin{equation}
    \label{eq:UpUCB-GapUp-radConf-larger-than}
    2\card\left(\va[\setId]\union\va[\setId][\sol[\arm]]\right)\vta[\radiusConf][\run-1]
    \ge
    \sum_{\indv\in\va[\setId]\union\va[\setId][\sol[\arm]]}\vav[\meanReward][\sol[\arm]]
    -\sum_{\indv\in\va[\setId]\union\va[\setId][\sol[\arm]]}\vav[\meanReward].
\end{equation}
To conclude, we claim that
\begin{equation}
    \label{eq:VV-equal}
    \va[\setId]\union\va[\setId][\sol[\arm]]=\va[\variables]\union\va[\variables][\sol[\arm]].
\end{equation}
Let us first show $\va[\variables]\subseteq\va[\setId]\subseteq\va[\variables]\union\va[\variables][\sol[\arm]]$.
The first inclusion is a consequence of $\vv[\setBaseEst]\subseteq\vv[\setBase]$:
\begin{equation}
    \notag
    \indv\in\va[\variables] \implies
    \arm\notin\vv[\setBase] \implies
    \arm\not\in\vv[\setBaseEst] \implies
    \indv\in\va[\setId].
\end{equation}
As for the second inclusion we prove it's contrapositive:
\begin{equation}
    \notag
    \indv\notin\va[\variables]\union\va[\variables][\sol[\arm]] \implies
    \arm,\sol[\arm]\in\arms_{\text{res}}\intersect\vv[\setBase] \implies
    \arm,\sol[\arm]\in\vv[\setBaseEst] \implies
    \indv\notin\va[\setId]
\end{equation}
Notice that the second implication holds because $\arm$ is suboptimal and in particular $\arm\neq\sol[\arm]$.
The inclusion $\va[\variables][\sol[\arm]]\subseteq\va[\setId][\sol[\arm]]\subseteq\va[\variables]\union\va[\variables][\sol[\arm]]$ can be proved in a similar way, and combining these results gives \eqref{eq:VV-equal}.
Therefore, the \acl{RHS} of \eqref{eq:UpUCB-GapUp-radConf-larger-than} is exactly $\va[\gapbandit]$ while the \acl{LHS} of \eqref{eq:UpUCB-GapUp-radConf-larger-than} is bounded from above by $2(\va[\nAffected]+\va[\nAffected][\sol[\arm]])\vta[\radiusConf][\run-1]$.
In consequence,
\begin{equation}
    \notag
    \vta[\pullcount][\run-1]
    \le \frac{8(\va[\nAffected]+\va[\nAffected][\sol[\arm]])^2\log(1/\alt{\smallproba})}{(\va[\gapbandit])^2}.
\end{equation}
Therefore, any suboptimal action that gets taken in the \textsc{UpUCB} phase satisfies
\begin{equation}
    \notag
    \vta[\pullcount][\nRuns]
    \le \frac{8(\va[\nAffected]+\va[\nAffected][\sol[\arm]])^2\log(1/\alt{\smallproba})}{(\va[\gapbandit])^2}+1.
\end{equation}
On the other hand, if an suboptimal action is not taken in the \textsc{UpUCB} phase then \eqref{eq:UpUCB-GapUp-noBL-pullcount-bound-I} applies, so in either case \eqref{eq:UpUCB-GapUp-noBL-pullcount-bound} is verified, which concludes the proof.
\end{proof}

\section{Lower Bounds}
\label{apx:lower-bound}

To prove our lower bounds, 
we fist establish in \cref{apx:lower-bound-information} a general lemma for deriving instance-dependent lower-bounds for bandit problems with underlying variables, side observations, and prior knowledge on the distribution of the variables.
Subsequently, in \cref{apx:proof-lower-bound} we regard uplifting bandit as a special case of this model and apply the lemma to prove \cref{prop:lower-bound,prop:lower-bound-L}.

\subsection{A General Information-theoretic Lower Bound}
\label{apx:lower-bound-information}


Let $\valuev=(\vv[\valuev][1], \ldots, \vv[\valuev][\nVariables])$ be a random vector of $\nVariables$ variables 
that entirely determines the reward so that we can write $\reward=\rewardfunc(\valuev)$ for some deterministic reward function $\rewardfunc\from\R^{\nVariables}\to\R$.
Let $\va[\distribution]$ be the distribution of action $\arm$ on the underlying variables and $\va[\distribution][0]$ be the baseline distribution on these variables.
At each round after taking an action, the learner observes a vector in $\R^{\vdim}$ that is itself a function of the variables, written as $\obs(\valuev)$ where $\obs\from\R^{\nVariables}\to\R^{\vdim}$ is the observation function.
The decision of which action to take can then only be based on the interaction history $(\vt[\arm][1],\obs(\vt[\valuev][1]),\ldots,\vt[\arm][\run-1],\obs(\vt[\valuev][\run-1]))$ and some prior knowledge of the learner about the distributions of the variables $\priorknow(\va[\distribution][0],\ldots,\va[\distribution][\nArms])$.\footnote{For our purpose, we can regard this as a partition of the set of all bandit instances that the learner may encounter.}

A class of action distributions is defined as $\banditclass=\va[\setDist][0]\times\cdots\times\va[\setDist][\nArms]$ where each $\va[\setDist]$ is a set of distributions on $\R^{\nVariables}$. 
We say that $\banditclass$ is indistinguishable under $\priorknow$ if for all $\distribution_{\ast}=(\va[\distribution])_{\arm\in\arms_0},\distributionalt_{\ast}=(\va[\distributionalt])_{\arm\in\arms_0}\in\banditclass$ we have $\priorknow(\va[\distribution][0],\ldots,\va[\distribution][\nArms])=\priorknow(\va[\distributionalt][0],\ldots,\va[\distributionalt][\nArms])$.
A bandit problem with underlying variables, side observations, and prior knowledge on the variable distribution is thus defined by the quadruple $(\banditclass,\rewardfunc,\obs,\priorknow)$.
A policy $\policy$ is said to be \emph{consistent} over $(\banditclass,\rewardfunc)$ if for all $p>0$ and all instance of $(\banditclass,\rewardfunc)$, the regret of $\policy$ on the instance satisfies $\vt[\reg][\run]=\smalloh(\run^{p})$.
A policy $\policy$ is compatible with observation $\obs$ and prior knowledge $\priorknow$ if it can be implemented by a learner that observes $\obs(\valuev)$ and has prior knowledge $\priorknow(\va[\distribution][0],\ldots,\va[\distribution][\nArms])$.

By abuse of notation, for any distributions $\distribution$ on $\R^{\nVariables}$ we will write $\rewardfunc(\distribution)=\ex_{\valuev\sim\distribution}(\rewardfunc(\valuev))$ as the expected reward when the variables follow distribution $\distribution$ whenever this quantity is defined.
We also write $\distribution_{\obs}$ as the pushforward of $\distribution$ along $\obs$.
Let $\setDist$ be a set of distributions such that $\rewardfunc(\distribution)$ is well-defined for any $\distribution\in\setDist$.
Let $\sol[\reward]\in\R$ and $\distribution\in\setDist$ such that $\rewardfunc(\distribution)<\sol[\reward]$.
The following quantity is crucial to the analysis
\begin{equation}
    \notag
    \infkl(\distribution,\setDist,\sol[\reward],\rewardfunc,\obs)=\inf_{\distributionalt\in\setDist}\setdef{\dkl(\distribution_{\obs},\distributionalt_{\obs})}{\rewardfunc(\distributionalt)>\sol[\reward]},
\end{equation}
where $\dkl(\distribution_{\obs},\distributionalt_{\obs})$ denotes the KL divergence of $\distribution_{\obs}$ from $\distributionalt_{\obs}$.
Intuitively, it quantities how difficult it is to learn that the expected reward $\rewardfunc(\distribution)$ is smaller than $\sol[\reward]$ (smaller the value of $\infkl$ the more difficult it is).
The next lemma is a straightforward adaptation of \citep[Th. 1]{BK96},\citep[Th. 16.2]{LS20} to our model.
It provides a lower bound on the asymptotic regret 


\begin{lemma}
\label{lem:lower-bound-instance-dependent}
Let $\banditclass=\va[\setDist][0]\times\cdots\times\va[\setDist][\nArms]$ be a class of action distributions and $\rewardfunc$ be the reward function.
Let $\policy$ be a consistent algorithm over $(\banditclass,\rewardfunc)$ that is compatible with prior knowledge function $\priorknow$ and observation function $\obs$. Then, if $\banditclass$ is indistinguishable under $\priorknow$, for all $\distribution_{\ast}\in\banditclass$, it holds that
\begin{equation}
    \label{eq:lower-bound-lemma}
    \liminf_{\toinf[\nRuns]} \frac{\ex[\vt[\reg][\nRuns]]}{\log\nRuns}
    \ge 
    \sum_{\arm\in\arms:\va[\gapbandit]>0}
    \frac{\va[\gapbandit]}{\infkl(\va[\distribution],\va[\setDist],\sol[\reward],\rewardfunc,\obs)}.
\end{equation}
\end{lemma}
\begin{proof}
The proof follow closely the one presented in \citep{LS20}.
The key observation is that the same proof still carries out if we have a random vector and our reward and observation depends on this random vector.
In particular, let $\measure(\policy,\distribution_{\ast})$ and $\measure(\policy,\distributionalt_{\ast})$ be the probability measure on the actions and the observations induced by the interaction of the policy with the bandit instances $(\distribution_{\ast},\rewardfunc)$ and $(\distributionalt_{\ast},\rewardfunc)$ for $\distribution_{\ast},\distributionalt_{\ast}\in\banditclass$.
The assumption that $\policy$ is compatible with $\obs$ and $\priorknow$ and that $\banditclass$ is indistinguishable under $\priorknow$ allows us to write 
$q_{\policy,\distribution_{\ast}}(\vt[\arm][1],\vt[\obs][1],\ldots,\vt[\arm][\nRuns],\vt[\obs][\nRuns])=\prod_{\run=1}^{\nRuns}\vt[\policy^{\banditclass}](\vt[\arm]|\vt[\arm][1],\vt[\obs][1],\ldots,\vt[\arm][\run-1],\vt[\obs][\run-1])\va[p_{\obs}][\vt[\arm]](\vt[\obs])$, where $\vt[\obs][1],\ldots,\vt[\obs][\nRuns]$ are the observations of the learner, $q_{\policy,\distribution_{\ast}}$ and $\va[p_{\obs}][\vt[\arm]]$ are respectively the probability density functions of $\measure(\policy,\distribution_{\ast})$ and $\va[\distribution_{\obs}]$, and $\vt[\policy^{\banditclass}]=\vt[\policy][\run,\priorknow(\distribution_{\ast})]=\vt[\policy][\run,\priorknow(\distributionalt_{\ast})]$ is a probability kernel that can be chosen using prior knowledge.\footnote{Many technical details are omitted here. We refer the interested readers to \citep{LS20} for a rigorous treatment of such proof.}
We have a similar decomposition for $q_{\policy,\distributionalt_{\ast}}$.
Then, following the proof of \citep[Lem. 15.1]{LS20} we get immediately
\begin{equation}
    \notag
    \dkl(\measure(\policy,\distribution_{\ast}),\measure(\policy,\distributionalt_{\ast}))=\sum_{\arm\in\arms}\ex_{\policy,\distribution}[\vta[\pullcount][\nRuns]]\dkl(\va[\distribution_{\obs}],\va[\distributionalt_{\obs}]).
\end{equation}
In the above, $\ex_{\policy,\distribution}$ means that the expectation is taken with respect to $\measure(\policy,\distribution_{\ast})$. The proof can be completed in the same way as done for \citep[Th. 16.2]{LS20}.
\end{proof}

\subsection{Lower Bounds for Uplifting Bandits-- Proof}
\label{apx:proof-lower-bound}

The following proposition includes both \cref{prop:lower-bound} and \cref{prop:lower-bound-L}.

\begin{proposition}
\label{prop:lower-bound-app}
Let $\policy$ be a consistent algorithm over the class of $1$-sub-Gaussian uplifting bandits that at most uses knowledge of $\va[\distribution][0]$, $(\va[\variables])_{\arm\in\arms}$, and the fact that the noise is $1$-sub-Gaussian.
Then, for any $\nArms,\nVariables>0$ and sequence $(\va[\nAffected],\va[\gapbandit])_{1\le\arm\le\nArms}\in(\oneto{\nVariables}\times\R_+)^{\nArms}$
with $\va[\gapbandit][1]=0$, there exists a $1$-sub-Gaussian uplifting bandit with parameters $(\nArms,\nVariables,(\va[\gapbandit],\va[\nAffected])_{\arm\in\arms})$
such that the regret induced by $\policy$ on it satisfies
\begin{equation}
    \label{eq:lower-bound-L}
    \liminf_{\toinf[\nRuns]} \frac{\ex[\vt[\reg][\nRuns]]}{\log\nRuns} \ge \sum_{\arm\in\arms:\va[\gapbandit]>0}\frac{2(\va[\nAffected])^2}{\va[\gapbandit]}.
\end{equation} 
%
Moreover, if $\policy$ can be implemented under either of the following conditions
\begin{enumerate}[(a)]
    \item Only the reward is observed.
    \item The learner does not have any prior knowledge about the arms' expected payoffs.
\end{enumerate}
Then, there exists a $1$-sub-Gaussian uplifting bandit with parameters $(\nArms,\nVariables,(\va[\gapbandit],\va[\nAffected])_{\arm\in\arms})$ where the regret of $\policy$ is
\begin{equation}
    \label{eq:lower-bound-m}
    \liminf_{\toinf[\nRuns]} \frac{\ex[\vt[\reg][\nRuns]]}{\log\nRuns} \ge \sum_{\arm\in\arms:\va[\gapbandit]>0}\frac{2\nVariables^2}{\va[\gapbandit]}.
\end{equation}
\end{proposition}

\begin{proof}
Let $\sol[\reward]=1+\max_{\arm\in\arms}\va[\gapbandit]>0$.
Throughout the proofs, the lower bound will be shown for problem instances with $\vav[\meanReward][0]\equiv0$ and the following mean values for $\arm\in\arms$
\begin{equation}
    \label{eq:lower-bound-proof-mean-value}
    \vav[\meanReward][\arm]=\begin{dcases}
    \frac{\sol[\reward]-\va[\gapbandit]}{\va[\nAffected]} & ~\text{if }  1\le\indv\le\va[\nAffected],\\
    \hfil 0 & ~\text{if } \va[\nAffected]+1\le\indv\le\nVariables.
    \end{dcases}
\end{equation}
By construction, we have 
$\vav[\meanReward]\neq\vav[\meanReward][0]$ if and only if $1\le\indv\le\va[\nAffected]$ and thus the number of mean-affected variables of $\arm$ is exactly $\va[\nAffected]$.
Clearly, $1$ is an optimal action and the suboptimality gap of action $\arm$ is $\va[\gapbandit]$.

Next, we will need to define the classes of action distributions $\banditclass_1, \banditclass_2, \banditclass_3$ that we will use for proving \eqref{eq:lower-bound-L}, (a)-\eqref{eq:lower-bound-m}, and (b)-\eqref{eq:lower-bound-m}.
To begin, we want these classes to be indistinguishable under a certain prior knowledge function, and since we allow the learner to know the baseline distribution, $\va[\setDist][0]$ must be a singleton.
As for $\va[\setDist]$, it will take the form $\va[\setDist]=\setdef{\gaussian(\meanReward,\va[\covmat])}{\meanReward\in\va[\meanRewards]}$ for some mean vector set $\va[\meanRewards]\subset \R^{\nVariables}$ and covariance matrix $\va[\covmat]\in[0,1]^{\nVariables\times\nVariables}$ to be specified.
This implies that every noise variable $\vav[\snoise]$ is $1$-sub-Gaussian and thus $\policy$ is consistent in each of these classes.
Moreover, the classes will also be defined in such a way that when the mean values are taken as in \eqref{eq:lower-bound-proof-mean-value}, $\va[\distribution][0]$ and $\va[\distribution]$ indeed have the same marginal distribution on all but the first $\va[\nAffected]$ variables.

\paragraph{The Importance of Limited Number of Affected Variables\afterhead}
To prove \eqref{eq:lower-bound-L}, we let the unique element of $\setDist_0$ be the Dirac measure that assigns full weight to the point $(0,\ldots,0)$. For any $\arm\in\arms$, we define $\va[\meanRewards]=(\setexclude{\R}{\{0\}})^{\va[\nAffected]}\times\{0\}^{\va[\nVariables-\nAffected]}$ and block matrix $\va[\covmat]$ such that $\vam[\covmat]=\one\{\max(\indv,\indvalt)\le\va[\nAffected]\}$. In other words, the random variable $\va[\valuev]$ can be written in the following form
\begin{equation}
    \notag
    \vav[\valuev][\arm]=\begin{dcases}
    \vav[\meanReward][\arm]+\va[\snoise] & ~\text{if } 1\le\indv\le\va[\nAffected],\\
    \hfil 0 & ~\text{if } \va[\nAffected]+1\le\indv\le\nVariables,
    \end{dcases}
\end{equation}
where $\va[\snoise]$ is Gaussian with zero mean and unit variance. Notice that the same noise applies to all the affected variables.
By construction, every $\va[\distribution]\in\va[\setDist]$ has exactly the first $\va[\nAffected]$ variables affected, \ie $\va[\variables]=\{1,\ldots,\va[\nAffected]\}$.
Thus $\banditclass_1$ defined in this way is indistinguishable under prior knowledge on baseline distribution $\va[\distribution][0]$ and the sets of affected variables $(\va[\variables])_{\arm\in\arms}$.
To conclude, we would like to apply \cref{lem:lower-bound-instance-dependent}.
For this, we observe that for any $\va[\distribution],\va[\distributionalt]\in\va[\setDist]$ with means $\va[\meanReward]$ and $\va[\meanRewardalt]$, it holds\footnote{This can for example be proved by applying the chain rule of KL divergence.}
\begin{equation}
    \label{eq:dkl-PQ-B1}
    \dkl(\va[\distribution],\va[\distributionalt]  )=\begin{dcases}
    \frac{\gapvariable^2}{2} & ~\text{if } \vav[\meanReward]-\vav[\meanRewardalt]=\gapvariable 
    \text{ for all } \indv\in\intinterval{1}{\va[\nAffected]},\\
    \hfil +\infty & ~\text{otherwise}.
    \end{dcases}
\end{equation}
With this we see immediately that for $\va[\distribution]$ with mean values \eqref{eq:lower-bound-proof-mean-value}, we have $\infkl(\va[\distribution],\va[\setDist],\sol[\reward],(+),\Idf)=\va[\gapbandit]/(2(\va[\nAffected])^2)$, where $(+)$ and $\Idf$ are respectively the sum and the identity function that gives the reward and the observations of the learner.
Plugging this into \eqref{eq:lower-bound-lemma} gives exactly \eqref{eq:lower-bound-L}.

\paragraph{The importance of Side Observation\afterhead}
To construct $\banditclass_2$, we draw the same noise for all the variables. In terms of covariance matrix, this means all the entries of $\covmat$ is equal to $1$. 
The baseline distribution is $\va[\distribution][0]=\gaussian(\bb{0},\covmat)$ and for every $\arm$, we set $\va[\covmat]=\covmat$ and choose again $\va[\meanRewards]=(\setexclude{\R}{\{0\}})^{\va[\nAffected]}\times\{0\}^{\va[\nVariables-\nAffected]}$.
This time, only the reward, that is, the sum of the variables, is observed,
and the distribution of the reward at action $\arm$ is $\gaussian(\sol[\reward]-\va[\gapbandit],\nVariables^2)$.
For any two real numbers $\scalar, \alt{\scalar}$, one has $\dkl(\gaussian(\scalar,\nVariables^2),\gaussian(\alt{\scalar},\nVariables^2))=(\scalar-\alt{\scalar})^2/(2\nVariables^2)$.
Therefore $\infkl(\va[\distribution],\va[\setDist],\sol[\reward],(+),(+))=\va[\gapbandit]/(2\nVariables^2)$ for $\va[\distribution]$ with mean values \eqref{eq:lower-bound-proof-mean-value}.
Notice that $\banditclass_2$ is also indistinguishable under prior knowledge on the baseline distribution and on the sets of affected variables $(\va[\variables])_{\arm\in\arms}$, and thus \cref{lem:lower-bound-instance-dependent} can be applied, leading to \eqref{eq:lower-bound-m}.

\paragraph{The Importance of Prior Knowledge on How the Variables are Affected\afterhead}
The class $\banditclass_3$ is defined almost in the same way as $\banditclass_2$ except for the fact that we now set $\va[\meanRewards]=\R^{\nVariables}$ for every $\arm\in\arms$.
Our condition (b) postulates that $\banditclass_3$ is indistinguishable under the prior knowledge of the learner.
Similar to \eqref{eq:dkl-PQ-B1}, for any $\va[\distribution],\va[\distributionalt]\in\va[\setDist]$ with means $\va[\meanReward]$ and $\va[\meanRewardalt]$, we now have
\begin{equation}
    \notag
    \dkl(\va[\distribution],\va[\distributionalt])=\begin{dcases}
    \frac{\gapvariable^2}{2} & ~\text{if } \vav[\meanReward]-\vav[\meanRewardalt]=\gapvariable 
    \text{ for all } \indv\in\intinterval{1}{\nVariables},\\
    \hfil +\infty & ~\text{otherwise}.
    \end{dcases}
\end{equation}
In particular, if $\va[\meanReward]$ is defined as in \eqref{eq:lower-bound-proof-mean-value} and $\vav[\meanRewardalt]=\vav[\meanReward]+(\va[\gapbandit]+\eps)/\nVariables$ for all $\indv$ for some $\eps\in\R$, we have $\dkl(\va[\distribution],\va[\distributionalt])=(\va[\gapbandit]+\eps)^2/(2\nVariables^2)$.
We deduce immediately $\infkl(\va[\distribution],\va[\setDist],\sol[\reward],(+),\Idf)=\va[\gapbandit]/(2\nVariables^2)$.
Invoking \cref{lem:lower-bound-instance-dependent} completes the proof.
\end{proof}


\section{Contextual Combinatorial Uplifting Bandits}
\label{apx:C2UpUCB}



In this appendix, we discuss in more detail the problem of contextual combinatorial uplifting bandit that we introduced in \cref{subsec:C2UpBandit}.
We extend the model and the algorithm to the case of multiple treatments in \cref{apx:c2-model}.
Subsequently, we specify the algorithm for the case of linear expected payoffs and present some associated analysis in \cref{apx:c2-linear}.
Numerical experiments conducted with the Criteo Uplift Prediction Dataset are presented in \cref{apx:exp-c2}.

\subsection{Model and Algorithm Template}
\label{apx:c2-model}

Let us consider a pool of $\nVariables$ individuals that may change from round to round, associated with (time-dependent) contexts $(\vtv[\feature])_{\indv\in\variables}$ with $\vtv[\feature]\in\R^\vdim$,
and a set of treatments that can be applied to the individuals.
For convenient, we also write $\treatments_0=\treatments\union\{0\}$, where, as before, $0$ indicates that no treatment is applied to an individual.
Following \cref{sec:context},
we consider the action set
$\arms=\setdef{\arm\in(\treatments_0)^{\nVariables}}{\norm{\arm}_0\le\nAffected}$,
which encodes the constraint that at most $\nAffected$ individuals can be treated in each run.
Similarly, we associate each treatment with a function $\va[\objalt][\indtreat]$.
The reward and the uplift of choosing action $\arm$ in round $\run$ are given then by
\begin{equation}
    \notag
    \va[\reward](\vt[\feature]) = \sum_{\allvariables}\va[\objalt][\vv[\arm]](\vtv[\feature]),
    ~~~
    \va[\up{\reward}](\vt[\feature]) = \sum_{\indv\in\va[\variables]}
    \va[\objalt][\vv[\arm]](\vtv[\feature])-\va[\objalt][0](\vtv[\feature]).
\end{equation}
The problem thus consists in finding both the set of customers to treat and the treatments to apply to these customers at each round.
To adapt \CUpUCB to this situation, we define a treatment-dependent uplifting index $\vtav[\UPindex][\run][\indtreat]=\vtav[\UCBindex][\run][\indtreat]-\vtav[\UCBindex][\run][0]$ where $\vtav[\UCBindex][\run][\indtreat]$ and $\vtav[\UCBindex][\run][0]$ are respectively upper confidence bounds on $\va[\objalt][\vv[\arm]](\vtv[\feature])$ and $\va[\objalt][0](\vtv[\feature])$.
Again, if $\va[\objalt][0]$ is known, 
$\vtav[\UCBindex][\run][0]$ can be simply taken as $\va[\objalt][0](\vtv[\feature])$.
Next, at each time step $\run$, we define the presumed optimal treatment for individual $\indv$ and the associated uplifting index as
\begin{equation}
    \label{eq:uplifting-treatment-customer}
    \vtv[\indtreat] = \argmax_{\indtreat\in\treatments}\vtav[\UPindex][\run][\indtreat],
    ~~~~~
    \vtv[\UPindex]
    = \vtav[\UPindex][\run][\vtv[\indtreat]]
    = \max_{\indtreat\in\treatments}\vtav[\UPindex][\run][\indtreat].
\end{equation}
The adapted \CUpUCB algorithm then chooses the $\nAffected$ customers with the largest uplifting index and for each of these customers chooses their presumed optimal treatment. The resulting algorithm (referred to as \CUpUCBMwb) is summarized in \cref{algo:C2UpUCB-mt} .

\begin{algorithm}[tb]
    \caption{\CUpUCBMwb}
    \label{algo:C2UpUCB-mt}
\begin{algorithmic}[1]
    \STATE {\bfseries Input:} Budget constraint $\nAffected$, 
    Parameters for computing the UCBs indices, Treatment set $\treatments_0$
    \FOR{$\run = 1, \ldots, \nRuns$}
    \FOR{$\indv\in\variables$}
    \STATE Compute UCB indices $(\vtav[\UCBindex][\run][\indtreat])_{\indtreat\in\treatments_0}$ for $(\va[\objalt][\indtreat](\vtv[\feature]))_{\indtreat\in\treatments_0}$
    \hfill
    \COMMENT{use \eqref{eq:UCBs-treatment-contextual} for linear payoffs}~~
    \STATE Compute uplifting indices $(\vtav[\UPindex][\run][\indtreat])_{\indtreat\in\treatments}$ 
    \hfill
    \COMMENT{$\vtav[\UPindex][\run][\indtreat]=\vtav[\UCBindex][\run][\indtreat]-\vtav[\UCBindex][\run][0]$}~~
    \STATE Compute presumed optimal treatment $\vtv[\indtreat]$ and uplifting index $\vtv[\UPindex]$ 
    \hfill
    \COMMENT{use \eqref{eq:uplifting-treatment-customer}}~~
    \ENDFOR
    \STATE Set $\vt[\variables]\subs\argmax_{\set\subset\variables,\card(\set)\le\nAffected}\sum_{\indv\in\set}\vtv[\UPindex]$
    \hfill
    \COMMENT{this defines the set of treated customers}~~
    \STATE Select action $\vt[\arm]\subs(\vtv[\indtreat]\one\{\indv\in\vt[\variables]\})_{\allvariables}$
    \hfill
    \COMMENT{$\vtv[\arm]=\vtv[\indtreat]$ if $\indv\in\vt[\variables]$; $\vtv[\arm]=0$ otherwise}~~
    \ENDFOR
\end{algorithmic}
\end{algorithm}

\subsection{Linear Expected Payoffs}
\label{apx:c2-linear}

To provide a concrete algorithm and its analysis thereof, let us assume that $\va[\objalt][\indtreat]$ is linear in the feature vector with some unknown coefficient vector $\va[\param][\indtreat]$, \ie $\va[\objalt][\indtreat](\vtv[\feature])=\product{\va[\param][\indtreat]}{\vtv[\feature]}$.
Then, we can construct upper confidence bounds for $\va[\objalt][\indtreat](\vtv[\feature])$ following the ideas of linear UCB \citep{APS11}.
To begin, for each $\indtreat\in\treatments_0$ and $\run\in\intinterval{1}{\nRuns}$, we define
\begin{equation}
    \notag
    \vta[\variables][\run][\indtreat]
    = \setdef{\indv\in\variables}{\vtv[\arm]=\indtreat}
\end{equation}
as the set of customers that receive treatment $\indtreat$ in round $\run$.
Fix a regularization parameter $\regpar>0$, we further define the following $\vdim\times\vdim$ matrix and $\vdim$-dimensional vector
\begin{equation}
    \label{eq:definition-Vtz}
    \vta[\LSmat][\run][\indtreat] = \regpar\Id + \sum_{\runalt=1}^{\run}\sum_{\indv\in\vta[\variables][\runalt][\indtreat]}
    \vtv[\feature][\runalt]\vtv[\feature][\runalt]^{\top},
    ~~~~~
    \vta[\LSvec][\run][\indtreat]
    = \sum_{\runalt=1}^{\run}\sum_{\indv\in\vta[\variables][\runalt][\indtreat]}
    \vtv[\valuev][\runalt]\vtv[\feature][\runalt].
\end{equation}
The regularized least-square estimate of $\va[\param][\indtreat]$ is then given by $\vta[\est{\param}][\run][\indtreat]= (\vta[\LSmat][\run][\indtreat])^{-1}\vta[\LSvec][\run][\indtreat]$.
The UCB index is of the form
\begin{equation}
    \label{eq:UCBs-treatment-contextual}
    \vtav[\UCBindex][\run][\indtreat] = \product{\vta[\est{\param}][\run-1][\indtreat]}{\vtv[\feature]}
    +\vt[\radiusEllipConf]\norm{\vtv[\feature]}_{(\vta[\LSmat][\run-1][\indtreat])^{-1}}.
\end{equation}
for some $\vt[\radiusEllipConf]$ that is properly chosen.

\paragraph{Regret Guarantee.}
In this part, we provide regret analysis for \cref{algo:C2UpUCB-mt} when
\begin{enumerate*}[\itshape i\upshape)]
\item the baseline payoffs are unknown,
\item the expected payoffs are linear, and
\item the UCB indices are computed following \eqref{eq:UCBs-treatment-contextual}.
\end{enumerate*}
For technical reasons, we will assume the noises $\vtv[\snoise]$ to be mutually independent (here $\vtv[\snoise]=\vtv[\valuev]-\product{\va[\param][\indtreat]}{\vtv[\feature]}$).
We comment on this assumption in \cref{rem:independence-noise}.

The theorem below establishes a high-probability regret bound for \CUpUCBMwb when the above conditions are satisfied.

\begin{theorem}
\label{thm:C2UpUCB-regret}
Assume that $\norm{\vtv[\feature]}\le1$ for all $\run$ and $\indv$ and $(\vtv[\snoise])_{\allvariables}$ are mutually independent and $1$-sub-Gaussian (conditioning on the past).
Let $\parambound>0$ such that $\norm{\va[\param][\indtreat]}\le\parambound$ for all $\indtreat\in\treatments_0$.
Then if \CUpUCBMwb (\cref{algo:C2UpUCB-mt}) is run with the UCB indices defined in \eqref{eq:UCBs-treatment-contextual} under the parameter choices
\begin{equation}
    \notag
    \vt[\radiusEllipConf]
    =\sqrt{\regpar}\parambound
    +\sqrt{2\log\left(\frac{\nTreatments+1}{\smallproba}\right)
    +\vdim\log\left(1+\frac{\nVariables\run}{\vdim\regpar}\right)},
\end{equation}
and $\regpar\ge\nAffected$, it holds, with probability at least $1-\smallproba$, that
\begin{equation}
    \label{eq:C2UpUCB-regret}
    \vt[\reg][\nRuns]
    \le
    4
    \sqrt{
    \vdim\nAffected(\nTreatments+1)\nRuns
    \log\left(1+\frac{\nAffected\nRuns}{\vdim\regpar}\right)
    }
    \left(
    \sqrt{\regpar}\parambound
    +\sqrt{2\log\left(\frac{\nTreatments+1}{\smallproba}\right)
    +\vdim\log\left(1+\frac{\nVariables\nRuns}{\vdim\regpar}\right)}
    \right)
    .
\end{equation}
\end{theorem}

\cref{thm:C2UpUCB-regret} provides a regret bound in $\tbigoh(\sqrt{\max(\vdim,\nAffected)\vdim\nTreatments\nAffected\nRuns})$.
We observe that the dependence on $\nVariables$ is only logarithmic.
In fact, $\nVariables$ plays no role in either the number of parameters to estimate (which is $(\nTreatments+1)\vdim$) or the scale of the noise (which is in $\bigoh(\sqrt{\nAffected})$).
Ideally, we would like to remove completely the dependence on $\nVariables$.
However, we are not able to achieve this in our proof, and we leave whether this is possible or not as an open question.
Finally, due the mutual independence of noises, the regret only scales with $\sqrt{\nAffected}$ as long as $\nAffected\le\vdim$.

\begin{remark}
\label{rem:independence-noise}
The $\log(\nVariables)$ factor appears because $\vta[\est{\param}][\run][0]$ is estimated using feedback from roughly $\nVariables\nRuns$ samples (see \cref{lem:conf-ellip} below).
This is also the reason why we need mutual independence of the noises, since otherwise, this estimation may cause the regret to get $\sqrt{\nVariables}$ larger.
It can be easily verified that the dependence on $\nVariables$ can be completely removed when the baseline payoffs are known and used for computing the uplifting indices.
In this case, we can also show if the noises are not mutually independent, multiplying the second term of $\vt[\radiusEllipConf]$ by $\sqrt{\nAffected}$ still guarantees a regret in $\tbigoh(\vdim\nAffected\sqrt{\nTreatments\nRuns})$.
What is at stake here is \cref{lem:conf-ellip} that we present below, as the size of the confidence ellipsoid varies for different underlying assumptions.
\end{remark}

\paragraph{Analysis\afterhead}
To prove \cref{thm:C2UpUCB-regret}, we need to show that $\vtav[\UCBindex]$ as defined in \eqref{eq:UCBs-treatment-contextual} is indeed an upper confidence bound on $\product{\va[\param][\indtreat]}{\vtv[\feature]}$.
This step is based on the following lemma which establishes the confidence ellipsoids of the estimated parameters.
In fact, $\vtav[\UCBindex]$ defined in \eqref{eq:UCBs-treatment-contextual} can be regarded as the largest estimated payoff that one can get from choosing a parameter from the confidence ellipsoid.

\begin{lemma}
\label{lem:conf-ellip}
Assume that $\norm{\vtv[\feature]}\le1$ for all $\run$ and $\indv$ and $(\vtv[\snoise])_{\allvariables}$ are mutually independent and $1$-sub-Gaussian (conditioning on the past).
For all $\indtreat\in\treatments_0$ and $\alt{\smallproba}\in(0,1)$, with probability at least $1-\alt{\smallproba}$ it holds for all $\run$ that
\begin{equation}
    \notag
    \begin{aligned}
    \norm{\va[\param][\indtreat]-\vta[\est{\param}][\run][\indtreat]}_{\vta[\LSmat][\run][\indtreat]}
    &\le
    \sqrt{\regpar}\norm{\va[\param][\indtreat]}
    + \sqrt{2\log\left(\frac{1}{\alt{\smallproba}}\right)
    +\log\left(\frac{\det(\vta[\LSmat][\run][\indtreat])}{\regpar^{\vdim}}\right)}\\
    &\le
    \sqrt{\regpar}\norm{\va[\param][\indtreat]}
    +\sqrt{2\log\left(\frac{1}{\alt{\smallproba}}\right)
    +\vdim\log\left(1+\frac{\nVariables\run}{\vdim\regpar}\right)}.
    \end{aligned}
\end{equation}
\end{lemma}
\begin{proof}
This is an immediate corollary of \cite[Th. 2]{APS11}.
Since $(\vtv[\snoise])_{\allvariables}$ are mutually independent, we can apply this theorem by arranging $(\vtv[\snoise])_{\indv\in\vta[\variables][\run][\indtreat]}$ in some arbitrary order $\vtv[\snoise][\run][\indv_1],\ldots,\vtv[\snoise][\run][\indv_{\card(\vta[\variables][\run][\indtreat])}]$ so that $\vtv[\snoise][\run][\indv_{\indg}]$ is $1$-sub-Gaussian conditioning on $(\bigcup_{\runalt<\run}\setdef{\vtv[\snoise][\runalt]}{\indv\in\vta[\variables][\runalt][\indtreat]})\union\{\vtv[\snoise][\run][\indv_{\indg'}]\}_{1\le \indg'\le \indg}$.

For the second inequality, we notice that $\tr(\vta[\LSmat][\run][\indtreat])\le\vdim\regpar+\sum_{\runalt=1}^{\run}\sum_{\indv\in\vta[\variables][\runalt][\indtreat]}\norm{\vtv[\feature]}^2\le\vdim\regpar+\nVariables\run$.
It thus follows from the AM–GM inequality on the eigenvalues of $\vta[\LSmat][\run][\indtreat]$, $\det(\vta[\LSmat][\run][\indtreat])\le(\tr(\vta[\LSmat][\run][\indtreat])/\vdim)^\vdim$.
\end{proof}

The lemma below will also be useful for bounding the regret.

\begin{lemma}
\label{lem:Vt-norm-det-bound}
Consider subsets of the variables $\vt[\variables][1],\ldots,\vt[\variables][\nRuns]\subset\variables$ and define $\vt[\LSmat]=\regpar\Id+\sum_{\runalt=1}^{\run}\sum_{\indv\in\vt[\variables][\runalt]}\vtv[\feature][\runalt]\vtv[\feature][\runalt]^{\top}$.
If $\regpar\ge\max_{\run\in\oneto{\nRuns}}\card(\vt[\variables])$ and $\norm{\vtv[\feature]}\le1$ for all $\run$ and $\indv$, it holds that
\begin{equation}
    \notag
    \sum_{\run=1}^{\nRuns}
    \sum_{\indv\in\vt[\variables]}\norm{\vtv[\feature]}_{\vt[\LSmat][\run-1]^{-1}}^2
    \le 2\log\left(\frac{\det(\vt[\LSmat][\nRuns])}{\regpar^{\vdim}}\right)
    \le
    2\vdim\log\left(
    1+\frac{\sum_{\run=1}^{\nRuns}\card(\vt[\variables])}{\vdim\regpar}
    \right)
    .
\end{equation}
\end{lemma}
\begin{proof}
For all $\run\ge1$, we have
\begin{equation}
    \notag
    \vt[\LSmat][\run]
    =\vt[\LSmat][\run-1]+\sum_{\indv\in\vt[\variables]}\vtv[\feature]\vtv[\feature]^{\top}
    =\vt[\LSmat][\run-1]^{1/2}
    \left(\Id + \sum_{\indv\in\vt[\variables]}\vt[\LSmat][\run-1]^{-1/2}\vtv[\feature]\vtv[\feature]^{\top}\vt[\LSmat][\run-1]^{-1/2}\right)
    \vt[\LSmat][\run-1]^{1/2}
\end{equation}
Thus, with $\vtv[\wvec]=\vt[\LSmat][\run-1]^{-1/2}\vtv[\feature]$, we get
\begin{equation}
    \label{eq:det-decomp}
    \det(\vt[\LSmat][\run])
    = \det(\vt[\LSmat][\run-1])
    \det\left(\Id+\sum_{\indv\in\vt[\variables]}\vtv[\wvec]\vtv[\wvec]^{\top}\right).
\end{equation}
Regarding $\{\vtv[\wvec]\}_{\indv\in\vt[\variables]}$ as column vectors, we denote by their horizontal concatenation as $\wmat\in\R^{\vdim\times\vt[\nVariables]}$, where $\vt[\nVariables]=\card(\vt[\variables])$.
We have $\wmat\wmat^{\top}=\sum_{\indv\in\vt[\variables]}\vtv[\wvec]\vtv[\wvec]^{\top}$.
Let the eigenvalues of $\wmat\wmat^{\top}$ be denoted by $\eigv_1,\ldots,\eigv_{\vdim}$, then clearly $\det(\Id+\wmat\wmat^{\top})=\prod_{\indg=1}^{\vdim}(1+\eigv_{\indg})$.
On the other hand, $\sum_{\indg=1}^{\vdim}\eigv_{\indg}=\tr(\wmat\wmat^{\top})=\tr(\wmat^{\top}\wmat)=\sum_{\indv\in\vt[\variables]}\norm{\vtv[\wvec]}^2$. From the non-negativity of the eigenvalues of $\wmat\wmat^{\top}$, we obtain
\begin{equation}
    \label{eq:det-inequa}
    \det(\Id+\wmat\wmat^{\top})=\prod_{\indg=1}^{\vdim}(1+\eigv_{\indg})
    \ge 1 + \sum_{\indg=1}^{\vdim}\eigv_{\indg}
    = 1 + \sum_{\indv\in\vt[\variables]}\norm{\vtv[\wvec]}^2
    = 1 + \sum_{\indv\in\vt[\variables]}\norm{\vtv[\feature]}_{\vt[\LSmat][\run-1]^{-1}}^2.
\end{equation}
Applying \eqref{eq:det-decomp} and \eqref{eq:det-inequa} recursively leads to
\begin{equation}
    \label{eq:det-inequa-recur}
    \det(\vt[\LSmat][\nRuns])
    \ge \det(\regpar\Id)
    \prod_{\run=1}^{\nRuns}
    \left(1 + \sum_{\indv\in\vt[\variables]}\norm{\vtv[\feature]}_{\vt[\LSmat][\run-1]^{-1}}^2\right)
    =
    \regpar^{\vdim}
    \prod_{\run=1}^{\nRuns}
    \left(1 + \sum_{\indv\in\vt[\variables]}\norm{\vtv[\feature]}_{\vt[\LSmat][\run-1]^{-1}}^2\right).
\end{equation}

We now turn back to bound $\sum_{\indv\in\vt[\variables]}\norm{\vtv[\feature]}_{\vt[\LSmat][\run-1]^{-1}}^2$.
Since $\regpar\ge\max_{\run\in\oneto{\nRuns}}\card(\vt[\variables])$ and $\norm{\vtv[\feature]}\le1$, we have
\begin{equation}
    \label{eq:sum-log-bound}
    \sum_{\indv\in\vt[\variables]}\norm{\vtv[\feature]}_{\vt[\LSmat][\run-1]^{-1}}^2
    = \min\left(1, \sum_{\indv\in\vt[\variables]}\norm{\vtv[\feature]}_{\vt[\LSmat][\run-1]^{-1}}^2\right)
    \le 2\log\left(1+ \sum_{\indv\in\vt[\variables]}\norm{\vtv[\feature]}_{\vt[\LSmat][\run-1]^{-1}}^2\right).
\end{equation}
Combining \eqref{eq:det-inequa-recur} and \eqref{eq:sum-log-bound} we get the first inequality.
The second inequality is proved as in \cref{lem:conf-ellip}.
\end{proof}

We are now ready to prove \cref{thm:C2UpUCB-regret}.
The first part of the proof reuses the trick that we have seen several times before, notably in the proof of \cref{thm:UpUCB-regret}.
We rearrange the UCB indices and show that the suboptimality gaps can be bounded from above by the sum of the radii of certain confidence intervals. The details will be omitted.
Next, we apply \cref{lem:Vt-norm-det-bound} to bound the sum of these upper bounds and conclude.

\begin{proof}[Proof of \cref{thm:C2UpUCB-regret}]
For any $\indtreat\in\treatments_0$,
choosing $\alt{\smallproba}=\smallproba/(\nTreatments+1)$ in \cref{lem:conf-ellip} gives 
\[
\prob\left(\{\exists \run, \norm{\va[\param][\indtreat]-\vta[\est{\param}][\run][\indtreat]}^2_{\vta[\LSmat][\run][\indtreat]}>\vt[\radiusEllipConf]\right)\ge \frac{\smallproba}{\nTreatments+1}.
\]
With a union bound we thus deduce that with probability at least $1-\smallproba$, the inequality $\norm{\va[\param][\indtreat]-\vta[\est{\param}][\run][\indtreat]}^2_{\vta[\LSmat][\run][\indtreat]}\le\vt[\radiusEllipConf]$ holds for all $\run$ and $\indtreat$.
This also implies that for all $\run$ and $\indtreat$, we have
\begin{equation}
    \label{eq:contextual-estimate-product-diff}
    \abs{
    \product{\va[\param][\indtreat]}{\vtv[\feature]}
    -\product{\vta[\est{\param}][\run-1][\indtreat]}{\vtv[\feature]}}
    \le \vt[\radiusEllipConf]\norm{\vtv[\feature]}_{(\vta[\LSmat][\run-1][\indtreat])^{-1}}
\end{equation}
In the following, we will prove \eqref{eq:C2UpUCB-regret} under the condition that \eqref{eq:contextual-estimate-product-diff} holds, which happens with probability at least $1-\smallproba$ as we just showed.

Let $\run\ge 1$ and $\vt[\sol[\arm]]\in\argmax_{\arm\in\arms} \sum_{\allvariables}\product{\va[\param][\vv[\arm]]}{\vtv[\feature]}$ be an optimal action of this round.
The algorithm chooses $\vt[\arm]$ that maximizes $\vta[\UPindex]=\sum_{\indv\in\va[\variables]}(\vtav[\UCBindex]-\vtav[\UCBindex][\run][0])$, and thus
\begin{equation}
    \label{eq:contextual-pf-UCB-diff}
    \sum_{\indv\in\va[\variables][\vt[\arm]]}(\vtav[\UCBindex][\run][\vtv[\arm]]-\vtav[\UCBindex][\run][0])
    \ge\sum_{\indv\in\va[\variables][\vt[\sol[\arm]]]}(\vtav[\UCBindex][\run][\vtv[\sol[\arm]]]-\vtav[\UCBindex][\run][0]).
\end{equation}
On the other hand, inequality \eqref{eq:contextual-estimate-product-diff} can be rewritten as
\begin{equation}
    \label{eq:contextual-pf-UCB-lu}
    \vtav[\UCBindex][\run][\indtreat]
    -2\vt[\radiusEllipConf]\norm{\vtv[\feature]}_{(\vta[\LSmat][\run-1][\indtreat])^{-1}}
    \le
    \product{\va[\param][\indtreat]}{\vtv[\feature]}
    \le\vtav[\UCBindex][\run][\indtreat].
\end{equation}
Combining \eqref{eq:contextual-pf-UCB-diff} and \eqref{eq:contextual-pf-UCB-lu} and rearranging, we get
\begin{equation}
    \notag
    \sum_{\indv\in\variables}\product{\va[\param][\vtv[\sol[\arm]]]-\va[\param][\vtv[\arm]]}{\vtv[\feature]}
    =
    \sum_{\indv\in\va[\variables][\vt[\arm]]\union \va[\variables][\vt[\sol[\arm]]]}
    \product{\va[\param][\vtv[\sol[\arm]]]-\va[\param][\vtv[\arm]]}{\vtv[\feature]}
    \le
    \sum_{\indv\in\va[\variables][\vt[\arm]]\union \va[\variables][\vt[\sol[\arm]]]}
    2\vt[\radiusEllipConf]\norm{\vtv[\feature]}_{(\vta[\LSmat][\run-1][\vtv[\arm]])^{-1}}.
\end{equation}
Summing the above from $\run=1$ to $\nRuns$ and applying the Cauchy–Schwarz inequality, we obtain
\begin{equation}
    \notag
    \begin{aligned}
    \vt[\reg][\nRuns]
    &\le\sum_{\run=1}^{\nRuns}
    \sum_{\indv\in\va[\variables][\vt[\arm]]\union \va[\variables][\vt[\sol[\arm]]]}
    2\vt[\radiusEllipConf]\norm{\vtv[\feature]}_{(\vta[\LSmat][\run-1][\vtv[\arm]])^{-1}}\\
    &\le
    2\vt[\radiusEllipConf][\nRuns]
    \sqrt{
    \left(\sum_{\run=1}^{\nRuns}
    \card(\va[\variables][\vt[\arm]]\union \va[\variables][\vt[\sol[\arm]]])\right)
    \left(\sum_{\run=1}^{\nRuns}
    \sum_{\indv\in\va[\variables][\vt[\arm]]\union \va[\variables][\vt[\sol[\arm]]]}
    \norm{\vtv[\feature]}_{(\vta[\LSmat][\run-1][\vtv[\arm]])^{-1}}^2\right)
    }
    \\
    &\le
    2\vt[\radiusEllipConf][\nRuns]
    \sqrt{2\nAffected\nRuns
    \sum_{\run=1}^{\nRuns}
    \left(
    \sum_{\indtreat\in\treatments}\sum_{\indv\in\vta[\variables][\run][\indtreat]}
    \norm{\vtv[\feature]}_{(\vta[\LSmat][\run-1][\indtreat])^{-1}}^2
    +
    \sum_{\indv\in\setexclude{\va[\variables][\vt[\sol[\arm]]]}{\va[\variables][\vt[\arm]]}}
    \norm{\vtv[\feature]}_{(\vta[\LSmat][\run-1][0])^{-1}}^2
    \right)}.\\
    \end{aligned}
\end{equation}
Since $\regpar\ge\nAffected$, by \cref{lem:Vt-norm-det-bound}, for all $\indtreat\in\treatments$ it holds that
\begin{equation}
    \notag
    \sum_{\run=1}^{\nRuns}\sum_{\indv\in\vta[\variables][\run][\indtreat]}
    \norm{\vtv[\feature]}_{(\vta[\LSmat][\run-1][\indtreat])^{-1}}^2
    \le
    2\vdim\log\left(
    1+\frac{\nAffected\nRuns}{\vdim\regpar}\right).
\end{equation}
Similarly, let us define $\vta[\widetilde{\LSmat}][\run][0]=\regpar\Id+\sum_{\runalt=1}^{\run}\sum_{\indv\in\setexclude{\va[\variables][\vt[\sol[\arm]]]}{\va[\variables][\vt[\arm]]}}\vtv[\feature][\runalt]\vtv[\feature][\runalt]^{\top}$.
Using $\vta[\widetilde{\LSmat}][\run][0]\mleq\vta[\LSmat][\run][0]$ and invoking \cref{lem:Vt-norm-det-bound} with $\vt[\variables]\subs\setexclude{\va[\variables][\vt[\sol[\arm]]]}{\va[\variables][\vt[\arm]]}$ gives
\begin{equation}
    \notag
    \sum_{\run=1}^{\nRuns}
    \sum_{\indv\in\setexclude{\va[\variables][\vt[\sol[\arm]]]}{\va[\variables][\vt[\arm]]}}
    \norm{\vtv[\feature]}_{(\vta[\LSmat][\run-1][0])^{-1}}^2
    \le
    \sum_{\run=1}^{\nRuns}
    \sum_{\indv\in\setexclude{\va[\variables][\vt[\sol[\arm]]]}{\va[\variables][\vt[\arm]]}}
    \norm{\vtv[\feature]}_{(\vta[\widetilde{\LSmat}][\run-1][0])^{-1}}^2
    \le
    2\vdim\log\left(
    1+\frac{\nAffected\nRuns}{\vdim\regpar}\right).
\end{equation}
Putting the above three inequalities together, we get exactly \eqref{eq:C2UpUCB-regret}.
\end{proof}

\subsection{Experiments for Targeted Campaign with Criteo Uplifting Dataset}
\label{apx:exp-c2}

In this section, we provide experiments for the contextual combinatorial uplifting model using the Criteo Uplift dataset.
At each round, $100$ individuals are sampled from the dataset and the problem consists in choosing $10$ of them to treat so that the number of visits over the $100$ individuals is maximized.  Here, we compare our methods based on \CUpUCB (\cref{algo:C2UpUCB}) with approaches based on greedy and $\varepsilon$-greedy (see~\citep{LS20} for a description) that are more appropriate baselines in this setup. 
Again, our results are averaged over $100$ independent runs and the shaded areas of the plots represent the standard errors.



\paragraph{Algorithms\afterhead} 
In all the considered algorithms, the learner learns a logistic model based on the feedback they have received from the individuals they select to treat.\!\footnote{Logistic models were also considered in the original Criteo dataset paper \citep{Diemert2018} to compute uplifts.}
If the baseline model is unknown, the learner also learns a baseline logistic model based on the feedback received from the individuals they do not treat.
In the following, we will denote respectively by $\vta[\est{\objalt}][\run][1]$ and $\vta[\est{\objalt}][\run][0]$ the conditional probability functions of the two fitted models using data up to and including round $\run$.
When the baseline model is known, we denote by $\va[\objalt][0]$ its conditional probability function.
The methods that we look into in this experiment mainly differ in how they select individuals in each round $\run$, as we detail below.
As before, we use the suffix (bl) to indicate modification of the respective algorithm to the case where the baseline payoffs are known.
\begin{itemize}[leftmargin=*]
    \item \textbf{Greedy\,(bl):} The learner chooses the $10$ individuals with the largest estimated uplifts $\vta[\est{\objalt}][\run-1][1](\feature)-\va[\objalt][0](\feature)$.
    \item $\boldsymbol{\varepsilon}$\textbf{-greedy\,(bl):} The learner reserves a exploration budget $b$ where $b$ is sampled from $B(10, \varepsilon)$ the binomial distribution with parameters $10$ and $\varepsilon$.
    $10-b$ individuals are chosen greedily as in the greedy strategy while the remaining $b$ individuals are randomly sampled from those that have not been selected.
    \item $\boldsymbol{\varepsilon}$\textbf{-greedy:} For the greedy selection part in $\varepsilon$-greedy, the estimated uplifts are computed by $\vta[\est{\objalt}][\run-1][1](\feature)-\vta[\est{\objalt}][\run-1][0](\feature)$.
    \item \textbf{\CUpUCB(bl):} For each individual $\indv$ we define their uplifting index
    $\vtav[\UPindex]=\vta[\est{\objalt}][\run-1][1](\feature)+\expparam\norm{\vtv[\feature]}_{(\vta[\LSmat][\run-1][1])^{-1}}-\va[\objalt][0](\feature)$ where $\vta[\LSmat][\run-1][1]$ is defined following \eqref{eq:definition-Vtz} with $\regpar=1$ and $\expparam$ is a exploration parameter to be tuned.
    The algorithm consists in choosing the $10$ individuals with the largest uplifting indices.
    \item \textbf{\CUpUCB:} It works in the same way as C2UPUCB with baseline knowledge but the uplifting index is now defined by 
    $\vtav[\UPindex]=\vta[\est{\objalt}][\run-1][1](\feature)+\expparam\norm{\vtv[\feature]}_{(\vta[\LSmat][\run-1][1])^{-1}}
    -\vta[\est{\objalt}][\run-1][0](\feature)-\expparam\norm{\vtv[\feature]}_{(\vta[\LSmat][\run-1][0])^{-1}}$;
    that is, it is computed as the difference between two upper confidence bounds.
    \item \textbf{C2UCB:} It works in the same way as \CUpUCB but the baseline is ignored, so $\vtav[\UPindex]=\vta[\est{\objalt}][\run-1][1](\feature)+\expparam\norm{\vtv[\feature]}_{(\vta[\LSmat][\run-1][1])^{-1}}$. The linear variant of this algorithm was introduced in \citep{QCZ14} for contextual combinatorial bandits.
\end{itemize}
Note that as the reward comes from a logistic model, we use a logistic regression for the bandit model based on the idea of GLM-UCB \citep{filippi2010parametric}.
For the results, we select the optimal parameters from $\varepsilon\in\{0.05, 0.1, 0.2, 0.3\}$ and $\expparam\in\{0.5, 1, 2, 3\}$.
More details on the tuning of the parameters are provided in \cref{apx:param-tuning}, and the chosen optimal parameters are reported in \cref{tab:params}.

\paragraph{Data Preprocessing\afterhead}
For this experiment, we normalize the data so that the features are contained in the range $[0, 1]$.
Moreover, we also subsample the dataset to increase the visit rate to $0.25$.
In fact, we found the logistic models to perform very poorly on the original dataset (with a f1-score around $0.4$) due to its small visit rate at $0.04$.

\begin{figure}[t]
    \centering
    \begin{subfigure}[b]{\linewidth}
    \centering
    \includegraphics[width=0.325\linewidth]{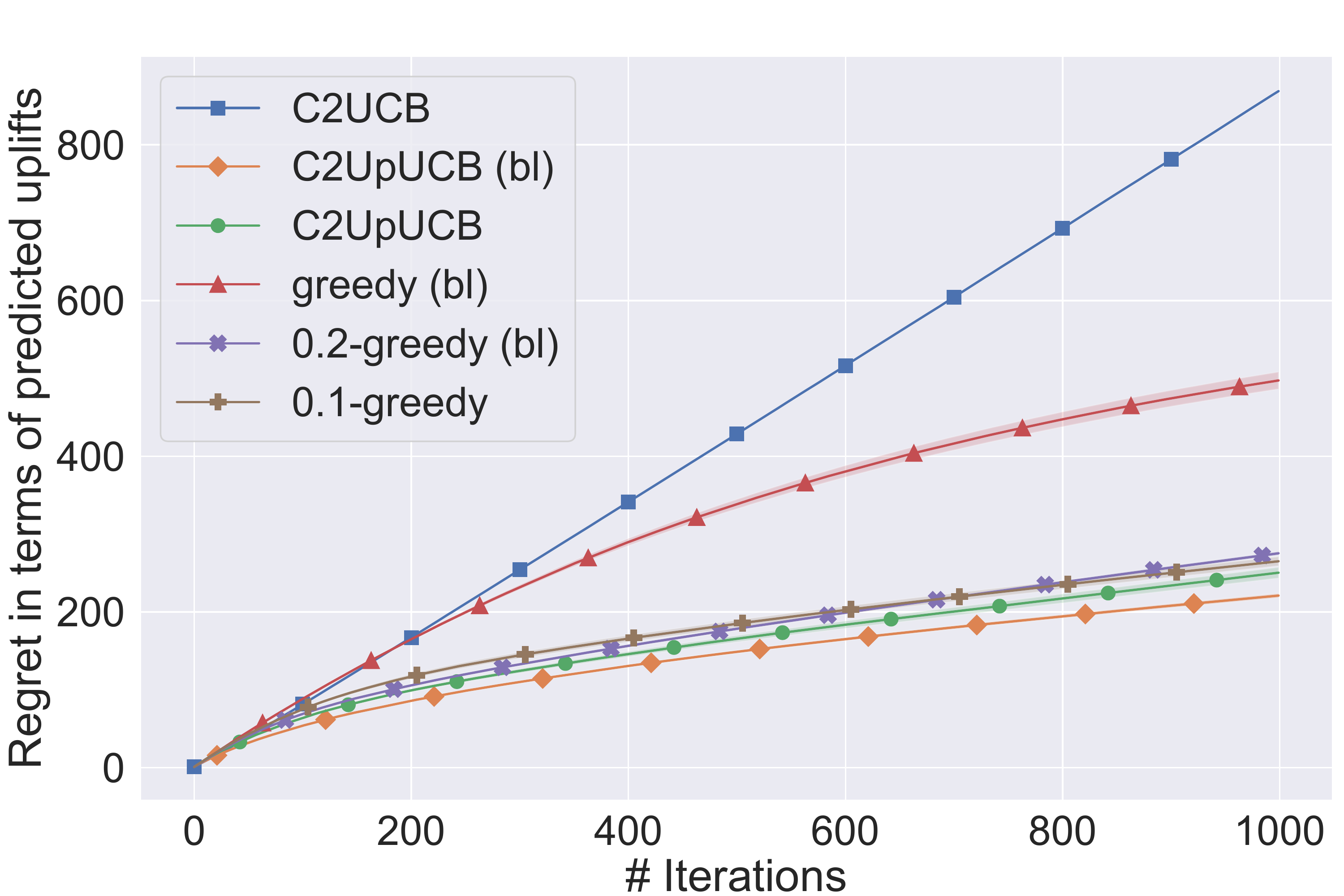}
    \includegraphics[width=0.325\linewidth]{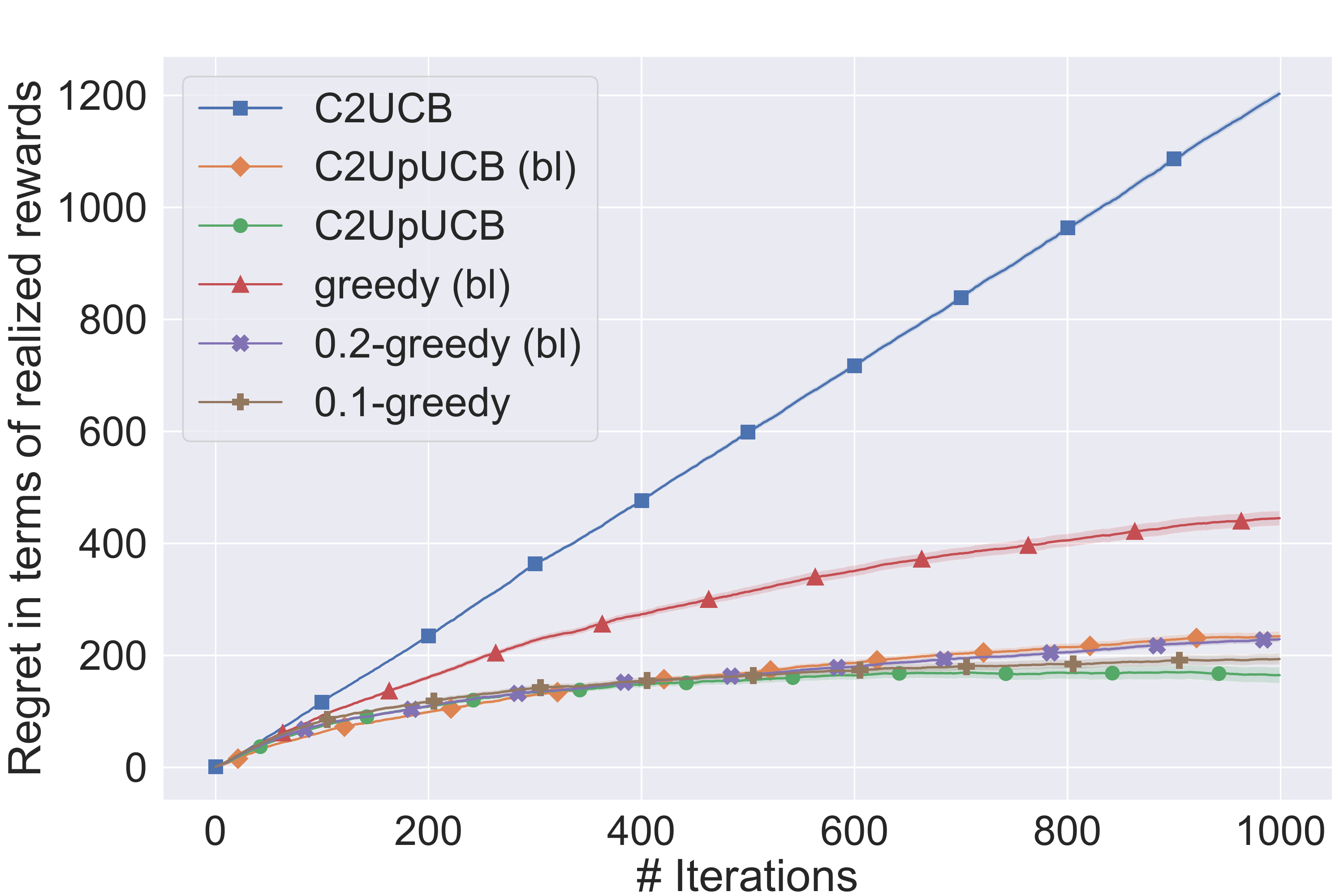}
    \includegraphics[width=0.325\linewidth]{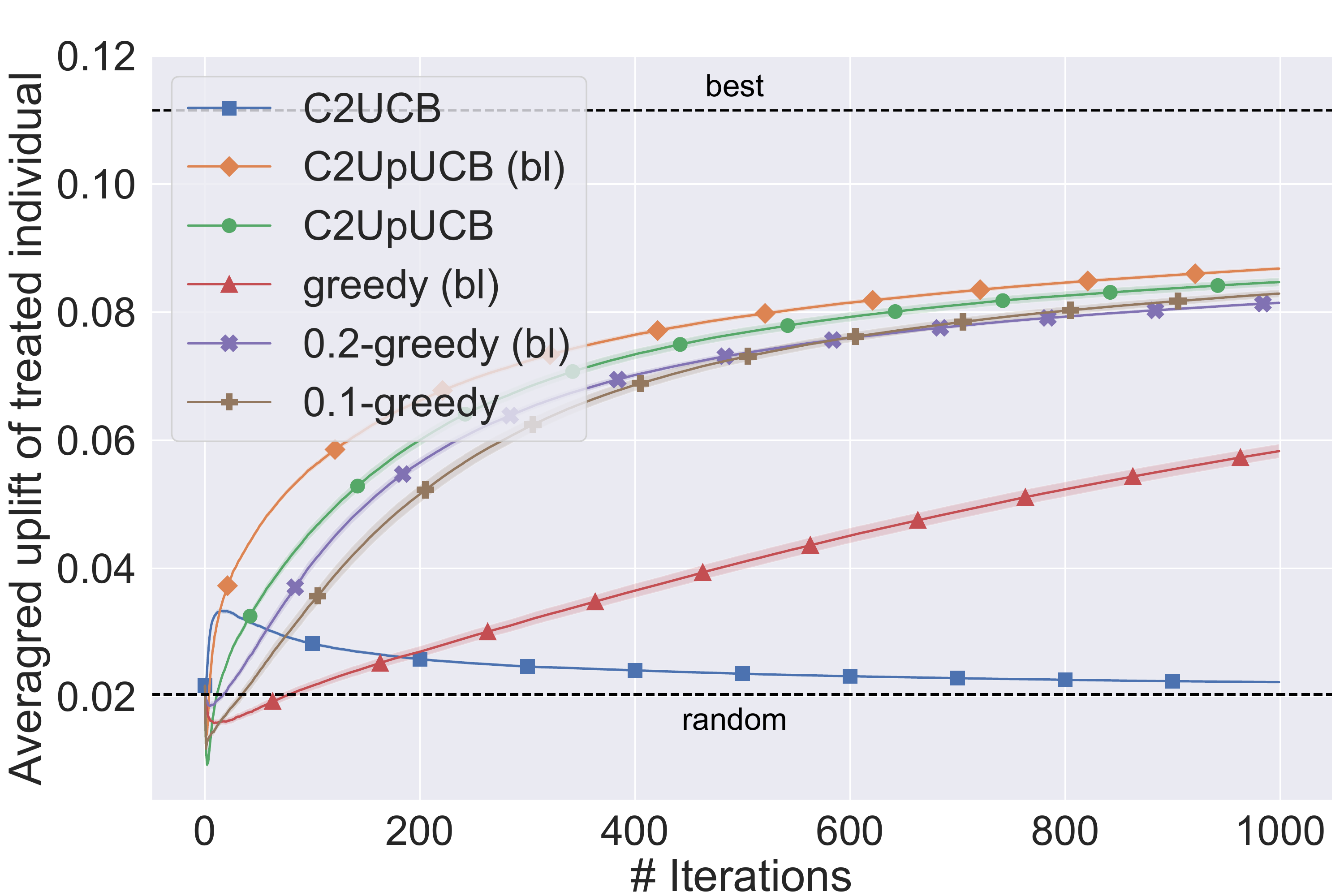}
    \caption{Logistic models only used to generate counterfactual outcomes.}
    \label{subfig:exp-c2-add-with-true}
    \end{subfigure}\\
    \vspace*{0.6em}
    \begin{subfigure}[b]{\linewidth}
    \centering
    \includegraphics[width=0.325\linewidth]{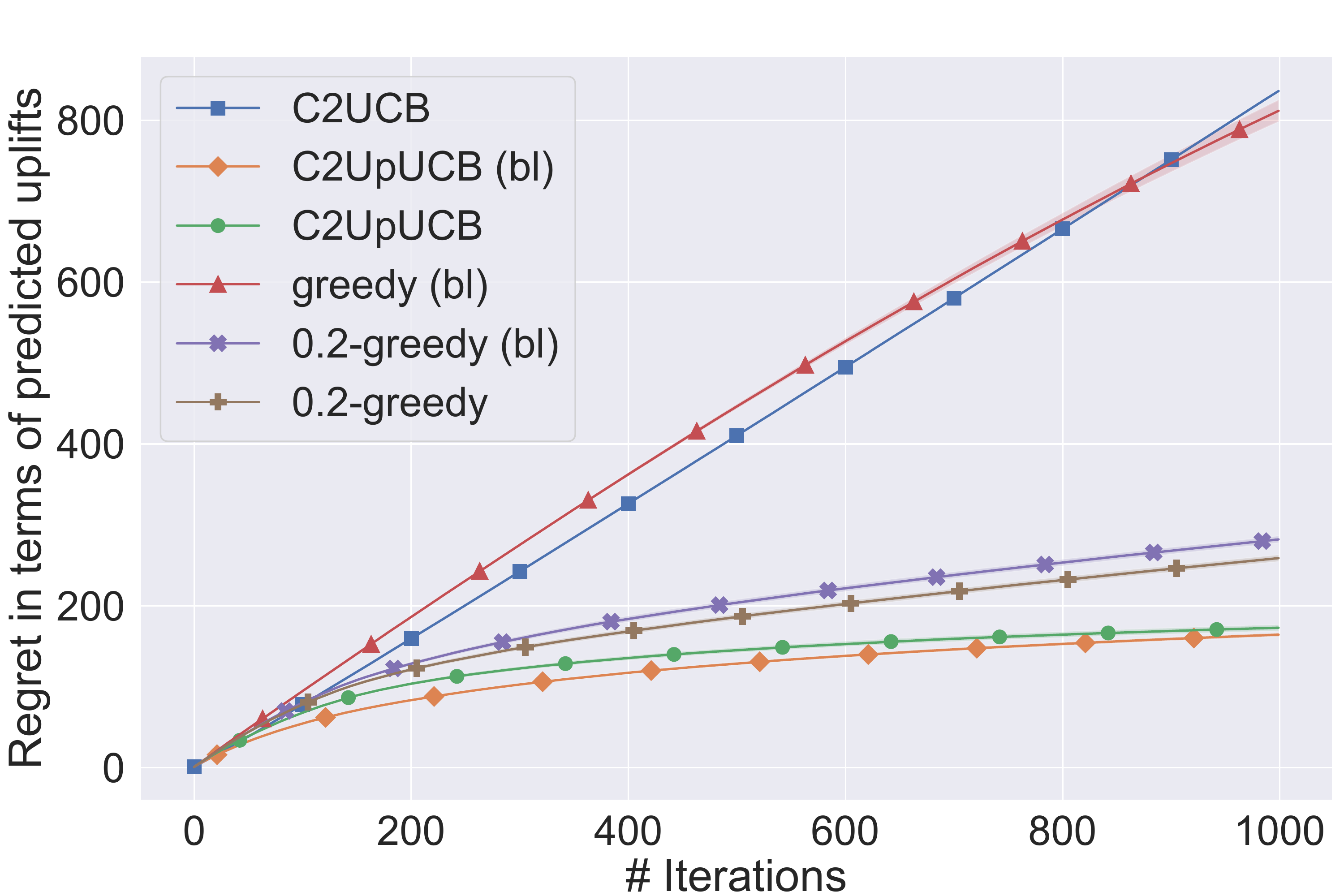}
    \includegraphics[width=0.325\linewidth]{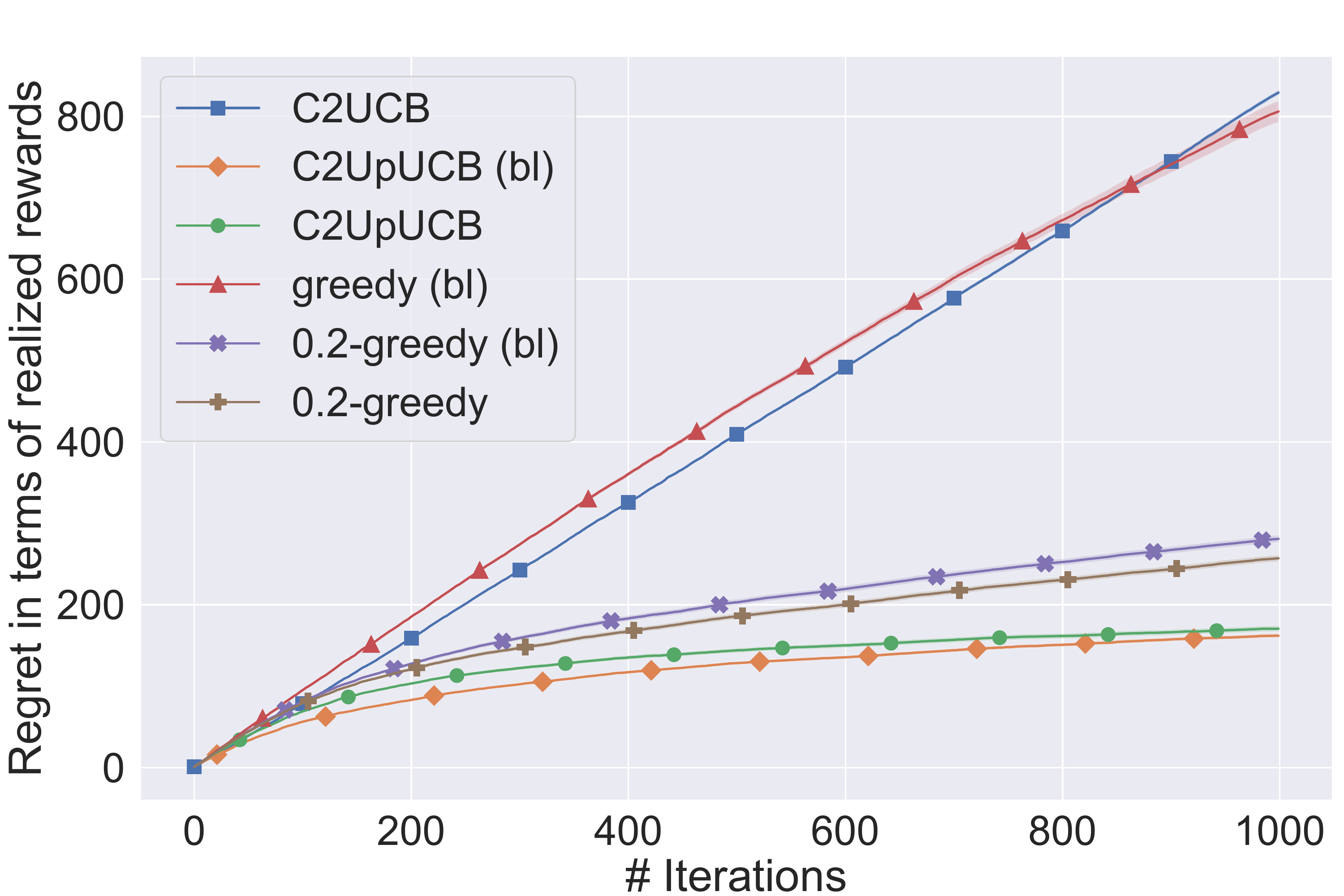}
    \includegraphics[width=0.325\linewidth]{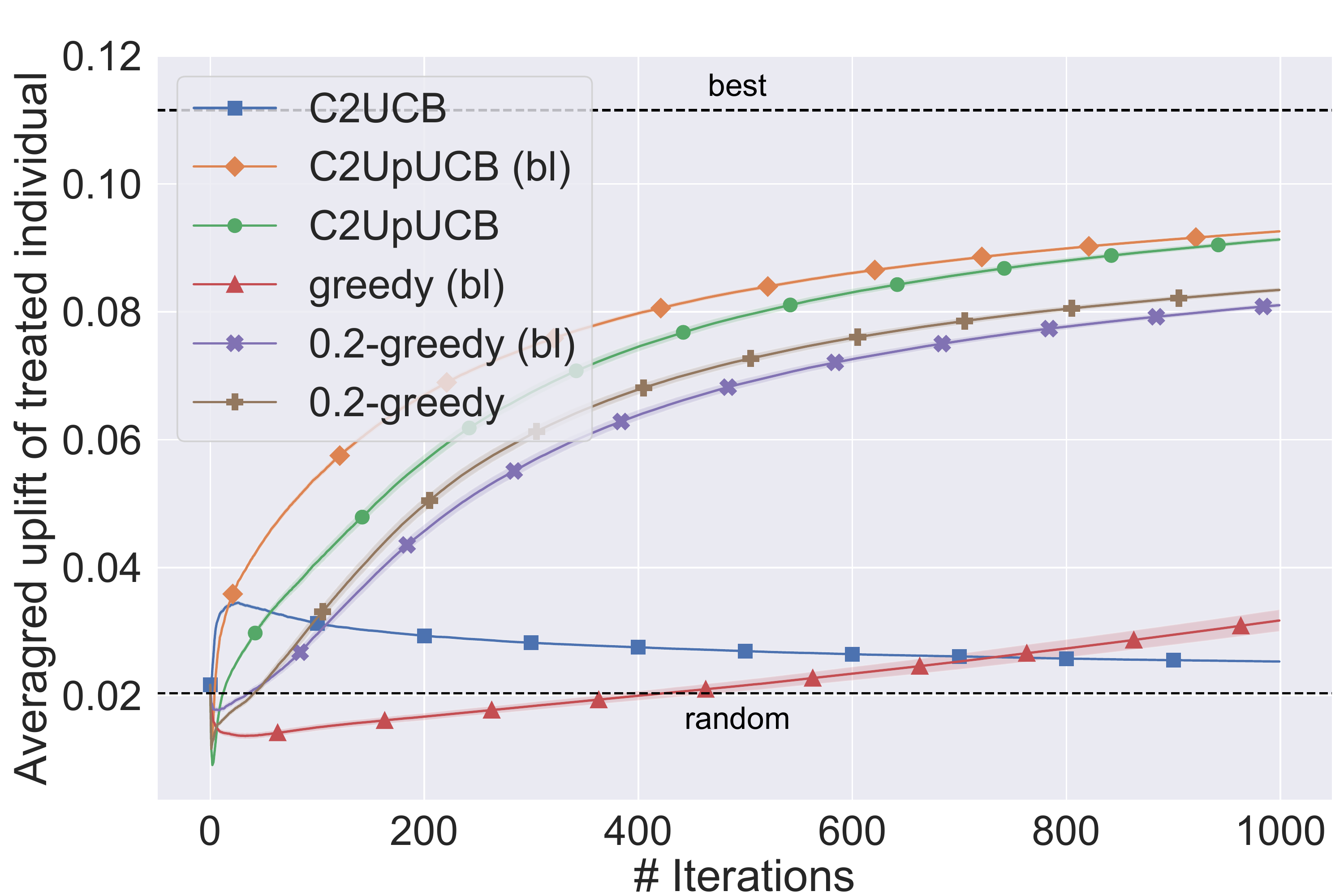}
    \caption{Logistic models used to generate all the outcomes.}
    \label{subfig:exp-c2-add-without-true}
    \end{subfigure}
    \caption{Results on the targeted campaign experiment. For the top row  the logistic models are only used to generate counterfactual outcomes, and in the bottom row all the outcomes are generated using the predicted conditional probabilities of the logistic models. We plot regret computed using the predicted uplifts and the realized rewards, and the predicted uplift averaged over all the treated individuals.}
    \label{fig:exp-c2-add}
    \vspace{-1em}
\end{figure}

\paragraph{Feedback\afterhead}
We consider two ways to use the fitted logistic models
\begin{enumerate}
    \item The logistic models are only used to generate the counterfactual outcomes.
    That is, the feedback of an individual is sampled from the Bernoulli distribution with mean predicted by either of the models only if the learner decides to treat the individual while the individual is not treated in the dataset or vice-versa.
    \item The logistic models are used to generate all the outcomes.
\end{enumerate}

\paragraph{Results\afterhead}

In \cref{fig:exp-c2-add} (Right), we plot the uplift averaged over all the individuals that have been treated up to a time horizon by an algorithm.
Higher this value is better it is because it means we are selecting the individuals for which the treatment is the most effective.
For comparison, we also indicate the average uplift of random selection and an oracle strategy that always chooses the $10$ individuals with the highest uplifts.
%
Both \CUpUCB and $\varepsilon$-greedy beat greedy and random selection,
which demonstrates the benefit of considering a bandit model for this problem.
$\varepsilon$-greedy turns out to be quite competitive, but \CUpUCB still has a slight advantage no matter whether the baseline is known or unknown.


For further comparison, we also plot the regret in terms of both the predicted uplifts (Left) and the rewards that the learner actually receives (Middle) in \cref{fig:exp-c2-add}.\footnote{We use occasionally the term predicted uplift here to remind that they are computed using logistic models fitted on the dataset and we do not have access to any type of \emph{ground-truth} uplift.}
For the two regrets, we define the optimal action as the one that selects the $10$ individuals with the highest predicted uplifts.
Then, if the logistic models correctly predict the expected rewards, the regret in terms of the predicted uplifts should be the expectation of the regret in terms of the realized rewards, and thus we should get similar curves after averaging over $100$ runs.
This is indeed the case in \cref{subfig:exp-c2-add-without-true}.
Nonetheless, 
when the outcomes also come from true data,
selecting the samples with higher predicted uplifts does not necessarily result in higher rewards, which explains the discrepancy between \cref{subfig:exp-c2-add-with-true} (Left) and \cref{subfig:exp-c2-add-with-true} (Middle).
\CUpUCB and $\varepsilon$-greedy perform quite similarly here when we look at the realized rewards.
Since the ground truth outcomes cannot be perfectly predict by logistic models,
we also observe linear regret of all the algorithms in \cref{subfig:exp-c2-add-with-true} (Left), while in \cref{subfig:exp-c2-add-without-true} (Left), both \CUpUCBMwb and \CUpUCB achieves sublinear regret and performs much better than $\varepsilon$-greedy.

In all the cases, Greedy and C2UCB perform poorly.
This points out the importance of exploration (which is taken into account by bandit algorithms) and uplift modeling in this problem.
Finally, to explain the surprising fact that an algorithm might perform better without knowledge of the baseline, we look into the data and find out that samples with higher baselines often have higher uplifts.
Therefore, when the learner is provided with the baselines of the samples, they may avoid choosing these samples at the beginning, which results a slower exploration and a worse performance.




\section{Additional Experimental Details}
\label{apx:exp-detail}
In this appendix, we present missing experimental details from~\cref{sec:exp}. The code implementation of the proposed methods is attached in supplementary material.
All our experiments are run on amazon m4.xlarge EC2 instances.
The Criteo Uplift Prediction Dataset that we use for our experiments has the `CC0: Public Domain' license.


\subsection{Choice of Algorithm Parameters}
\label{apx:param-tuning}

As mentioned in the main text, we report the results for the parameters that yield the best average performance.
Precisely, we conduct $100$ independent runs for each experimental setup, and in each run, we fix the random seed and test all the candidate algorithms with their parameters chosen from a predefined set.
The performance of a specific parameter choice is then evaluated by the sum of the mean and one standard deviation of the regrets of the final run.
In fact, when the algorithm puts too much weight on exploitation, we may get lower mean regret while the standard deviation gets extremely large because the algorithm fails more frequently. 

Below we explain in detail what are the parameters that we tune in the non-contextual experiments.
The optimal parameters that we have found are summarized in \cref{tab:params}.


\begin{table}[t]
\setlength{\tabcolsep}{1pt}
\centering
\caption{Parameter values used for the reported experimental results. TS stands for Thompson sampling. Please refer to the text for a explanation on what is the parameter that we tune for each of the algorithms.}
\label{tab:params}
\vskip 0.4em
\begin{tabularx}{\textwidth}{lcccccc}
    \toprule
    Non-contextual & UCB & \UpUCB & \UpUCBwb & \UpUCBL & \UpUCBLwb & TS \\
    \cmidrule{2-7}
    Gaussian 
    & $2$ & $5$ & $3$ & $5$ & $3$ & $8\cdot 10^{-2}$ \\
    Bernoulli &
    $7 \cdot 10^{-7}$ & $8 \cdot 10^{-5}$ & $8 \cdot 10^{-5}$ &  $5 \cdot 10^{-7}$ & $2 \cdot 10^{-6}$  & $2 \cdot 10^{-7}$
    \\
    \midrule
    Contextual && C2UCB & \CUpUCB(b) & \CUpUCB & $\varepsilon$-greedy\,(b) & $\varepsilon$-greedy \\
    \cmidrule{3-7}
    \multicolumn{2}{l}{Logistic for counterfactual} & 2 & 2 & 2 & 0.2  & 0.1\\
    \multicolumn{2}{l}{Logistic for all} & 2 & 2 & 2 & 0.2  & 0.2\\
    \bottomrule
\end{tabularx}
\end{table}

\paragraph{UCB and \textsc{UpUCB}s\afterhead}
For all UCB-type methods, we tune the exploration parameter $\expparam=1/\alt{\smallproba}$, which plays a crucial role in determining the widths of the confidence intervals-- see \eqref{eq:mean-estimate-a} and \eqref{eq:mean-estimate-0}.
Since $\vta[\radiusConf]$ represents the width of the confidence interval of a single variable's payoff in UpUCB algorithms, we set the widths of the confidence intervals of UCB to $\nVariables\vta[\radiusConf]=\nVariables\sqrt{2\expparam/\vta[\pullcount]}$, \ie the UCB indices in UCB are $\vta[\UCBindex]=\vta[\est{\reward}]+\nVariables\vta[\radiusConf]$, where $\vta[\est{\reward}]$ is the empirical mean of action $\arm$'s reward up to time $\run$.
In this way, we can choose exploration parameters of roughly the same order for different algorithms.
In the Gaussian uplifting bandit experiment (\cref{subsec:exp-gaussian}), we pick $\expparam\in\oneto{10}=\intinterval{1}{10}$.
As for the Bernoulli uplifting bandit experiment with Criteo dataset (\cref{subsec:exp-ber}), we select $\expparam\in\setdef{k\times10^{-\gamma}}{k\in\oneto{10},\gamma\in\{5,6,7\}}$.
The optimal performances are achieved with much smaller exploration parameters here because the noises in the variables are independent.

\paragraph{Thompson Sampling.}
We use Thompson sampling with Gaussian prior and Gaussian noise.
The mean and the variance of the prior are set to the mean and the variance of the actions' rewards.
Therefore, Thompson sampling is actually provided with more information than UCB.
Similar to before, the variance of the noise is set to $\nVariables^2\noisedev^2$ since the noise in the total reward scales with $\nVariables$.
The parameter $\noisedev^2$ plays a similar role as $\expparam$ in UCB.
In the Gaussian and Bernoulli uplifting experiments we respectively choose $\noisedev^2\in\setdef{k\times10^{-\gamma}}{k\in\oneto{10},\gamma\in\{0,1,2\}}$ and $\noisedev^2\in\setdef{k\times10^{-\gamma}}{k\in\oneto{10},\gamma\in\{5,6,7\}}$.

\subsection{More Details on Bernoulli Uplifting Bandit}
\label{apx:exp-ber}


For sake of completeness, we report below 
the visit rate of the treated and the untreated individuals of the clusters (rounded to three decimal places) that we obtained for our experiments
\begin{equation}
    \notag
    \begin{aligned}
    \meanReward_{\text{treated}}
    = [&.001 ~.037 ~.003 ~.001 ~.003 ~.377 ~.237 ~.309 ~.071 ~.287 \\
    &~.531 ~.044 ~.007 ~.086 ~.002 ~.019 ~.028 ~.007 ~.265 ~.013]\\
    \meanReward_{\text{untreated}}
    = [&.001 ~.023 ~.002 ~.002 ~.004 ~.289 ~.206 ~.229 ~.073 ~.289 \\
    &~.464 ~.035 ~.004 ~.052 ~.001 ~.011 ~.022 ~.004 ~.165 ~.000].
    \end{aligned}
\end{equation}

\noindent
The sizes of the clusters, which also correspond to $(\va[\nAffected])_{\arm\in\arms}$, are
\begin{align}
    \notag
    [&10600 ~2764 ~7222 ~11128 ~6385 ~1630 ~2806 ~1089 ~3018
     ~4594\\
     \notag
     &~594 ~7020 ~12654 ~2186 ~9609 ~5101 ~3714 ~4569
     ~1158 ~2159].
\end{align}

\newcommand{\shortminus}{\scalebox{0.75}[1.0]{\( - \)}}

\noindent
The uplifts of the actions (rounded to one decimal place) are
\begin{equation}
    \notag
    \begin{aligned}
    \up{\reward} =
    [&0.6 ~\,39.3 ~\,6.6 ~\,\shortminus5.0 ~\,\shortminus5.7 ~\,143.5 ~\,86.0 ~\,87.1 ~\,\shortminus4.5 ~\,\shortminus6.9 
    \\
    &~\,39.8 ~\,66.3 ~\,34.7 ~\,75.4 ~\,4.0
    ~\,40.3 ~\,21.1 ~\,12.3 ~\,115.7 ~\,28.3].
    \end{aligned}
\end{equation}

\end{document}